\pdfoutput=1
\documentclass[dvipsnames]{article}
\PassOptionsToPackage{numbers,sort&compress}{natbib}
\usepackage[T1]{fontenc}
\usepackage{unipat_report}
\setcitestyle{numbers,square,citesep={,}}

\usepackage[utf8]{inputenc}
\usepackage{booktabs}
\usepackage{amsfonts}
\usepackage{amsmath}
\usepackage{nicefrac}
\usepackage{colortbl}
\usepackage{graphicx}
\usepackage{float}
\usepackage{multirow}
\usepackage{makecell}
\usepackage[inline]{enumitem}
\usepackage[skins,breakable]{tcolorbox}
\usepackage{pifont}
\usepackage{subcaption}
\usepackage{placeins}
\usepackage{xspace}
\newcommand{\cmark}{\textcolor{ForestGreen}{\ding{51}}}
\newcommand{\xmark}{\textcolor{BrickRed}{\ding{55}}}
\newcommand{\pmark}{\textcolor{BurntOrange}{$\boldsymbol{\triangle}$}}

\tcbset{
  taskbox/.style={
    enhanced, breakable,
    colframe=blue!55!black,
    colback=blue!2!white,
    colbacktitle=blue!55!black,
    coltitle=white,
    fonttitle=\small\bfseries,
    fontupper=\small,
    boxrule=0.8pt,
    arc=3pt,
    top=5pt, bottom=5pt, left=7pt, right=7pt,
    before upper={\setlength{\parindent}{0pt}\setlength{\parskip}{3pt}},
  },
  tracebox/.style={
    enhanced, breakable,
    colframe=#1!70!black,
    colback=#1!4!white,
    colbacktitle=#1!70!black,
    coltitle=white,
    fonttitle=\small\bfseries,
    fontupper=\small,
    boxrule=0.6pt,
    arc=2pt,
    top=4pt, bottom=4pt, left=6pt, right=6pt,
    before upper={\setlength{\parindent}{0pt}\setlength{\parskip}{3pt}},
  },
  findingbox/.style={
    enhanced,
    colframe=blue!45!black,
    colback=blue!3!white,
    boxrule=0.5pt,
    arc=2pt,
    left=6pt, right=6pt, top=3pt, bottom=3pt,
    fontupper=\small,
  }
}

\definecolor{pendingbg}{HTML}{FFF0B3}
\newtcolorbox{pending}{enhanced, breakable, colback=pendingbg, colframe=pendingbg, boxrule=0pt, arc=1.5pt,
  left=4pt, right=4pt, top=2pt, bottom=2pt, before skip=6pt, after skip=6pt,
  before upper={\setlength{\parindent}{0pt}\setlength{\parskip}{.5pc}}}

\newtcolorbox{pendingfloat}{enhanced, colback=pendingbg, colframe=pendingbg, boxrule=0pt, arc=1.5pt,
  left=3pt, right=3pt, top=2pt, bottom=2pt, before skip=1pt, after skip=1pt,
  before upper={\setlength{\parindent}{0pt}\setlength{\parskip}{.5pc}}}

\newcommand{\MYBENCH}{\textsc{SPU}\xspace}

\title{Science Prediction Universe}

\author{}

\newcommand{\SLEInstitutionMark}[3]{%
  \makebox[27pt][c]{%
    \raisebox{-0.5\height}{%
      \includegraphics[#1,height=#2pt]{logo/institution-marks/#3}%
    }%
  }%
}

\newcommand{\SLEInstitutionMarks}{%
  \makebox[\linewidth][c]{\hbox{%
    \SLEInstitutionMark{keepaspectratio}{22}{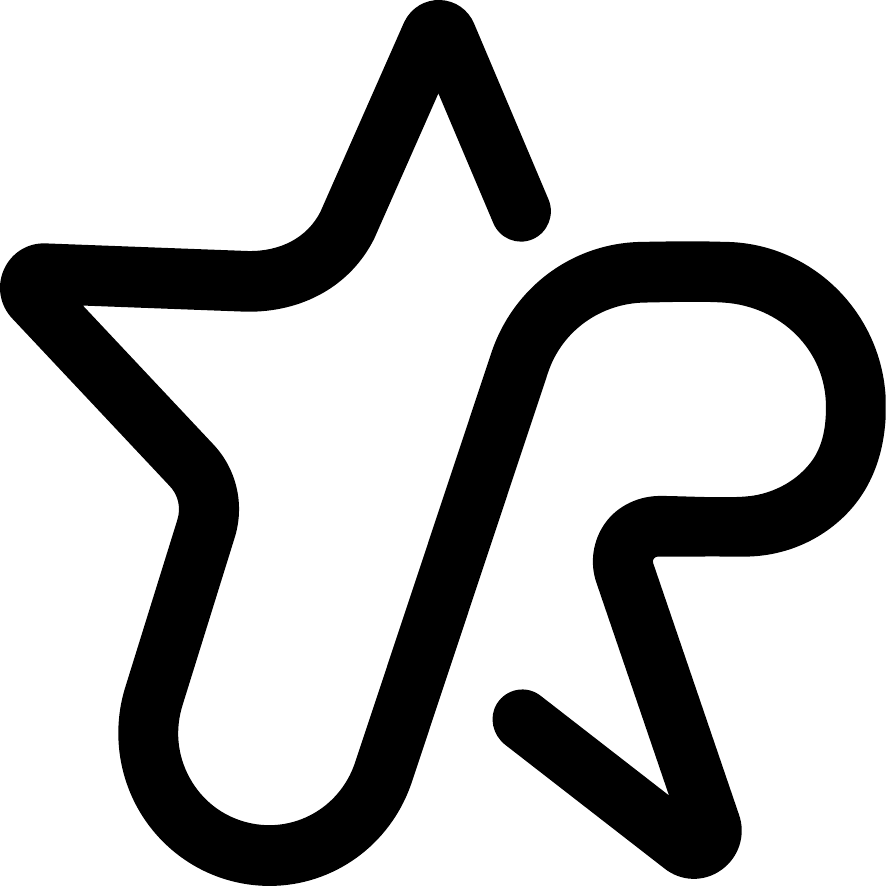}\hspace{7pt}%
    \SLEInstitutionMark{trim={100bp 50bp 100bp 55bp},clip}{26}{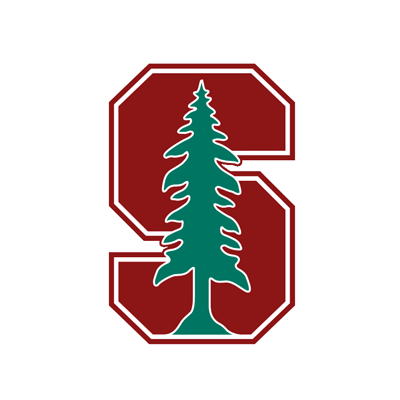}\hspace{7pt}%
    \SLEInstitutionMark{keepaspectratio}{24.44}{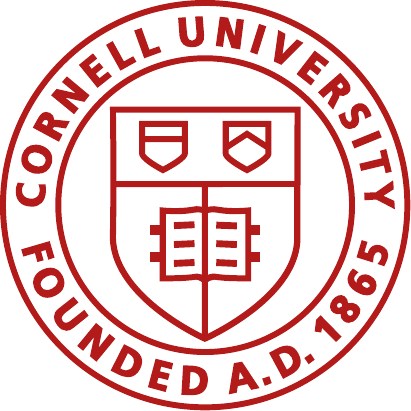}\hspace{7pt}%
    \SLEInstitutionMark{trim={0bp 14bp 0bp 14bp},clip}{26}{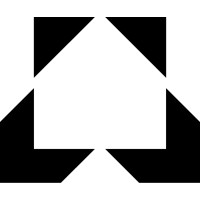}\hspace{7pt}%
    \SLEInstitutionMark{keepaspectratio}{24.44}{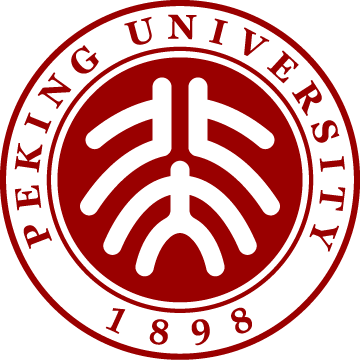}\hspace{7pt}%
    \SLEInstitutionMark{trim={26bp 14bp 26bp 14bp},clip}{26}{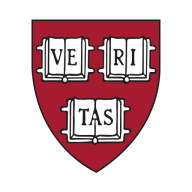}\hspace{7pt}%
    \SLEInstitutionMark{trim={0bp 0bp 492bp 0bp},clip}{26}{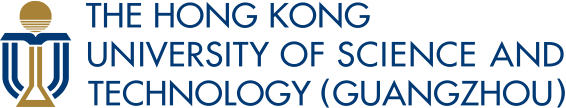}\hspace{7pt}%
    \SLEInstitutionMark{keepaspectratio}{24.44}{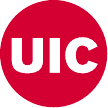}\hspace{7pt}%
    \SLEInstitutionMark{trim={84bp 57.84bp 2040.24bp 0bp},clip}{26}{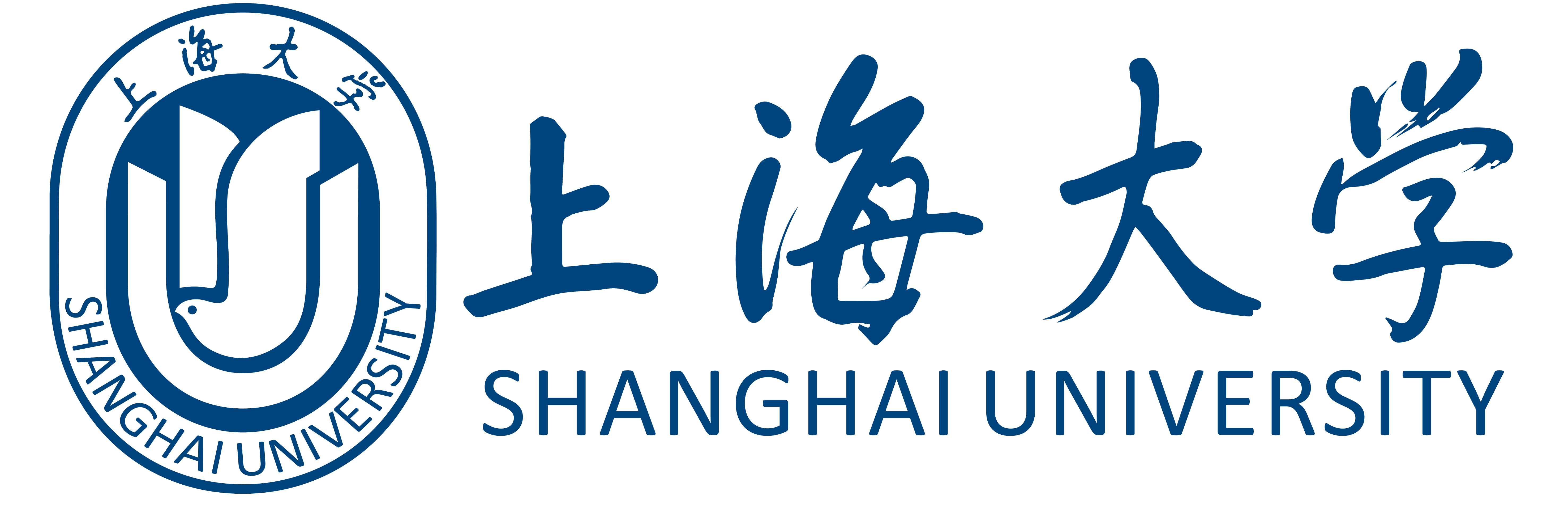}\hspace{7pt}%
    \SLEInstitutionMark{trim={0bp 0bp 567bp 0bp},clip}{26}{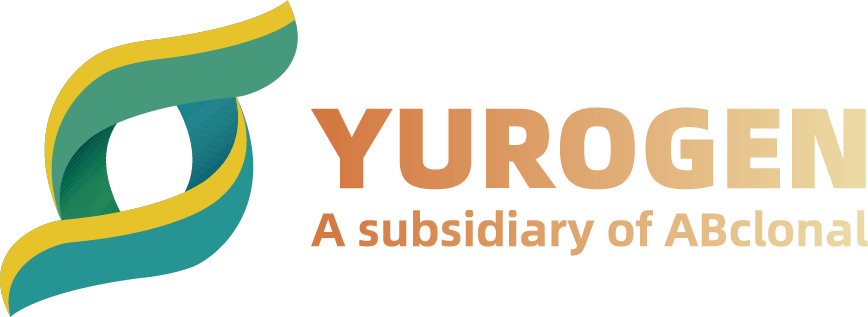}\hspace{7pt}%
    \SLEInstitutionMark{trim={0bp 0bp 316bp 0bp},clip}{26}{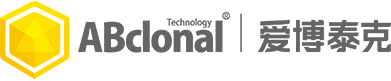}\hspace{7pt}%
    \SLEInstitutionMark{trim={43.2bp 60.72bp 42.72bp 7.2bp},clip}{26}{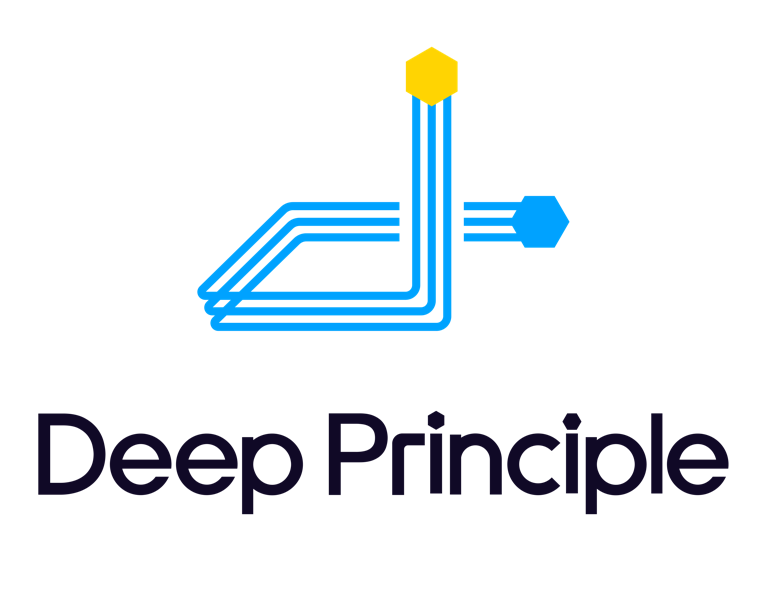}\hspace{7pt}%
    \SLEInstitutionMark{trim={0bp 0bp 692bp 0bp},clip}{26}{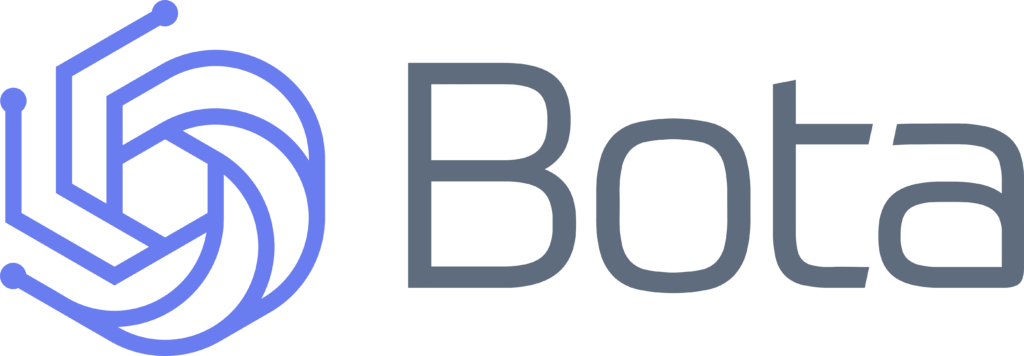}%
  }}%
}

\begin{document}

\maketitle
\vspace{-2.8em}
\begingroup
\setlength{\parindent}{0pt}
\setlength{\parskip}{0pt}
\centering\small
\rule{\linewidth}{0.6pt}\par\vspace{4pt}

{\normalsize\bfseries Project Team}\par\vspace{1pt}
Zengji Tu\textsuperscript{*}, Xiang Liu\textsuperscript{*}, Jiarui Sun, Ruoyu Wu, Zhihui Qi, Chengda Wu,\\
Yutao Zhou, Bo Yang, Ziyan Zhang, Jialong Wu, Liang Chen\textsuperscript{\dag,\ddag}, Kuan Li\textsuperscript{\dag}\par
\vspace{0pt}
{\footnotesize \textsuperscript{*}\,Core contributors.\quad \textsuperscript{\dag}\,Corresponding authors.\quad \textsuperscript{\ddag}\,Project leader.\par}
\vspace{3pt}

{\normalsize\bfseries Advisory Committee}\par\vspace{1pt}
Steven Dillmann, Yuanqi Du, Soojung Yang, Shizhe Diao, Jianjun Wu, Junlang Liu,\\
Xuming Hu, Lin Chang\textsuperscript{\dag}, Baobao Chang, Xiaowen Chu, Philip S. Yu, Tong-Yi Zhang\par
\vspace{3pt}

{\normalsize\bfseries Wet-Lab Collaborators}\par\vspace{1pt}
Hao Cheng, Qin Han, Yanye Feng, Chenru Duan, Chuncheng Liu, Yan Liu,\\
Yuchen Gao, Zhao Li, Ronghua Zhang, Tory Xu, Renjian Song, Yaokai Fu\par
\vspace{3pt}

{\footnotesize Leading institution: UniPat AI.\quad Affiliations and contact details in Appendix~\ref{app:authors}.\par}
\vspace{4pt}
\noindent\SLEInstitutionMarks\par
\endgroup
\vspace{2pt}

\begin{abstract}
Humanity has long sought to explain natural phenomena. Reliable prediction of experimental outcomes is a central test of our understanding of the processes that produce them. If artificial general intelligence (AGI) is to surpass human scientific capabilities, it must also predict experimental outcomes more reliably than humans. Such capabilities could accelerate scientific discovery across disciplines, help address humanity's most pressing challenges, and improve human well-being. We introduce \textbf{Science Prediction Universe (\MYBENCH)}, a benchmark of hidden experimental outcomes with 100 public verified tasks from chemistry, materials science, biology and medicine, and 16 private tasks from unpublished wet-lab experiments. Given an experimental setup and candidate conditions, an agent works with any resources it needs to rank the candidates by a measured outcome.

We evaluate thirteen agent configurations spanning twelve frontier models across two harnesses (Codex and Claude Code). Current agents remain far from saturating wet-lab outcome prediction, with substantial performance disparities across models: on the public benchmark, configurations separate into two distinct tiers, with task-adjusted Spearman correlations ranging from 0.227 for the lowest-performing configuration to 0.478 for the best. Analysis of 1,155 scored public trajectories (227,103 tool calls) suggests that the bottleneck lies in scientific judgment rather than tool execution: agents rely on shared monotone priors, agreeing far more with one another than with experimental reality (0.52 vs.\ 0.33), and frequently compute trivial physical or chemical proxies that do not influence the final outcome.
\end{abstract}

\begin{figure}[!htbp]
  \centering
  \includegraphics[width=\linewidth]{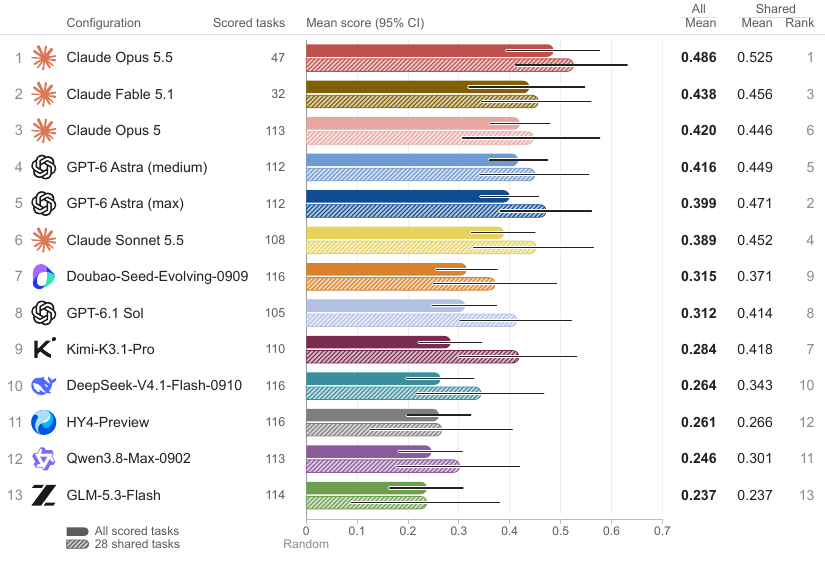}
  \caption{\textbf{Configurations ranked by mean score on all scored tasks.} Solid bars are means over the tasks, among the 100 public and 16 private tasks, on which the configuration has a scored run (Scored tasks). Hatched bars are means over the 28 shared tasks, the 27 public tasks and one private task on which all thirteen configurations have a scored run; the far-right column ranks the configurations on these tasks. Claude Opus 5.5 and Claude Fable 5.1 ended most of their runs in safety refusals, mainly on chemistry, biology and medicine tasks, so they are scored on fewer tasks and only 28 tasks are shared by all thirteen configurations, against 94 shared by the other eleven. Lines are 95\% bootstrap intervals over tasks, with the same resampled tasks for every configuration on the shared tasks. A perfect ranking scores 1, and random rankings average 0. GPT-6 Astra ran in Codex at medium and max reasoning effort, GPT-6.1 Sol in Codex at max effort, and the other models in Claude Code at max effort.}
  \label{fig:model-score-ranking}
\end{figure}

\section{Introduction}
\label{sec:intro}

Science advances by prediction. A theory earns trust when it correctly forecasts what an experiment not yet performed will show, and loses it when the measurement disagrees~\citep{popper1959logic}: Mendeleev's periodic table predicted the density and atomic weight of gallium years before it was isolated, and measurement settled the matter. For an AI system, predicting a wet-lab outcome before the experiment is run is therefore the step that connects reasoning to reality. Such an outcome cannot be recalled from training data or won by flattering a judge model. We gather such predictions, across chemistry, materials science, biology and medicine, into the \textbf{Science Prediction Universe (\MYBENCH)}, a growing collection of experiments whose outcomes stay hidden until a prediction is committed.

Language-model agents can now search the literature~\citep{skarlinski2024paperqa2}, install scientific software, run quantum-chemistry and molecular-dynamics calculations, and write analysis code for hours without supervision~\citep{yang2024sweagent,wang2025openhands}. Chemistry agents plan and execute syntheses~\citep{boiko2023coscientist,bran2024chemcrow} and drive simulation workflows~\citep{zou2025elagente,campbell2025mdcrow}, and designs from multi-agent systems have been validated in the wet lab~\citep{swanson2025virtuallab,gottweis2026coscientist,ghareeb2025robin}. Yet many questions that matter in experimental science have no computed answer. Which electrolyte additive will extend the life of a zinc battery, or which enzyme variant will shift oxidation selectivity toward the desired product, is settled by synthesis and measurement. Computation usually narrows the candidate list~\citep{greeley2006screening,mamun2019bimetallic}, but the answer still comes from the experiment. The question is therefore:

\begin{center}
\begin{minipage}{0.92\linewidth}\centering
\textbf{Given the same experimental definition a scientist would have before running the experiment, can an autonomous agent predict how it will turn out?}
\end{minipage}
\end{center}

Exam-style frontier benchmarks such as Humanity's Last Exam~\citep{phan2025hle} and Agents' Last Exam~\citep{sun2026ale} push the boundaries of expert-level reasoning and long-horizon tool use, but they grade answers that are already known, whereas discovery turns on outcomes that are not~\citep{du2026measuring}. Current scientific evaluations answer this question only partly (Section~\ref{sec:related}; Table~\ref{tab:comparison}). Outcome-prediction benchmarks such as BrainBench~\citep{luo2024brainbench}, SciPredict~\citep{sehwag2026scipredict} and AssayBench~\citep{debrouwer2026assaybench} ask models to predict real experimental results, but mostly as single question-answer calls without tools, so they measure what a model already knows rather than what it can find out. Agentic science benchmarks such as ScienceAgentBench~\citep{chen2025scienceagentbench} and BixBench~\citep{mitchener2025bixbench} let agents write and execute code, but the target is an analysis of data that is already in hand, and long-horizon container benchmarks such as Terminal-Bench-Science~\citep{tbscience2026} check answers with deterministic tests rather than with unknown wet-lab measurements. None of these settings tests the step that is most consequential in practice: choosing which evidence and which calculation bear on a measured endpoint, and committing to a prediction when no experimental feedback is available.

\begin{figure}[!htbp]
  \centering
  \includegraphics[width=\linewidth]{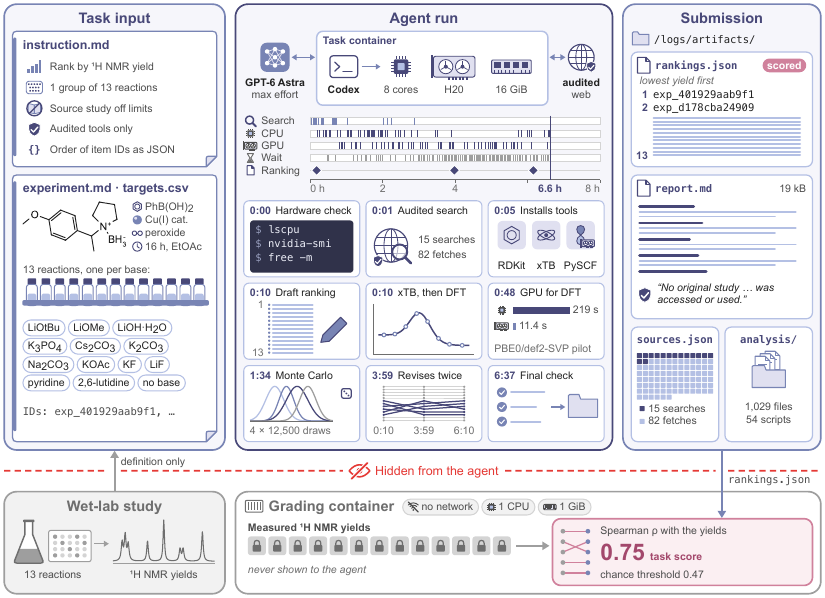}
  \caption{\textbf{The \MYBENCH setting, shown with one official run.} Left, the task: rank 13 copper-catalyzed deaminative arylations of one amine-borane, which differ only in the base additive, by measured \textsuperscript{1}H NMR yield. Middle, GPT-6 Astra at max reasoning effort worked in Codex for 6.6 of the 8 available hours, reaching the web only through the audited service and revising its ranking twice. Right, the files the run submitted. Bottom, the measured yields stay in a separate grading container and are never shown to the agent; this submission scores 0.75, above the task's chance threshold of 0.47.}
  \label{fig:teaser}
\end{figure}

To address this gap we introduce the \textbf{Science Prediction Universe (\MYBENCH)}: 100 public tasks built from 85 papers published in the last two years in 11 Nature Portfolio journals, and 16 private tasks that experts designed from unpublished wet-lab experiments carried out in collaboration with scientific laboratories, none of whose results has been made public. Each task fixes the candidates, materials, conditions and measurement protocol of one experiment and asks the agent to rank the candidates by the measured outcome, which is held only by an isolated grader (Figure~\ref{fig:teaser}; \S\ref{sec:evaluation}). The agent runs in its own container with 8 CPU cores, 16\,GiB of memory, one NVIDIA H20 GPU and an eight-hour budget, and may search the web, download software and compute freely. Because the answers to public tasks can in principle be found by search, we combine four safeguards against retrieval and memorization: an isolated grader and anonymized tasks, an information policy with audited tools and a stop-on-exposure rule, automated screening of tool outputs for source identifiers, and private tasks that are not online (\S\ref{sec:audit}).

We evaluate thirteen agent configurations (Figure~\ref{fig:model-score-ranking}): GPT-6 Astra at medium and at max reasoning effort and GPT-6.1 Sol at max effort in the Codex CLI, and ten other models at max effort in Claude Code (Claude Opus 5.5, Claude Opus 5, Claude Sonnet 5.5, DeepSeek-V4.1-Flash-0910, Qwen3.8-Max-0902, GLM-5.3-Flash, Doubao-Seed-Evolving-0909, HY4-Preview, Kimi-K3.1-Pro and Claude Fable 5.1). Each configuration runs once on every task, totaling 1,300 official runs on the public tasks and 208 on the private tasks; across configurations, the median run time of scored public runs ranges from 0.26 to 3.59 hours and the median number of tool calls from 29 to 231. Our evaluation yields four main findings:
\emph{First}, wet-lab outcome prediction remains far from solved. On the public tasks, the best task-adjusted mean is 0.478 (Table~\ref{tab:public-results}), and configurations exceed the chance threshold on only 34\% to 52\% of their scored tasks (Figure~\ref{fig:public-scores}c). On the 16 private tasks, which involve no model-based selection, twelve of the thirteen configurations rank the unpublished experiments significantly better than random.
\emph{Second}, the configurations fall into two tiers. Claude Opus 5.5, Claude Fable 5.1, Claude Opus 5, the two GPT-6 Astra configurations and Claude Sonnet 5.5 achieve task-adjusted means of 0.393 to 0.478, whereas the other seven score between 0.227 and 0.315; all 17 pairwise differences that survive Holm correction lie between these two groups, while the ordering within each group remains unresolved at one run per task. GPT-6.1 Sol, at 0.308, differs detectably from no configuration.
\emph{Third}, more computation did not yield detectably more accurate predictions. Running GPT-6 Astra at max instead of medium effort increases run time by a median factor of 4.8 without raising the score (the 95\% interval excludes a gain above 0.031), and across configurations score shows no detectable correlation with run time, tool calls, web fetches or GPU use. Runs flagged for encountering identifiers of the source study show no statistically detectable advantage in any configuration.
\emph{Fourth}, trajectory analyses and a close reading of selected runs point to why predictions fall short (\S\ref{sec:analysis}). Agents rely heavily on monotone priors and agree much more with one another than with the measurements (0.52 vs.\ 0.33); when they run quantum-chemistry or biophysical simulations, the runs that fall short typically compute tractable proxies that do not govern the measured endpoint, or revise accurate early drafts on the basis of unverified assumptions.

\vspace{-5pt}
\paragraph{Contributions.}
\begin{itemize}[leftmargin=*,itemsep=2pt,topsep=2pt]
  \item \textbf{A benchmark of hidden experimental outcomes for autonomous agents.} 100 public ranking tasks in chemistry, materials science, biology and medicine, built from 85 recent papers in 11 Nature Portfolio journals, with source studies and data chosen by domain experts and every task reviewed independently (\S\ref{sec:construction}), plus 16 private tasks designed by experts from unpublished wet-lab experiments that we carried out in collaboration with scientific laboratories (\S\ref{sec:private}).
  \item \textbf{An evaluation protocol against retrieval and memorization.} An isolated grader with anonymized tasks, an information policy with audited tools and a stop-on-exposure rule, automated screening of tool outputs for source identifiers, and private tasks that cannot be found online: four safeguards that jointly constrain retrieval and memorization under open web access and keep scientific scores separate from compliance verdicts (\S\ref{sec:audit}).
  \item \textbf{A systematic evaluation of thirteen agent configurations.} Twelve models in two harnesses, Codex and Claude Code, with each configuration run once on all 116 tasks; we report coverage alongside task-adjusted means, Holm-corrected paired tests and run-to-run variation estimated from repeated runs (\S\ref{sec:experiments}).
  \item \textbf{A trajectory-level analysis of how science agents work and where they fail.} Across 1,155 scored public trajectories (227,103 tool calls) and a close reading of selected runs, we examine how agents allocate compute, construct and revise rankings, use the GPU and adhere to the information policy (\S\ref{sec:experiments}, \S\ref{sec:analysis}), suggesting that runs are separated by the quantity they choose to rank by rather than by the tools they execute: no method family or behavior is detectably associated with score, while the configurations agree with one another more than with the measurements.
\end{itemize}

\section{Science Prediction Universe}
\label{sec:benchmark}
\label{sec:integrity}

The Science Prediction Universe evaluates whether an autonomous agent, given the setup of real wet-lab experiments, can predict how the experiments rank on measured outcomes that are withheld from it. Public tasks are built from studies published in peer-reviewed journals, and the information policy prohibits retrieving their results or reconstructing them from memory (Section~\ref{sec:audit}). Private tasks are designed by experts from real wet-lab experiments that we carried out in collaboration with scientific laboratories. These experiments have not been published or released in any form, so no model could have been trained on their results and no agent can find them online. This section defines the task, summarizes the benchmark composition, describes how the public tasks were constructed, presents the private tasks and specifies the evaluation protocol.

\subsection{Task Definition}
\label{sec:task}

Each public task is built from experiments reported in a published wet-lab study. The experiments are organized into comparison groups, each with one measured endpoint, such as a reaction yield or an ionic conductivity. For each group, the agent orders the experiments from the lowest to the highest measured value; it does not have to predict the values themselves (Figure~\ref{fig:teaser}). The task instruction states the prediction question, the comparison groups, the information policy of Section~\ref{sec:audit} and the submission format. A set of input files describes the shared procedure, the settings of each experiment and, where needed, structures or sequences. The agent works in a computing environment with 8 CPU cores, 16\,GiB of memory and one NVIDIA H20 GPU, where it can run code and install software for up to 8 hours. Experiments carry only anonymized identifiers, and the agent is not given the source study or told which study it is.

The agent submits \texttt{rankings.json} with one strict ordering per comparison group, together with a report, a source record and supporting scripts. Appendix~\ref{app:task_example} reproduces a task instruction with light abridgment. The measured values and the grading code stay outside the agent's environment. The grader scores each group by the rank correlation between the submitted order and the measured values (Section~\ref{sec:evaluation}).

\subsection{Task Statistics}
\label{sec:stats}

\begin{figure}[!htbp]
  \centering
  \includegraphics[width=\linewidth]{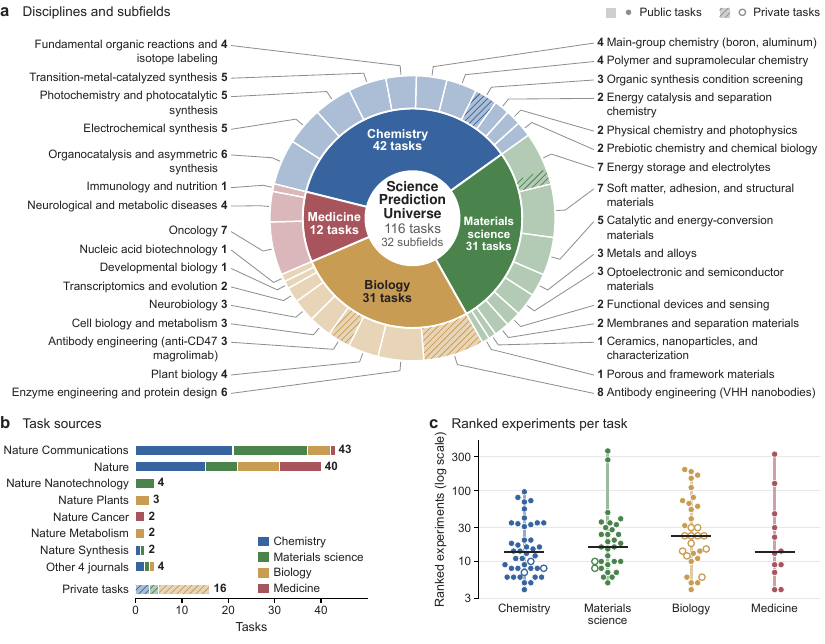}
  \caption{\textbf{Coverage of the Science Prediction Universe.} \textbf{a}, The 116 tasks by discipline and subfield; sector angles are proportional to task counts, and hatching marks private tasks. \textbf{b}, Sources of the tasks: 85 papers in 11 Nature Portfolio journals, published online between January 2025 and August 2026, for the 100 public tasks, and four series of unpublished wet-lab experiments for the 16 private tasks; journals with one source paper are pooled. \textbf{c}, Number of experiments in each task on a logarithmic scale, with open circles for private tasks and black ticks at discipline medians.}
  \label{fig:task-characteristics}
\end{figure}

The Science Prediction Universe contains 116 tasks in four disciplines: chemistry (42 tasks), materials science (31), biology (31) and medicine (12). The tasks fall into 32 subfields, such as organocatalysis and asymmetric synthesis, energy storage and electrolytes, enzyme engineering and protein design, and oncology (Figure~\ref{fig:task-characteristics}a). The 100 public tasks are built from 85 papers published online between January 2025 and August 2026 in 11 Nature Portfolio journals. Seventy papers contribute one task each and 15 contribute two; most appeared in \emph{Nature Communications} (41 papers, 43 tasks) or \emph{Nature} (28 papers, 40 tasks; Figure~\ref{fig:task-characteristics}b). The 16 private tasks come from four series of unpublished experiments and add three subfields: antibody engineering with VHH nanobodies, antibody engineering with an anti-CD47 monoclonal antibody, and organic synthesis condition screening (Section~\ref{sec:private}).

Task size varies widely (Figure~\ref{fig:task-characteristics}c). A task contains 4 to 362 experiments (median 16) in 1 to 117 comparison groups. The benchmark has 4,207 experiments in 517 groups, of which the public tasks account for 3,947 experiments and 489 groups. The measured endpoints include reaction yields, ionic conductivities, tensile strengths, fluorescence signals and RNA-seq gene counts.

\subsection{Public Task Construction}
\label{sec:construction}

\begin{figure}[!htbp]
  \centering
  \includegraphics[width=\linewidth]{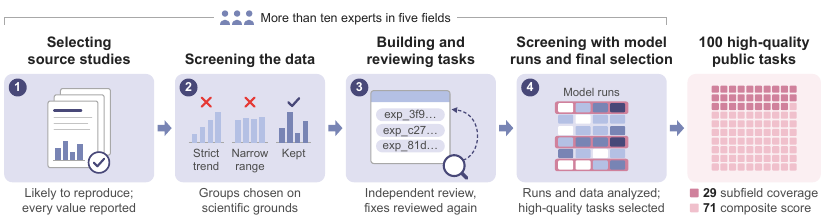}
  \caption{\textbf{Public Task Construction.} More than ten experts in five fields built the public tasks in four steps. Step 2 avoids two kinds of comparison (crossed charts): orderings that rise or fall strictly with one varied parameter, and values that differ only slightly. In Step 3, each task is reviewed independently, and a task with a blocking issue is fixed and reviewed again (dashed arrow). Step 4 selects tasks by pilot scores, mainly those of GPT-6 Astra; the 100 squares on the right are the public tasks, 29 giving every subfield a task and 71 chosen by composite score.}
  \label{fig:public-task-pipeline}
\end{figure}

More than ten experts in biology, chemistry, materials science, medicine and computer science took part in building the public tasks. Construction took four steps (Figure~\ref{fig:public-task-pipeline}).

\vspace{-5pt}
\paragraph{Step 1: Selecting source studies.}
Directly harvesting experimental data from arbitrary published wet-lab papers without screening often encounters uneven reproducibility, incomplete protocol descriptions and selective reporting of outcomes~\citep{begley2012raise,errington2021replicability}. To ensure high-fidelity ground truth, our domain experts curated source studies exclusively from recent Nature Portfolio journals, prioritizing research groups with an established track record in the relevant experimental methods and requiring each study to report complete quantitative measurements for every experiment.

\vspace{-5pt}
\paragraph{Step 2: Screening the data.}
Within each study, the experts avoided two kinds of comparison: orderings that rise or fall strictly with one varied parameter, such as a concentration series, where a model can score highly just by extending the trend; and groups whose values differ only slightly within a narrow range, where the order more likely reflects experimental error than a real difference. Domain experts then selected, on scientific grounds, the groups of experiments that could be built into tasks and assigned each task to a discipline and subfield.

\vspace{-5pt}
\paragraph{Step 3: Building and reviewing tasks.}
Each selected data set was built into a task under a written protocol, which reconstructs every experiment from the source and links it to its reported value, groups experiments with a shared endpoint into comparison groups, and describes all experiments with one neutral template, anonymized identifiers and a random order. Before being frozen, each task was reviewed against the original sources independently of its construction, and any blocking issue was fixed and reviewed again.

\vspace{-5pt}
\paragraph{Step 4: Quality and difficulty calibration.}
We ran pilot agent evaluations on the candidate tasks from Step 3 and analyzed the trajectories alongside each task's experimental data. We then selected 100 high-quality tasks as the public benchmark: one task in each of the 29 subfields, and 71 more on scientific quality, nontrivial computational demand and informative difficulty. Difficulty was judged mainly from the pilot scores of GPT-6 Astra.

\subsection{Private Tasks}
\label{sec:private}

The 16 private tasks draw on four series of real wet-lab experiments carried out in collaboration with scientific laboratories, and experts designed each task from the measured results. None of these results has been published, so none can be found online or recalled from training data. Across all four series, measurements were performed with independent experimental replicates or same-plate replicate wells to ensure the accuracy and reliability of the experimental outcomes. Together the four series contain 260 experiments in 28 comparison groups.

\vspace{-5pt}
\paragraph{Nanobody variants.}
The first series comes from an engineering campaign on VHH nanobodies that bind the SARS-CoV-2 spike protein, and it covers four sets of constructs: six original clones, a reference with fourteen single amino-acid substitutions, combinations of those substitutions, and humanized constructs, each variant set measured together with its reference. Every set was ranked once by binding signal, measured by ELISA against the WT or the BA.4/5 spike protein, and once by expression yield, which gives eight tasks with 155 experiments in 16 comparison groups. The agent receives each construct's sequence, its substitutions relative to the reference of its group, the expression host and plasmid, the antigen and the dilution series, and the shared expression and assay procedure. It orders the constructs in each group by the area under the ELISA dilution curve, or by the expression yield in mg/L.

\vspace{-5pt}
\paragraph{Reaction conditions.}
The second series comes from three condition screens: benzylation of 2-pyridone with benzyl bromide over eight conditions, sulfonylation of 3-aminophenol with \textit{p}-toluenesulfonyl chloride over ten, and reductive amination of \textit{N}-methylethylenediamine with benzaldehyde over seven. The first two screens vary the base and the solvent and the third the reducing agent and the solvent, while each screen holds its substrates, stoichiometry, temperature and reaction time fixed. The agent receives the identity, CAS number, structure and amount of every varied reagent, together with the shared procedure and the definition of the measured endpoint, and orders the conditions by that endpoint: the analytical yield of the target product in the first two screens, and an NMR yield estimate for the candidate product in the third.

\vspace{-5pt}
\paragraph{Electrolytes.}
The third series dissolves 1\,M LiPF$_6$ in single solvents, without a co-solvent or an additive, and measures each solution at 25 degrees Celsius. Ten solutions were measured for ionic conductivity and eight for the apparent electrochemical window, taken from linear sweep voltammetry as the difference between the oxidation and the reduction onset. The agent receives the name, structure, formula and CAS number of each solvent, the salt and its concentration, the amounts used and the measurement definition, and orders the solutions by the measured conductivity or window.

\vspace{-5pt}
\paragraph{Magrolimab variants.}
The fourth series measures variants of Magrolimab, a monoclonal antibody against human CD47, each prepared in CHO or 293F cells with the same vector and measured against a wild-type reference. Three tasks follow the three measurements of the campaign. The first ranks the reported yield and the reported purity of seven CHO preparations, 14 experiments in two comparison groups. The second ranks the ELISA response of ten variants at antibody-diluent pH 6.0, 6.8 and 7.4, taken over eight concentrations as the normalized trapezoidal area of the mean reported OD on the log-concentration axis, 30 experiments in three groups. The third ranks the SPR affinity of nine prepared samples against human CD47 at pH 7.4 and 6.0, taken as the mean of the two reported $K_\mathrm{D}$ values, 18 experiments in two groups. The agent receives the light- and heavy-chain variable regions of every variant and of the reference, each sample's expression system and vector, the shared preparation and assay procedures, the concentration and pH series, and the definition of each endpoint, and orders the variants within each group.

The private tasks use the same format, computing environment, submission files, information policy and scoring as the public tasks, and each passed an independent review before it was frozen. No model scores were used to decide which experiments the private tasks include.

\subsection{Evaluation}
\label{sec:evaluation}

The Science Prediction Universe grades only the submitted rankings, against measured values that the agent never sees. Grading is automatic and deterministic: no model judges a submission, and the same file always receives the same score.

\vspace{-5pt}
\paragraph{Grading environment.}
After the agent stops, the grader reads \texttt{rankings.json} in a separate container that has no network access and holds the measured values and the grading code. The report, source record and scripts are kept with the run but are not graded.

\vspace{-5pt}
\paragraph{Scoring.}
For a comparison group $g$ with $n_g$ experiments, let $r_i$ be the measured rank of experiment $i$ (ties receive average ranks) and $\hat r_i$ its submitted rank. The group score is the Spearman correlation
\begin{equation}
  \rho_g=\frac{\sum_i (r_i-\bar r)(\hat r_i-\bar{\hat r})}{\sqrt{\sum_i (r_i-\bar r)^2}\,\sqrt{\sum_i (\hat r_i-\bar{\hat r})^2}}\in[-1,1],
\end{equation}
which reduces to $1-6\sum_i (r_i-\hat r_i)^2/\big(n_g(n_g^2-1)\big)$ without ties. The task score is $S_t=\sum_g w_g\rho_g$, where the group weights $w_g$ are fixed when the task is built and sum to one; 97 of the 100 public tasks weight their groups equally. Because 33 public tasks contain tied measured values, their attainable maximum is slightly below 1 (0.956 at the lowest).

\vspace{-5pt}
\paragraph{Valid submissions.}
A valid \texttt{rankings.json} is UTF-8 JSON of at most 1\,MiB with unique keys. It states the task identity and lists every comparison group exactly once, each with an ordering that contains every experiment of the group once. For a missing file, or an error that affects the whole file, such as unparsable JSON, a wrong task identity or an unknown group, the grader returns $S_t=-1$. For an error confined to one group, such as an absent group or an incomplete ordering, it sets that group's $\rho_g$ to $-1$; the group keeps its weight and the other groups are scored normally. Because a uniformly random valid ranking has an expected score of 0, a failed file scores below chance, and a failed group lowers the task score. Section~\ref{sec:experiments} reports scores both under these penalties and with refusals and formatting errors separated from prediction accuracy (Sections~\ref{sec:exp-setup} and~\ref{sec:run-outcomes}).

\vspace{-5pt}
\paragraph{Chance level.}
To separate a high score from chance, especially on small tasks, for each task we draw 4,000 random valid submissions, each permuting every group, and take the 95th percentile of the resulting $S_t$ as the chance threshold; a score above it beats random ranking at $P<0.05$. Across the public tasks the threshold has a median of 0.44, and it lies between 0.71 and 0.80 for the 17 tasks with fewer than seven experiments. Besides the mean score, we report the fraction of tasks on which a configuration exceeds its threshold. Appendix~\ref{app:score-examples} shows what scores look like on real runs.

\vspace{-5pt}
\paragraph{Run records.}
Each run keeps its full trajectory, with every model message, tool call and tool output, including all calls to the audited search and fetch tools, as well as the submitted report and source record. These records do not change the grader's score. Section~\ref{sec:experiments} specifies how refusals, infrastructure failures and policy stops enter the reported results.

\section{Experiments}
\label{sec:experiments}

We evaluate thirteen agent configurations on the 100 public tasks and the 16 private tasks. The two task sets differ in size and origin, published literature against unpublished wet-lab campaigns, so we report them separately (Sections~\ref{sec:public-results} and~\ref{sec:private-results}).

\subsection{Experimental Setup}
\label{sec:exp-setup}

\vspace{-5pt}
\paragraph{Configurations.}
GPT-6 Astra~\citep{openai2026astra} runs in the Codex CLI (v0.156.1)~\citep{openai2026codex} at medium and at max reasoning effort, giving the configurations Astra Medium and Astra Max, and GPT-6.1 Sol~\citep{openai2026sol} runs in the same Codex version at max effort. Ten other models run in Claude Code (v2.1.260)~\citep{anthropic2025claudecode} at max reasoning effort: Claude Opus 5.5~\citep{anthropic2026opus55}, Claude Opus 5~\citep{anthropic2026opus5}, Claude Sonnet 5.5~\citep{anthropic2026sonnet55}, DeepSeek-V4.1-Flash-0910~\citep{deepseek2026flash}, Qwen3.8-Max-0902~\citep{qwen2026qwen38}, GLM-5.3-Flash~\citep{zhipu2026glm53}, Doubao-Seed-Evolving-0909~\citep{bytedance2026seed}, HY4-Preview~\citep{tencent2026hy4}, Kimi-K3.1-Pro~\citep{moonshot2026kimik31pro} and Claude Fable 5.1~\citep{anthropic2026fable51}, shortened below to Opus 5.5, Opus 5, Sonnet 5.5, DeepSeek, Qwen3.8, GLM-5.3, Doubao, HY4, Kimi-K3.1-Pro and Fable 5.1. A configuration combines a model, a reasoning effort and an agent harness, and the results describe each configuration as deployed. All configurations receive the same task instructions and run in the environment of Section~\ref{sec:task}.

\vspace{-5pt}
\paragraph{Runs.}
Each configuration is evaluated on every task, totaling 1,300 official runs on the 100 public tasks and 208 on the 16 private tasks. We also conduct five repeated runs on a stratified sample of 29 public tasks to evaluate run-to-run stability (Section~\ref{sec:repeated-runs}). Any attempt interrupted before completion by an API error, a truncated session or the memory limit was rerun once under the same configuration.

\vspace{-5pt}
\paragraph{Scoring.}
By default, the grader assigns $-1$ to any run without a valid \texttt{rankings.json} (Section~\ref{sec:evaluation}; Appendix~\ref{app:exp-recovery}, Table~\ref{tab:outcomes-grader}). To separate scientific prediction accuracy from safety refusals and formatting errors in our main analysis, a run is scored when its \texttt{rankings.json} is valid or when its group orderings can be unambiguously recovered from a formatting-rejected file (Section~\ref{sec:run-outcomes}; Appendix~\ref{app:exp-recovery}); safety refusals and runs without a recoverable submission are excluded from the mean and reported explicitly via task coverage alongside each score. Runs that stopped under the information policy are scored if they submitted a ranking.

\vspace{-5pt}
\paragraph{Statistics.}
Configurations are scored on different tasks, so we report two means. The raw mean averages a configuration's scores over its scored tasks. The task-adjusted mean fits the additive model $y_{tc}=\mu_t+\alpha_c$ to the scores $y_{tc}$ of all scored public runs, where $t$ indexes tasks and $c$ configurations, and reports $\alpha_c+\bar\mu$, with $\bar\mu$ the average task effect; when task and configuration effects add, it estimates the configuration's mean over all 100 tasks. Because Opus 5.5 and Fable 5.1 have substantially lower coverage due to safety refusals (44 and 30 scored public tasks), their means reflect the subsets of tasks they completed; we therefore explicitly mark the number of scored tasks alongside every reported mean. Confidence intervals are 95\% percentile intervals from 10,000 bootstrap resamples of tasks. Two configurations are compared on the tasks scored for both, by the mean paired difference and a two-sided Wilcoxon signed-rank test with Holm correction over the 78 pairs.

\subsection{Run Outcomes}
\label{sec:run-outcomes}

\begin{figure}[!htbp]
  \centering
  \includegraphics[width=\linewidth]{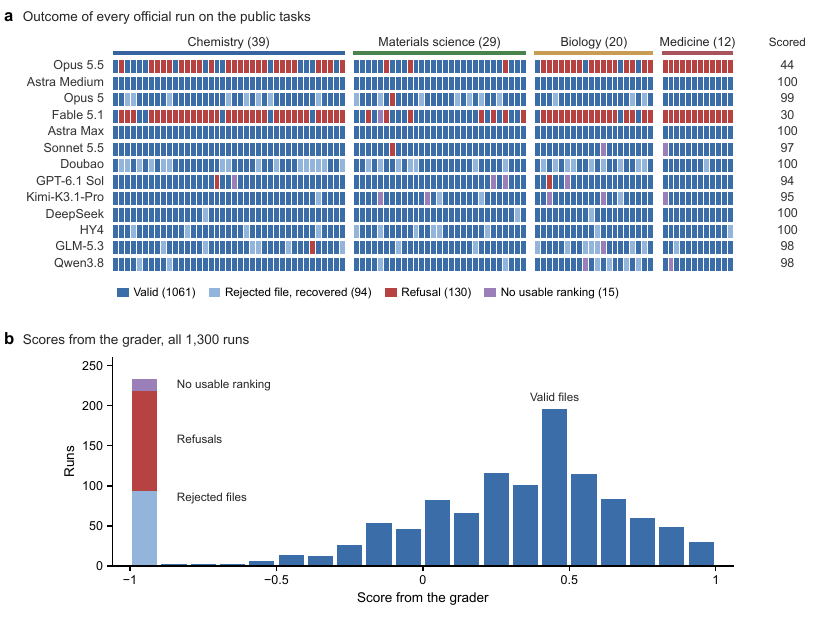}
  \caption{\textbf{Outcomes of the official runs on the public tasks before exclusion.} \textbf{a}, Outcome of each of the 1,300 official runs, with tasks grouped by discipline and ordered within each by mean score. Valid submissions and recovered files are scored, and all other runs are excluded. \textbf{b}, The grader's scores of all 1,300 runs, with every run without a valid file at $-1$.}
  \label{fig:public-outcomes-raw}
\end{figure}

Across the 1,300 official runs on the public tasks, 1,061 were classified as valid submissions and 94 left a \texttt{rankings.json} that the grader rejected for formatting errors; 130 ended in a safety refusal and 15 were excluded for other outcomes (Figure~\ref{fig:public-outcomes-raw}a), ten of them after stopping under the information policy, four without a submission and one at the memory limit. Five refusal runs and the memory-limit run nevertheless left files that the grader accepted (Appendix~\ref{app:exp-recovery}, Table~\ref{tab:outcomes-grader}). The grader therefore assigns $-1$ to the remaining 233 runs whose files it did not accept (Figure~\ref{fig:public-outcomes-raw}b).

\begin{figure}[!htbp]
  \centering
  \includegraphics[width=\linewidth]{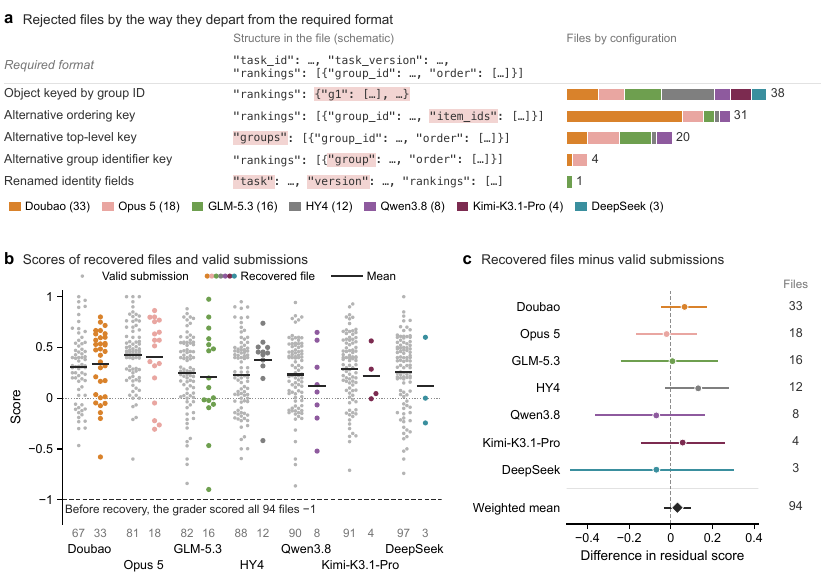}
  \caption{\textbf{Files rejected for formatting errors before and after recovery.} \textbf{a}, The 94 formatting-rejected files by how they depart from the required JSON schema, with a schematic of each type (departures shaded) and the files from each configuration. \textbf{b}, Scores of the recovered files and of the valid submissions of the same seven configurations; numbers under the columns count runs. \textbf{c}, Recovered files minus valid submissions within each configuration, with each run's score taken relative to the mean of the other configurations on the same task; the weighted mean weights each configuration by its recovered files. Lines are 95\% bootstrap intervals.}
  \label{fig:public-rejected-files}
\end{figure}

\vspace{-5pt}
\paragraph{Files rejected for formatting errors.}
The 94 files rejected by the grader for formatting errors come from seven configurations and fall into five kinds of schema departure, each of which renames or renests a JSON key while leaving the experiment ordering lists intact (Figure~\ref{fig:public-rejected-files}a): most hold the rankings as an object keyed by group where a list was required, put each ordering under a key other than \texttt{order}, or put the groups under a top-level key other than \texttt{rankings}. Every ordering in the 94 files could therefore be recovered unchanged and without ambiguity by a deterministic schema-normalization script (Appendix~\ref{app:exp-recovery}), and two independent implementations of the recovery give scores that agree to within $10^{-16}$. Recovered files do not score detectably differently from the valid submissions of the same configuration (Figure~\ref{fig:public-rejected-files}b). We compare the two within each configuration, measuring each run by its score minus the mean of the other configurations on the same task; weighted by the number of recovered files, recovered files average 0.032 higher (95\% CI $-0.033$ to 0.097; permutation $P=0.36$; Figure~\ref{fig:public-rejected-files}c).

\begin{figure}[!htbp]
  \centering
  \includegraphics[width=\linewidth]{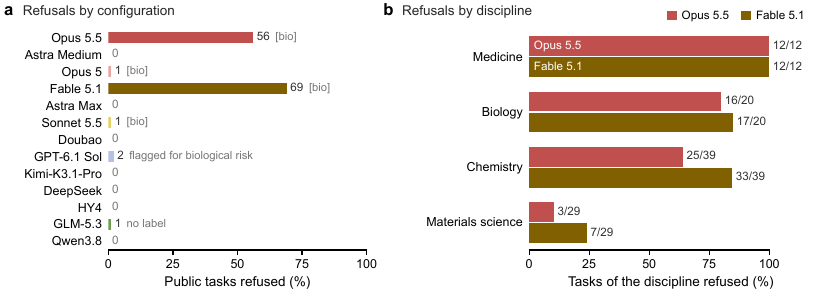}
  \caption{\textbf{Refusals in the official runs on the public tasks.} \textbf{a}, Share of the 100 public tasks refused by each configuration; every refusal by Opus 5.5, Fable 5.1, Opus 5 and Sonnet 5.5 carries the [bio] label, and GPT-6.1 Sol's two were flagged by its provider for possible biological risk. \textbf{b}, Share of the tasks in each discipline refused by Opus 5.5 and Fable 5.1.}
  \label{fig:public-refusals}
\end{figure}

\vspace{-5pt}
\paragraph{Refusals.}
Opus 5.5 refused 56\% of the 100 public tasks and Fable 5.1 69\% (Figure~\ref{fig:public-refusals}a). All 125 of these refusals came from the provider's safety classifier for biological risk and carry its [bio] label. Both models refused every medicine task and most of the biology and chemistry tasks, against a small minority of the materials science tasks (Figure~\ref{fig:public-refusals}b). The same classifier stopped Opus 5 on one task and Sonnet 5.5 on one, GPT-6.1 Sol was blocked upstream for biological risk on two, GLM-5.3 refused one task without the label, and the other seven configurations refused none.

Most runs that stopped under the information policy still submitted something, either a provisional ranking written before the stop or the order of the input files, and these submissions are scored (Appendix~\ref{app:exp-policy}). Including runs that recorded no stop, 22 official runs on 15 tasks submitted the input order. Nine runs reached the 8-hour limit after writing a file, and five ended early after writing a valid one (two of HY4 and one each of Opus 5, Kimi-K3.1-Pro and Sonnet 5.5); these 14 runs are scored as well.

\begin{figure}[!htbp]
  \centering
  \includegraphics[width=\linewidth]{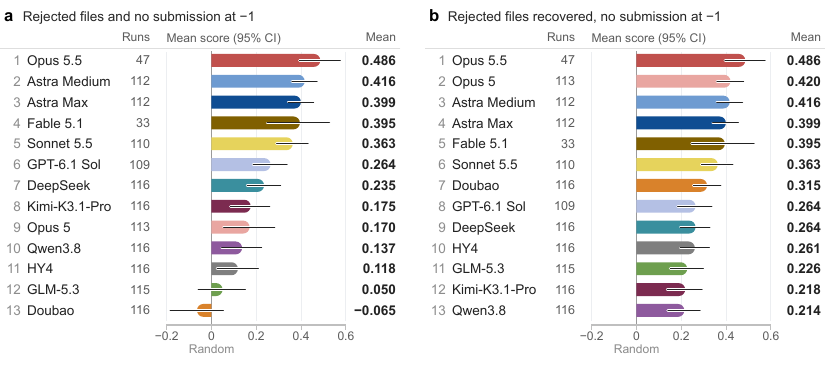}
  \caption{\textbf{Mean score of each configuration under the two scoring rules that penalize a missing or formatting-rejected file.} \textbf{a}, Formatting-rejected files and runs without a submission scored at $-1$. \textbf{b}, Formatting-rejected files scored by their recovered rankings, with runs without a submission still at $-1$. Both rules cover the public and the private tasks and exclude refusals; Runs gives the number of runs behind each mean, and lines are 95\% bootstrap intervals over tasks.}
  \label{fig:public-scoring-rules}
\end{figure}

\vspace{-5pt}
\paragraph{Scoring rule.}
Table~\ref{tab:outcomes-grader} in Appendix~\ref{app:exp-recovery} traces how each configuration's mean on the 1,300 public runs changes across scoring rules. Under the grader's default convention, which scores every run without a valid file at $-1$, the two Astra configurations lead and Opus 5.5 and Fable 5.1 come last, at $-0.329$ and $-0.537$, because most of their runs are safety refusals. The rules that follow exclude refusals and cover the public and private tasks together (1,508 official runs in all). Keeping formatting-rejected files and runs without a submission at $-1$ leaves Doubao last at $-0.065$, against $-0.126$ on its public runs alone, and Opus 5 ninth at 0.170 (Figure~\ref{fig:public-scoring-rules}a). Scoring the formatting-rejected files by their recovered rankings lifts Doubao, Opus 5, GLM-5.3 and HY4 by 0.14 to 0.38, moving Doubao to seventh and Opus 5 to second (Figure~\ref{fig:public-scoring-rules}b). Excluding the 17 runs without a submission as well gives the means shown as solid bars in Figure~\ref{fig:model-score-ranking}, where Opus 5.5 (0.486) and Fable 5.1 (0.438) rise to the top on the tasks they completed; averaged over the thirteen configurations, that mean is 0.344 over all the tasks each one completed and 0.396 over the 28 tasks that every one of them completed (hatched bars). Reporting both coverage and scored means thus disentangles operational reliability and refusal behavior from underlying scientific reasoning.

\subsection{Results on the Public Tasks}
\label{sec:public-results}

\begin{figure}[!htbp]
  \centering
  \includegraphics[width=\linewidth]{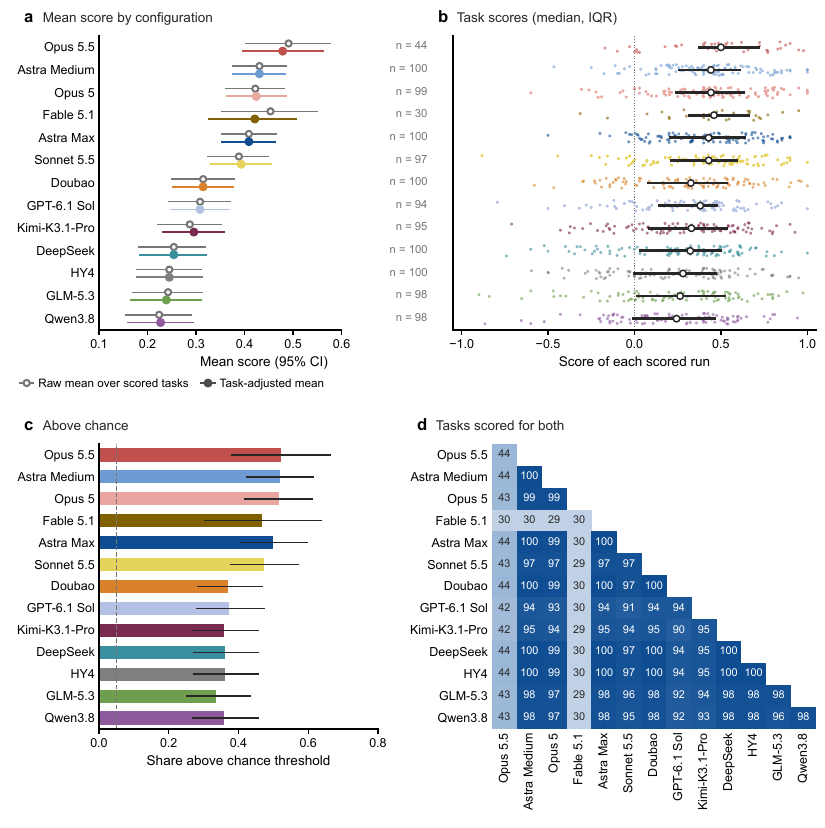}
  \caption{\textbf{Coverage and scores on the public tasks.} \textbf{a}, Raw mean (open circles) and task-adjusted mean (filled circles), with 95\% bootstrap intervals over tasks; n is the number of scored tasks. \textbf{b}, Scores of all scored runs, with medians and interquartile ranges. \textbf{c}, Share of scored tasks whose score exceeds the task's chance threshold, with Wilson 95\% intervals; the dashed line marks the 5\% expected from random rankings. \textbf{d}, Number of tasks scored for both configurations of each pair.}
  \label{fig:public-scores}
\end{figure}

\vspace{-5pt}
\paragraph{Coverage.}
Of the 1,300 official runs, 1,155 are scored (Section~\ref{sec:run-outcomes}). Safety refusals make up 130 of the 145 exclusions and all but five of them come from Opus 5.5 and Fable 5.1, which are left scored on 44 and 30 tasks, most of them in materials science. The tasks excluded for these two are not detectably harder or easier for the rest: the other eleven configurations score 0.023 lower on the tasks excluded for Opus 5.5 than on its scored tasks (95\% CI $-0.100$ to 0.055; Holm-corrected $P=0.30$) and 0.049 lower on those excluded for Fable 5.1 ($-0.130$ to 0.033; $P=0.09$). Where coverage is not missing at random, the task-adjusted mean accounts for it: on the five tasks where Kimi-K3.1-Pro left no usable file, the other configurations score 0.178 higher than on its 95 scored tasks (0.045 to 0.295; $P=0.009$). Of the other eleven configurations, five are scored on all 100 tasks and six on 94 to 99 (Figure~\ref{fig:public-scores}a, d), and all thirteen are scored on 27 of the 100 public tasks.

\vspace{-5pt}
\paragraph{Scores.}
Table~\ref{tab:public-results} and Figure~\ref{fig:public-scores}a give the means. On a scale on which a perfect ranking scores 1, the highest task-adjusted mean is Opus 5.5's 0.478, and the thirteen configurations separate into a leading six between 0.393 and 0.478 and a trailing seven between 0.227 and 0.315. The adjustment changes no mean by more than 0.032 and moves only Fable 5.1, from second on the raw means to fourth. Each configuration scores above zero on most of its scored tasks but beats the chance threshold on only 34\% to 52\% of them (Figure~\ref{fig:public-scores}c), against 5\% for random rankings. Over the scored runs, whose median score is 0.40, a median of 64\% of experiment pairs fall in the measured order, against 50\% for random rankings (Appendix~\ref{app:score-examples}). Scores vary more across tasks than across configurations: a task's mean over the configurations scored on it averages 0.330 and ranges from $-0.163$ to 0.698, and a Kruskal-Wallis test over these task means detects no difference among the four disciplines ($P=0.18$; Section~\ref{sec:analysis-discipline}).

\begin{table}[!htbp]
  \centering
  \caption{\textbf{Results on the public tasks.} Configurations are ordered by task-adjusted mean, and brackets give 95\% bootstrap intervals over tasks. Above chance counts the scored tasks whose score exceeds the task's chance threshold (Section~\ref{sec:evaluation}).}
  \label{tab:public-results}
  \small
  \begin{tabular}{lrllr}
    \toprule
    Configuration & Scored tasks & Task-adjusted mean [95\% CI] & Raw mean [95\% CI] & Above chance \\
    \midrule
    Opus 5.5 & 44 & 0.478 [0.395, 0.564] & 0.491 [0.401, 0.578] & 23 (52\%) \\
    Astra Medium & 100 & 0.431 [0.373, 0.486] & 0.431 [0.373, 0.487] & 52 (52\%) \\
    Opus 5 & 99 & 0.424 [0.361, 0.487] & 0.422 [0.359, 0.484] & 51 (52\%) \\
    Fable 5.1 & 30 & 0.421 [0.326, 0.508] & 0.453 [0.351, 0.551] & 14 (47\%) \\
    Astra Max & 100 & 0.409 [0.352, 0.466] & 0.409 [0.352, 0.466] & 50 (50\%) \\
    Sonnet 5.5 & 97 & 0.393 [0.328, 0.457] & 0.388 [0.323, 0.450] & 46 (47\%) \\
    Doubao & 100 & 0.315 [0.250, 0.377] & 0.315 [0.249, 0.379] & 37 (37\%) \\
    GPT-6.1 Sol & 94 & 0.308 [0.246, 0.371] & 0.308 [0.243, 0.371] & 35 (37\%) \\
    Kimi-K3.1-Pro & 95 & 0.296 [0.230, 0.361] & 0.287 [0.220, 0.353] & 34 (36\%) \\
    DeepSeek & 100 & 0.254 [0.183, 0.323] & 0.254 [0.181, 0.321] & 36 (36\%) \\
    HY4 & 100 & 0.245 [0.176, 0.313] & 0.245 [0.176, 0.311] & 36 (36\%) \\
    GLM-5.3 & 98 & 0.239 [0.164, 0.312] & 0.242 [0.167, 0.315] & 33 (34\%) \\
    Qwen3.8 & 98 & 0.227 [0.157, 0.296] & 0.224 [0.154, 0.291] & 35 (36\%) \\
    \bottomrule
  \end{tabular}
\end{table}

\vspace{-5pt}
\paragraph{Paired comparisons.}
To compare configurations with different coverage fairly, we evaluate all 78 configuration pairs strictly on their shared scored tasks using a two-sided Wilcoxon signed-rank test with Holm correction, identifying significant differences in 17 pairs (Figure~\ref{fig:public-paired}a, c; Appendix~\ref{app:exp-runs}). Every one of the 17 separates five of the leading six configurations (Opus 5.5, Astra Medium, Opus 5, Astra Max and Sonnet 5.5) from six of the seven below them (Doubao, Kimi-K3.1-Pro, DeepSeek, HY4, GLM-5.3 and Qwen3.8), by 0.14 to 0.29 on 43 to 100 shared tasks. GPT-6.1 Sol, the remaining configuration of the lower group, differs detectably from no configuration, and neither does Fable 5.1, which shares only 29 to 30 tasks with each other configuration, so the placement of these two rests on their point estimates. No difference is detected within either group (Appendix~\ref{app:exp-runs}).

\vspace{-5pt}
\paragraph{The tasks every configuration completed.}
On the 28 tasks that all thirteen configurations completed, 27 public and one private, the ranking is free of any coverage difference (hatched bars in Figure~\ref{fig:model-score-ranking}). So few tasks are shared because Opus 5.5 and Fable 5.1 refused most of the chemistry, biology and medicine tasks; the other eleven configurations alone share 94, and 20 of the 28 come from materials science. All but GLM-5.3 score higher on these tasks than on their own scored public tasks, by 0.003 to 0.131, and GLM-5.3 is 0.006 lower, so the level is raised throughout and only the order is comparable with Table~\ref{tab:public-results}. The leading six hold the first six places here as well, with Opus 5.5 first and Astra Max rising from fifth place to second, and the other seven hold the last seven, so the shared tasks reproduce the two groups.

\vspace{-5pt}
\paragraph{Effort.}
Longer runs did not score higher. Over the 100 tasks, Astra Medium averages 0.022 above Astra Max (95\% CI $-0.031$ to 0.074; Figure~\ref{fig:public-paired}d), although Astra Max runs a median of 4.8 times as long on the same task (Section~\ref{sec:analysis-effort}). With task and configuration effects removed, the correlation between score and log run time is 0.043 (95\% CI $-0.026$ to 0.113; Appendix~\ref{app:exp-effort}). Agents choose how long to run, so this correlation is observational, whereas the two Astra configurations differ only in the reasoning effort we set.

\subsection{Results on the Private Tasks}
\label{sec:private-results}

\vspace{-5pt}
\paragraph{Runs and coverage.}
Each configuration ran once on each of the 16 private tasks with the same settings as on the public tasks, 208 runs in total. Under the scoring rule of Section~\ref{sec:exp-setup}, 159 runs are scored: 156 valid submissions and three recovered files. The 49 excluded runs are 47 refusals, one run that reached the 8-hour limit without a submission and one that stopped under the information policy (Appendix~\ref{app:exp-private}). The refusals again fall on Opus 5.5 and Fable 5.1, which refused all but three and all but two of the 16 tasks, including every VHH and Magrolimab task, with a few more from GPT-6.1 Sol, Sonnet 5.5 and Opus 5; an upstream biological-risk block also ended both Astra configurations on the four VHH binding tasks. Coverage therefore ranges from 2 and 3 tasks for those two configurations to between 11 and 16 for the rest (Table~\ref{tab:private-results}).

\begin{figure}[!htbp]
  \centering
  \includegraphics[width=\linewidth]{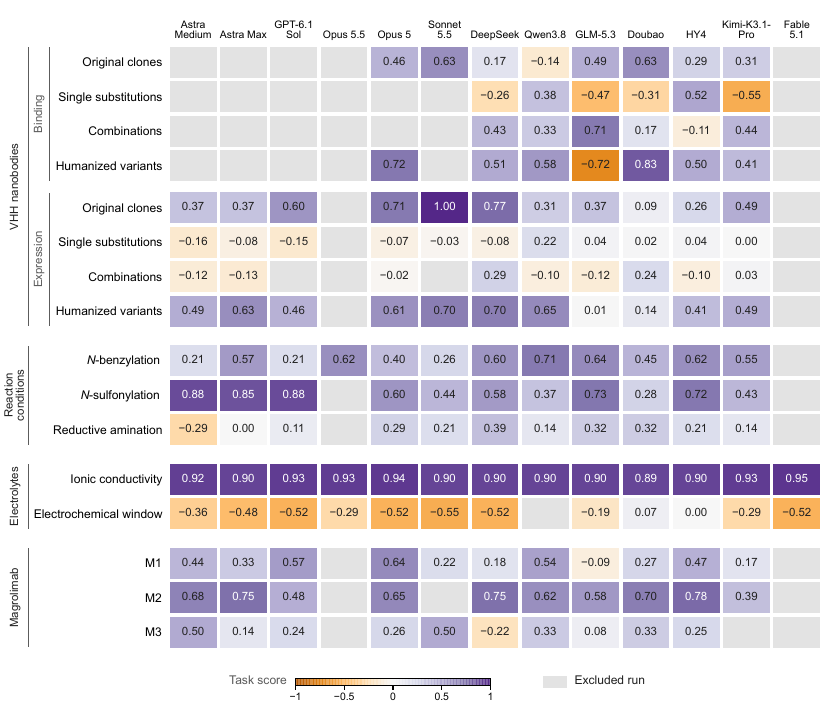}
  \caption{\textbf{Scores on the private tasks.} Score of each configuration on each private task, grouped into the VHH (eight tasks), reaction-condition (three), electrolyte (two) and Magrolimab (three) series. Gray cells mark excluded runs: refusals, one timeout and one information-policy stop.}
  \label{fig:private-scores}
\end{figure}

\vspace{-5pt}
\paragraph{Scores.}
Table~\ref{tab:private-results} and Figure~\ref{fig:private-scores} give the results. Among the eleven configurations scored on at least nine tasks, means range from 0.206 to 0.405, with 95\% intervals 0.28 to 0.49 wide; Opus 5.5 (0.420 on 3 tasks) and Fable 5.1 (0.214 on 2 tasks) have too few scored tasks for a stable mean, and Table~\ref{tab:private-results} marks their intervals, and those of GPT-6.1 Sol and Sonnet 5.5 on 11 tasks, as descriptive. Compared with uniformly random rankings of the same tasks, twelve configurations score significantly higher after Holm correction over the thirteen (every configuration except Fable 5.1).

\begin{table}[!htbp]
  \centering
  \caption{\textbf{Results on the private tasks.} Rows follow the configuration order of Section~\ref{sec:exp-setup}, and brackets give 95\% bootstrap intervals over tasks, descriptive only for the four configurations scored on fewer than 12 tasks (Opus 5.5, Fable 5.1, GPT-6.1 Sol and Sonnet 5.5). Above chance counts the scored tasks whose score exceeds the task's chance threshold, as in Table~\ref{tab:public-results}; the last column gives the Holm-corrected one-sided $P$ for a mean above that of uniformly random rankings, a different comparison.}
  \label{tab:private-results}
  \small
  \begin{tabular}{lrlrr}
    \toprule
    Configuration & Scored tasks & Mean [95\% CI] & Above chance & $P$ vs random (Holm) \\
    \midrule
    Astra Medium & 12 & 0.297 [0.056, 0.531] & 5 of 12 & 0.0020 \\
    Astra Max & 12 & 0.322 [0.083, 0.556] & 4 of 12 & 0.0013 \\
    GPT-6.1 Sol & 11 & 0.346 [0.091, 0.579] & 5 of 11 & 0.0013 \\
    Opus 5.5 & 3 & 0.420 [$-0.286$, 0.927] & 1 of 3 & 0.045 \\
    Opus 5 & 14 & 0.405 [0.191, 0.588] & 6 of 14 & 0.0003 \\
    Sonnet 5.5 & 11 & 0.391 [0.138, 0.631] & 5 of 11 & 0.0005 \\
    DeepSeek & 16 & 0.324 [0.115, 0.514] & 6 of 16 & 0.0003 \\
    Qwen3.8 & 15 & 0.392 [0.250, 0.531] & 7 of 15 & 0.0003 \\
    GLM-5.3 & 16 & 0.206 [$-0.016$, 0.424] & 5 of 16 & 0.0086 \\
    Doubao & 16 & 0.320 [0.171, 0.471] & 4 of 16 & 0.0004 \\
    HY4 & 16 & 0.360 [0.216, 0.506] & 6 of 16 & 0.0003 \\
    Kimi-K3.1-Pro & 15 & 0.264 [0.079, 0.434] & 5 of 15 & 0.0019 \\
    Fable 5.1 & 2 & 0.214 [$-0.524$, 0.952] & 1 of 2 & 0.203 \\
    \bottomrule
  \end{tabular}
\end{table}

Scores depend strongly on the task (Figure~\ref{fig:private-scores}; Appendix~\ref{app:exp-private}). On ionic conductivity, all thirteen configurations score between 0.89 and 0.95, whereas on the electrochemical window of the same electrolytes the twelve scored configurations range from $-0.55$ to 0.07.

No pair of configurations differs detectably on the 16 private tasks: pairs share 1 to 16 tasks, and the smallest uncorrected $P$ of an exact sign-flip test is 0.06. Together, these four unpublished experimental series confirm that twelve of the thirteen configurations rank prospective wet-lab outcomes significantly better than chance in a setting completely free of literature retrieval and pretraining contamination, while highlighting that prediction accuracy depends strongly on the physical endpoint.

\subsection{Cost Analysis}
\label{sec:cost}

\begin{figure}[!htbp]
  \centering
  \includegraphics[width=\linewidth]{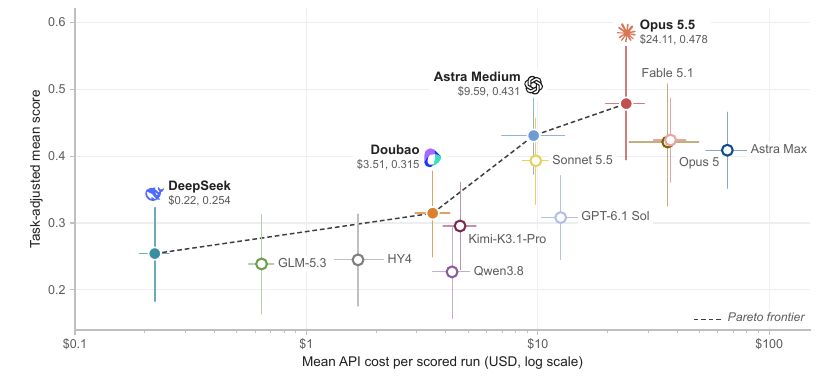}
  \caption{\textbf{Cost--performance Pareto efficiency curve on the public tasks.} Each point represents one configuration, plotting estimated mean API cost per scored run (Appendix~\ref{app:exp-cost}) against task-adjusted mean score, with 95\% bootstrap intervals. Filled markers connected by the dashed line define the cost--efficiency Pareto frontier. Opus 5.5 and Fable 5.1 are evaluated on their 44 and 30 scored tasks.}
  \label{fig:cost-pareto}
\end{figure}

\vspace{-5pt}
\paragraph{Estimation.}
We estimate the API cost of each official public run from the token usage recorded in its trajectory and the providers' list prices (Table~\ref{tab:prices}; Appendix~\ref{app:exp-cost}). The trajectories record which prompt tokens were read from or written to the provider's prompt cache, and these tokens are billed at the cache prices; in every configuration, cache reads make up 80\% to 99\% of the prompt tokens. The estimate covers model API calls only (Appendix~\ref{app:exp-cost}). Table~\ref{tab:cost} gives usage and cost per scored run, and Figure~\ref{fig:cost-pareto} plots cost against the task-adjusted mean of Table~\ref{tab:public-results}.

\begin{table}[!htbp]
  \centering
  \caption{\textbf{Estimated API cost on the public tasks.} Rows follow Table~\ref{tab:public-results}. Token counts, run time and cost are means over the scored official runs of each configuration, and the last column sums the cost of all 100 official runs. Input counts all prompt tokens, Cache reads is the share of input read from the prompt cache, and Output includes reasoning tokens. Costs use the list prices of Table~\ref{tab:prices}, with cache reads and writes billed at their own prices; brackets give 95\% bootstrap intervals over runs.}
  \label{tab:cost}
  \small
  \setlength{\tabcolsep}{4pt}
  \begin{tabular}{lrrrrrlr}
    \toprule
    Configuration & \makecell[r]{Scored\\tasks} & \makecell[r]{Input\\(M tokens)} & \makecell[r]{Cache\\reads} & \makecell[r]{Output\\(M tokens)} & \makecell[r]{Run time\\(h)} & \makecell[l]{Cost per scored run\\(USD) [95\% CI]} & \makecell[r]{Total, 100\\runs (USD)} \\
    \midrule
    Opus 5.5 & 44 & 25.6 & 93\% & 0.49 & 2.74 & 24.11 [19.67, 28.85] & 1,075 \\
    Astra Medium & 100 & 5.1 & 96\% & 0.05 & 0.46 & 9.59 [6.99, 13.05] & 959 \\
    Opus 5 & 99 & 35.4 & 96\% & 0.46 & 3.81 & 37.44 [31.59, 43.61] & 3,717 \\
    Fable 5.1 & 30 & 10.8 & 80\% & 0.15 & 1.55 & 36.42 [24.81, 49.60] & 1,158 \\
    Astra Max & 100 & 30.4 & 95\% & 0.34 & 2.90 & 65.91 [53.22, 79.84] & 6,591 \\
    Sonnet 5.5 & 97 & 20.7 & 94\% & 0.49 & 2.63 & 9.80 [8.57, 11.09] & 954 \\
    Doubao & 100 & 14.8 & 97\% & 0.13 & 2.42 & 3.51 [2.95, 4.16] & 351 \\
    GPT-6.1 Sol & 94 & 34.0 & 93\% & 0.31 & 4.19 & 12.55 [10.37, 14.81] & 1,216 \\
    Kimi-K3.1-Pro & 95 & 15.9 & 96\% & 0.15 & 1.92 & 4.61 [3.89, 5.41] & 440 \\
    DeepSeek & 100 & 13.0 & 99\% & 0.25 & 1.02 & 0.22 [0.19, 0.25] & 22 \\
    HY4 & 100 & 15.1 & 96\% & 0.21 & 2.29 & 1.67 [1.32, 2.15] & 167 \\
    GLM-5.3 & 98 & 15.4 & 97\% & 0.23 & 2.41 & 0.64 [0.56, 0.72] & 63 \\
    Qwen3.8 & 98 & 10.7 & 94\% & 0.17 & 2.15 & 4.26 [3.51, 5.08] & 420 \\
    \bottomrule
  \end{tabular}
\end{table}

\vspace{-5pt}
\paragraph{Cost range.}
The mean cost of a scored run covers a 300-fold range, from \$0.22 to \$65.91, and the 1,300 official runs cost about \$17,100 in total, 38\% of it for Astra Max (Table~\ref{tab:cost}). Prices, rather than how much the agents read, account for most of that range: prompt tokens per scored run vary only sevenfold, the cheapest configuration reads 2.6 times as many of them as Astra Medium at about 2\% of the cost, and GPT-6.1 Sol reads slightly more per scored run than Astra Max at a fifth of the price. Fable 5.1 reads fewer prompt tokens than Opus 5.5 yet costs more, because its input, output and cache-write prices are 2.5 times as high and less of its input comes from the cache, which leaves cache writes at 74\% of its cost. The two Astra configurations differ only in reasoning effort: Astra Max reads six times as many prompt tokens per scored run as Astra Medium and its runs last six times as long on average, costs a median of 5.3 times as much on the same task, and scores no higher (Section~\ref{sec:public-results}).

Four configurations define the cost--performance Pareto efficiency curve (Figure~\ref{fig:cost-pareto}): DeepSeek, Doubao, Astra Medium and Opus 5.5. Opus 5.5 reaches a higher mean score than Fable 5.1, Opus 5 and Astra Max at a lower cost, and Astra Medium scores 0.010 above Fable 5.1 at roughly one-quarter of the expense. Below the top tier, GLM-5.3, HY4 and Qwen3.8 spend 3 to 19 times as much as DeepSeek and score no higher, by margins the paired tests do not resolve (Section~\ref{sec:public-results}), and GPT-6.1 Sol costs 57 times as much as DeepSeek for a gain of 0.054; notably, within the five cheapest configurations, the highest-spending configuration (Qwen3.8) yields the lowest task-adjusted mean of the set. The frontier gains only Sonnet 5.5, which then costs less than Astra Medium, when every prompt token is billed at the full input price~\citep{li2026longhorizonterminal}, which multiplies the mean cost per scored run by 3.2 (Fable 5.1) to 9.5 (DeepSeek), and when the means are restricted to the 27 public tasks scored for all thirteen configurations, where they range from \$0.22 to \$48.44; it is unchanged when the cost of the excluded runs is charged to the scored runs of the same configuration, which raises the mean by at most 6\%, for Fable 5.1 to \$38.60.

\subsection{Repeated Runs}
\label{sec:repeated-runs}

\vspace{-5pt}
\paragraph{Design.}
To evaluate the robustness of \MYBENCH evaluations and quantify run-to-run stability across independent trials, we conducted five additional repeated runs of Astra Max on a stratified sample of 29 public tasks, drawing one representative task with a fixed seed from each discipline--subfield group (10 in chemistry, 9 in materials science, 7 in biology and 3 in medicine; 145 runs in total).

\begin{figure}[!htbp]
  \centering
  \includegraphics[width=\linewidth]{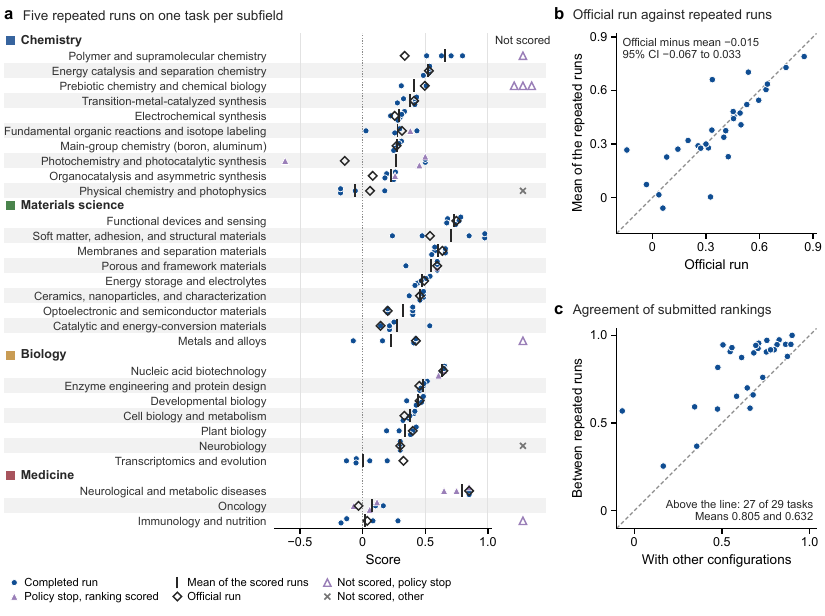}
  \caption{\textbf{Five repeated runs of Astra Max on 29 public tasks.} One task was drawn at random from each subfield, and Astra Max ran five more times on each. \textbf{a}, Scores of the repeated runs and of the official run on each task. \textbf{b}, Official score against the mean of the scored repeated runs. \textbf{c}, Agreement between two repeated runs against agreement between a repeated run and the official runs of the other twelve configurations. Dashed lines mark equality.}
  \label{fig:public-repeated-runs}
\end{figure}

Under the rule of Section~\ref{sec:exp-setup}, 137 of the 145 runs are scored (Figure~\ref{fig:public-repeated-runs}a). Twenty runs (13.8\%) stopped under the information policy; 14 of them still submitted a ranking and are scored, and 6 left no usable ranking. The other two excluded runs ended without a submission. Stops cluster on a few tasks: four tasks stopped in three of their five runs, whereas the pooled stop rate predicts 0.61 tasks with three or more stops (permutation $P=0.0019$), and no task stopped more than three times.

\vspace{-5pt}
\paragraph{Score variation.}
Scores differ more between tasks than between runs of the same task (Figure~\ref{fig:public-repeated-runs}a). The pooled within-task standard deviation is 0.150 (95\% CI 0.090 to 0.209; Figure~\ref{fig:public-runs}a), and the task means have a standard deviation of 0.219. The median within-task standard deviation is 0.089, and on one task all four scored runs submitted the same ranking. Most of the run-to-run variation comes from a few tasks: the four tasks with a standard deviation above 0.2 account for 71\% of the within-task sum of squares.

\subsection{Guarding Against Shortcuts}
\label{sec:audit}

\begin{figure}[!htbp]
  \centering
  \includegraphics[width=\linewidth]{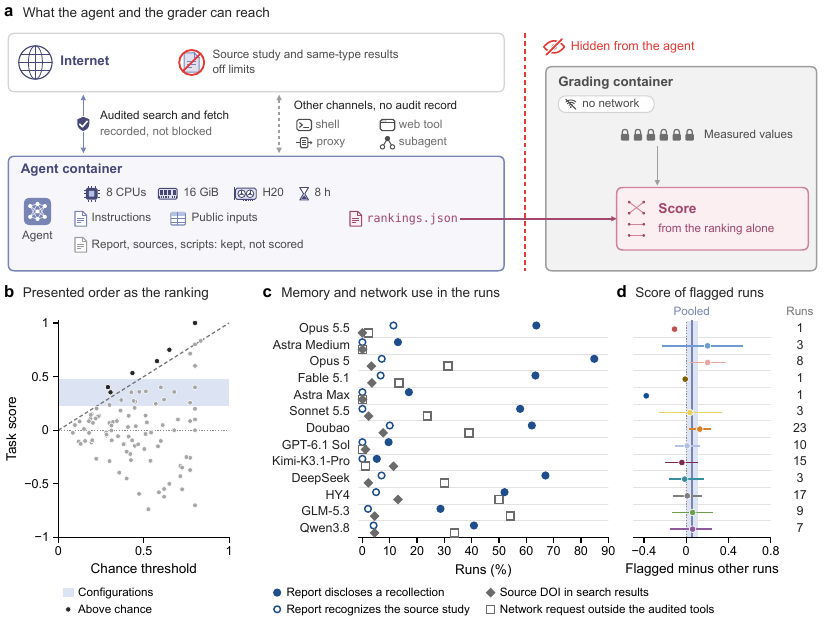}
  \caption{\textbf{Guarding against shortcuts.} \textbf{a}, What the agent can reach, what the audited service records and what stays in the grading container. \textbf{b}, Score of the order in which the experiments are presented on each public task, against the task's chance threshold, with the range of the configurations' task-adjusted means. \textbf{c}, Indicators of memory and network use, as percentages of each configuration's scored runs (for the source DOI, of the scored runs whose task has one). \textbf{d}, Difference in score between the runs flagged by the identifier detector and the unflagged runs of the same configuration, both measured against the other configurations on the same task, with 95\% intervals; the band shows the pooled difference.}
  \label{fig:exp-integrity}
\end{figure}

The automatic grading of Section~\ref{sec:evaluation} reads only the submitted order, in a container that the agent never reaches, so nothing a run writes about its own work can move its score (Figure~\ref{fig:exp-integrity}a). Nor does the task leave a trace of the measured order for an agent to copy. Experiments carry anonymized identifiers, one neutral template describes all of them, their order within a group is randomized, and the agent is not told which study the task comes from (Sections~\ref{sec:task} and~\ref{sec:construction}). Submitting the experiments in the order in which they are presented scores a mean of 0.030 over the 100 public tasks, below every configuration, and it beats the task's chance threshold about as often as a random ranking would (Figure~\ref{fig:exp-integrity}b; Appendix~\ref{app:exp-policy}). Construction also set aside comparisons that extending a trend would rank well and comparisons whose values differ within noise (Section~\ref{sec:construction}).

\vspace{-5pt}
\paragraph{The information policy.}
Every public task comes from a published study, so the instruction forbids the use of that study, its supplementary materials and data, and of any result of the target or same-type experiments, whether found online or reconstructed from memory; predictors fitted to such results are forbidden as well. All searches and downloads must go through audited tools. If a permitted query exposes a prohibited result, the agent must stop and record the step that triggered it, and it must disclose any recollection of results that it notices. The environment stays connected to the internet, the audited tools record requests rather than block them, and compliance is checked afterwards in the full trajectory that every run keeps (Section~\ref{sec:evaluation}).

\vspace{-5pt}
\paragraph{What the records show.}
A detector that scans every tool output for identifiers of the source study flags 110 of the 1,300 official runs, which shows that a tool returned an identifier and not that the run used it; no run requested a source DOI itself. Across the scored runs, 42\% of reports disclose a recollection of results as the policy requires, 42 runs on 34 tasks identify the source study once the disclosures are read by hand, the source DOI appears in the output of an audited tool in 46 of the 1,066 runs whose task has one, and almost a quarter of runs send at least one request outside the audited tools, most often to download software; Figure~\ref{fig:exp-integrity}c gives each rate by configuration. The audited service never refused a request: no policy block appears among the 8,780 error messages of its 38,152 audited calls.

\vspace{-5pt}
\paragraph{Scores of the flagged runs.}
In no configuration do the runs flagged for encountering identifiers of the source study score significantly higher than the unflagged runs after multiple-testing correction. Among the flagged runs, 41\% exceed the chance threshold of Section~\ref{sec:evaluation}, compared with 42\% among unflagged runs on the same tasks. Measured relative to the other configurations on the same task, flagged runs average +0.053 above unflagged runs of the same configuration (95\% CI 0.001 to 0.112; permutation $P=0.094$; Figure~\ref{fig:exp-integrity}d). Moreover, all eight flagged runs scoring above 0.9 occur on small tasks of at most 14 experiments, whose chance thresholds already lie between 0.62 and 0.83. Disclosing a recollection is not detectably associated with score either (Section~\ref{sec:analysis-score}). These comparisons concern exposure that leaves an identifier in a tool output; the pooled difference of +0.053 is small next to the gaps of 0.14 to 0.29 that separate the two groups of configurations (Section~\ref{sec:public-results}), and the 16 unpublished private tasks give a check that depends on no detector (Section~\ref{sec:private-results}).

\vspace{-5pt}
\paragraph{What a flag changes.}
A flag does not change the score. The grader never sees a flag, no run was excluded, rerun or rescored because of an exposure, and a run that stopped under the policy is scored if it submitted a ranking (Section~\ref{sec:exp-setup}). Of the 46 runs that saw a DOI, 40 recorded it in their report, and every trajectory is kept. The 16 private tasks give a comparison that does not depend on these records: their results have never been published, so they cannot be retrieved or recalled (Section~\ref{sec:private}), and twelve of the thirteen configurations still score above random rankings on them (Section~\ref{sec:private-results}).

\section{Analysis}
\label{sec:analysis}

Section~\ref{sec:experiments} measured how well the configurations rank experiments. This section uses their trajectories to examine how the rankings are produced, and their scores to ask which tasks separate the configurations. We analyze the 1,155 scored official runs on the public tasks, which contain 227,103 tool calls, and, where noted, the 137 scored repeated runs of Astra Max (Section~\ref{sec:repeated-runs}). Rules label each tool call with an activity and with the method families of the code it ran, and fixed detectors read the submitted reports; Appendix~\ref{app:exp-trajectory} gives the rules, and a hand-checked sample of 60 calls found 56 of the activity labels correct.

\subsection{How Agents Spend a Run}
\label{sec:analysis-run}

\begin{figure}[!htbp]
  \centering
  \includegraphics[width=\linewidth]{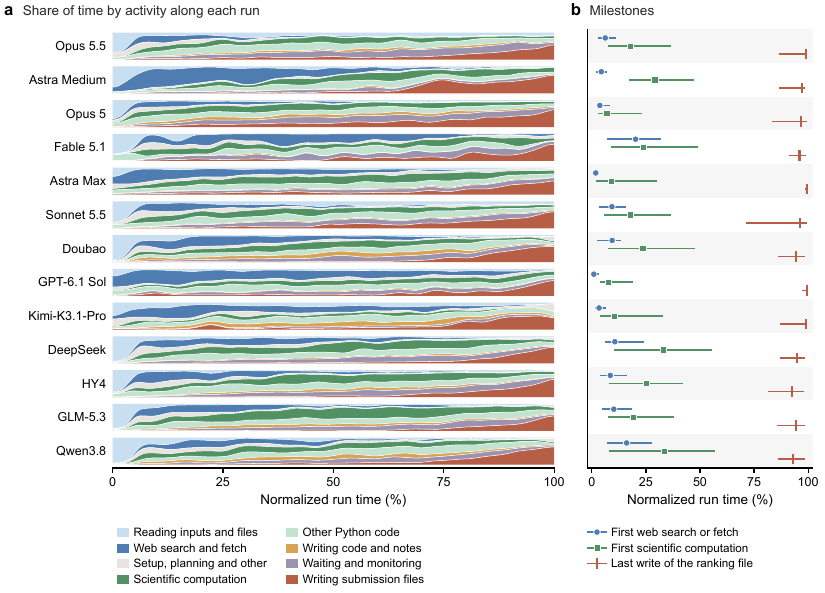}
  \caption{\textbf{How agents spend a run.} \textbf{a}, Share of time spent on each activity along a run, for the 1,155 scored official runs on the public tasks; each run is normalized to its own length and counts equally, the shares are smoothed with a Gaussian kernel of s.d.\ 2\% of the run, and the activities are defined in Appendix~\ref{app:exp-trajectory}. \textbf{b}, Median and interquartile range of the point in the run at which each milestone is reached.}
  \label{fig:analysis-workflow}
\end{figure}

A run has a consistent shape (Figure~\ref{fig:analysis-workflow}a): reading the task inputs fills most of the first 5\% of run time, web search and fetch peak within the first quarter, scientific computation leads in the middle, and writing the submission files takes over in the last tenth. Over all runs, other Python code and scientific computation take the largest shares of the time, just ahead of web search and fetch. Where the time goes differs by configuration: Astra Medium and GPT-6.1 Sol spend 30\% of a run on the web against 6\% for Opus 5.5, the first web access comes at a median of 1\% to 20\% of run time (Figure~\ref{fig:analysis-workflow}b), and 17\% of DeepSeek runs and 12\% of both Sonnet 5.5 and Qwen3.8 runs never call the audited search or fetch tools.

\vspace{-5pt}
\paragraph{Two rhythms.}
The configurations differ most in when they first commit to a complete ranking, the earliest ordering in the trajectory that has the submission format, covers every group and item and differs from the input order (Appendix~\ref{app:exp-trajectory}; Figure~\ref{fig:analysis-rankings}b). Opus 5.5, GPT-6.1 Sol, Opus 5, Fable 5.1, Sonnet 5.5, Astra Max and Kimi-K3.1-Pro write one at a median of 10\% to 32\% of run time and then keep working on it, continuing for a median of 1.7 hours after the first draft in Opus 5.5 and 3.7 hours in GPT-6.1 Sol. DeepSeek, Qwen3.8, GLM-5.3, Doubao and Astra Medium write their first complete ranking at 84\% to 87\% of run time, close to the end; Doubao writes its ranking file once (median) and revised its first draft in only one of the 54 runs where a draft is visible. HY4 falls between the two groups, at 63\%.

\begin{figure}[!htbp]
  \centering
  \includegraphics[width=\linewidth]{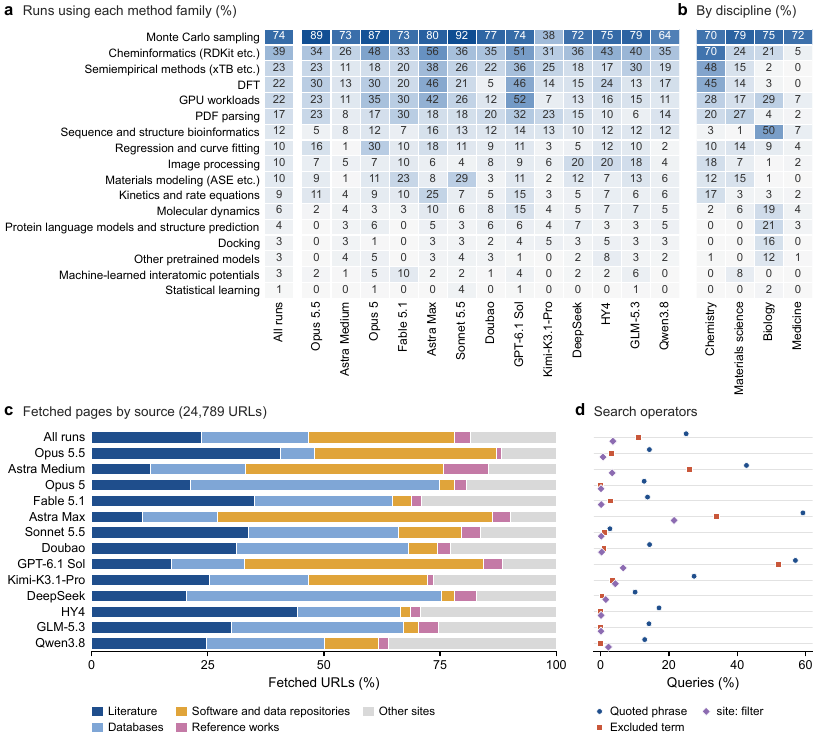}
  \caption{\textbf{Methods and information sources.} \textbf{a}, Share of the 1,155 scored official runs that used each method family, over all runs and for each configuration; the families are defined in Appendix~\ref{app:exp-trajectory}. \textbf{b}, The same share by discipline. \textbf{c}, Source class of the 24,789 URLs fetched through the audited fetch tool. \textbf{d}, Share of a run's search queries that use each operator, averaged over the 1,002 runs that searched.}
  \label{fig:analysis-methods}
\end{figure}

\vspace{-5pt}
\paragraph{Computational methods and domain specialization.}
Monte Carlo sampling~\citep{metropolis1949monte} is the most common computational tool, used in nearly three-quarters of all 1,155 scored runs and in a majority of the runs of every configuration except Kimi-K3.1-Pro (38\%; Figure~\ref{fig:analysis-methods}a), mostly to quantify the uncertainty of a ranking rather than to model an experiment. The domain-specific tools, which two runs in five invoke at least once, follow the discipline of the task (Figure~\ref{fig:analysis-methods}b): cheminformatics, semiempirical quantum chemistry and density functional theory concentrate in chemistry, sequence and structure bioinformatics in biology, and medicine tasks rarely invoke a physical simulation at all. Consequently, tool selection is driven primarily by the scientific problem rather than the model: two repeated Astra Max runs on the same task share a mean Jaccard similarity of 0.65 across method families, compared with 0.29 across different tasks. What separates the configurations is breadth, from a median of four method families per run in Astra Max to one in Astra Medium.

\vspace{-5pt}
\paragraph{Information sources.}
The mix of fetched sources varies by configuration (Figure~\ref{fig:analysis-methods}c). The three configurations under Codex also write their queries differently (Figure~\ref{fig:analysis-methods}d): Astra Medium excludes terms in about a quarter of its queries and Astra Max and GPT-6.1 Sol in more than a third, while in each of the ten Claude Code configurations fewer than 4\% of a run's queries do. The terms that the Astra runs exclude most often name a disease, material or reaction class from the task, which keeps pages about the application under study out of the results.

\subsection{How Rankings Are Produced}
\label{sec:analysis-rankings}

\begin{figure}[!htbp]
  \centering
  \includegraphics[width=\linewidth]{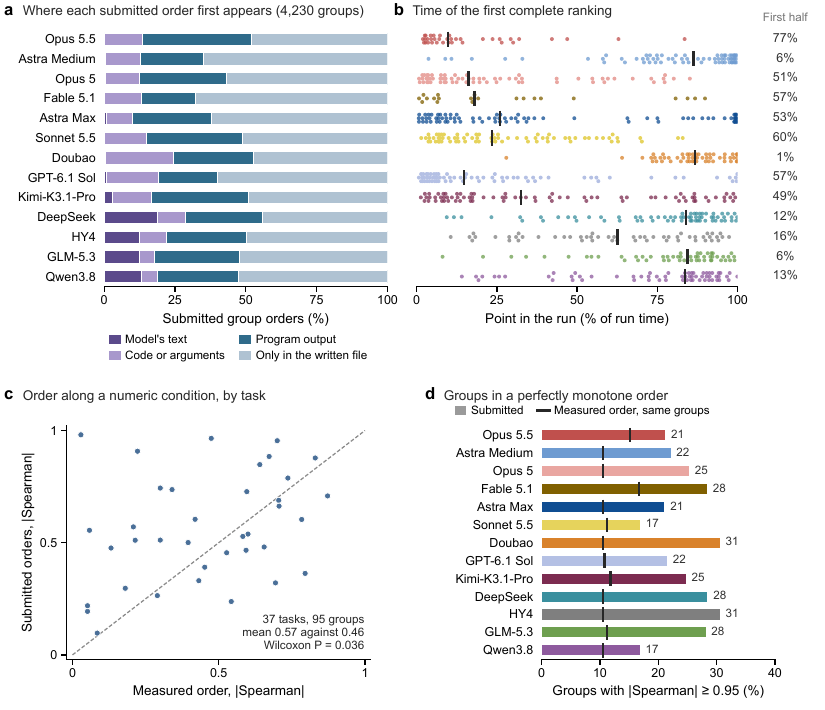}
  \caption{\textbf{How the submitted rankings are produced.} \textbf{a}, Where the submitted order of each of 4,230 groups first appears in the trajectory. \textbf{b}, Point in the run at which the first complete ranking was written, with medians (bars) and, on the right, the share of all runs that wrote it in the first half. \textbf{c}, On 37 tasks, the absolute Spearman correlation between a numeric experimental condition and the submitted orders against that for the measured order; the dashed line marks equality. \textbf{d}, Share of groups whose submitted order is monotone in the condition ($|\rho|\ge 0.95$), with ticks for the measured order.}
  \label{fig:analysis-rankings}
\end{figure}

For each submitted group of at least three items whose order differs from the input order, we located where the complete order first appears in the trajectory (Figure~\ref{fig:analysis-rankings}a). Of the 4,230 such groups, 54\% appear only in the submitted file, which a program wrote; 28\% first appear in the output of a program, 12\% in the code or arguments of a tool call and 5\% in the model's own text, a lower bound (Appendix~\ref{app:exp-trajectory}). The order of up to 82\% of the groups therefore first takes shape in a program rather than in the model's own words. The programs often work from values the model supplied itself: in 22\% of runs the code contains a table of numbers that the model typed in.

\vspace{-5pt}
\paragraph{A shared recipe.}
Most runs follow the same five steps, and the configurations differ mainly in the effort each step receives. The agent parses the items, groups and protocol; searches for background on the system and for software documentation; builds a proxy for each item, such as an RDKit descriptor, an xTB or DFT energy, a docking score or a value it assigns from its own knowledge; combines the proxies with weights it chooses, often sampling the weights or parameters by Monte Carlo and ranking by expected position; and writes the ranking with a report on its uncertainty. Two judgments in this recipe are never tested against a measurement: whether a proxy governs the measured endpoint, and how much weight each proxy deserves.

\vspace{-5pt}
\paragraph{The monotone prior.}
When the items of a group differ in a numerical condition, such as temperature, concentration or composition, agents tend to rank them along it. We examined the 95 groups of at least four items, on 37 tasks, in which such a condition takes at least three values (Figure~\ref{fig:analysis-rankings}c). Submitted orders follow the condition more closely than measured ones: their absolute Spearman correlation with it averages 0.57, against 0.46 for the measured orders (Wilcoxon $P=0.036$ over tasks). A quarter of the submitted groups are monotone in the condition, with an absolute correlation of at least 0.95, against 11\% of the measured orders of the same groups, and every one of the thirteen configurations exceeds the measured share (Figure~\ref{fig:analysis-rankings}d). Where the measured order is nearly unrelated to the condition, 30\% of the submitted groups still follow it with a correlation of at least 0.9. Because benchmark construction excluded trivially monotonic series (Section~\ref{sec:construction}), these tasks expose how heavily agents fall back on simple monotonic heuristics when the underlying mechanism is non-monotonic. Priors this uniform leave the configurations agreeing with one another more than with the measurements, and averaging their rankings scores above the average submission (Appendix~\ref{app:exp-agreement}).

\subsection{What Relates to Score}
\label{sec:analysis-score}

\begin{figure}[!htbp]
  \centering
  \includegraphics[width=\linewidth]{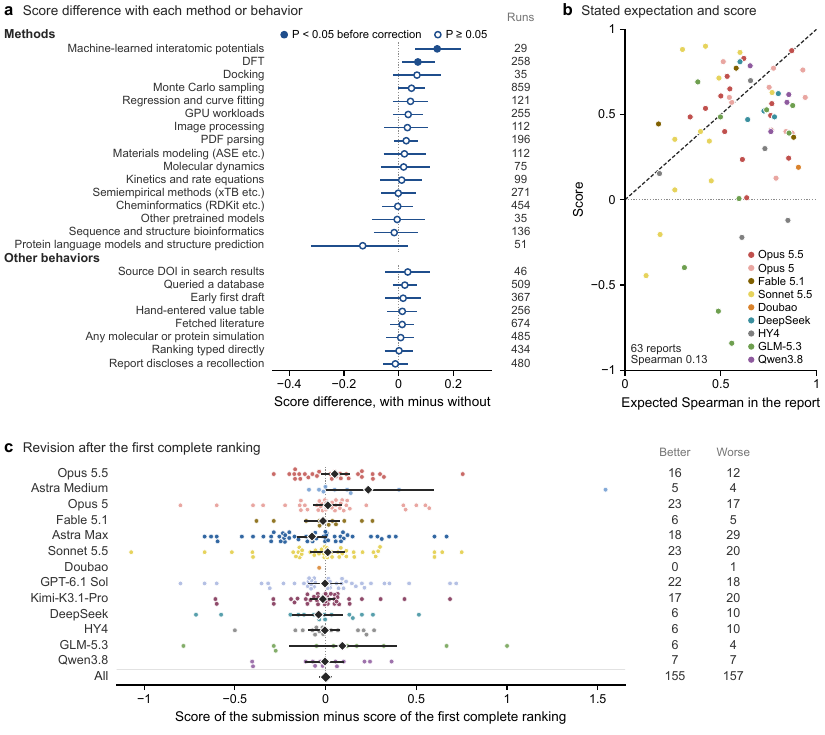}
  \caption{\textbf{What relates to score.} \textbf{a}, Score difference between runs with and without each method or behavior, with configuration and task fixed effects (Appendix~\ref{app:exp-trajectory}); filled markers have $P<0.05$ before correction, and none remains below 0.05 after Holm correction within each group. \textbf{b}, Expected Spearman correlation stated in 63 reports against the score of the run; the dashed line marks equality. \textbf{c}, Score of the submission minus that of the first complete ranking, for the 317 runs whose submission differs from it; diamonds are means. Lines are 95\% intervals from resampling tasks.}
  \label{fig:analysis-score}
\end{figure}

No method family or behavior is detectably associated with score after Holm correction (Figure~\ref{fig:analysis-score}a). Each estimate compares runs on the same task after adjusting for configuration (Appendix~\ref{app:exp-trajectory}); agents decide for themselves what to do in a run, so these associations are observational. For running any simulation, fetching literature, typing a table of values, writing an early draft and disclosing a recollection, the estimates lie within 0.02 of zero. On a single task the quantity a run ranked by can still decide its score (Section~\ref{sec:analysis-separation}).

\vspace{-5pt}
\paragraph{Revision.}
Changing a complete ranking helped about as often as it hurt. Of the 772 runs whose trajectory contains a complete ranking before the submission, 317 submitted a different one, and against their own first draft these revisions gained as much as they lost (mean change +0.001, 95\% CI $-0.034$ to 0.036; sign test $P=0.95$). The per-configuration means straddle zero as well (Figure~\ref{fig:analysis-score}c): Astra Max, which revised in the most runs, lost score on balance, and Opus 5.5 gained.

\vspace{-5pt}
\paragraph{Self-assessment.}
Reports that state an expected correlation overestimate it. Sixty-three reports across 40 tasks (all from configurations under Claude Code) explicitly state an expected Spearman correlation for the final submission. Their self-estimated median is 0.61 against an achieved median of 0.49, an average overestimation of 0.20 (95\% CI 0.11 to 0.29; Wilcoxon $P = 3.1 \times 10^{-4}$), and most of the 63 overpredict. The stated expectations also carry almost no information about the score (Spearman $r = 0.13$; Figure~\ref{fig:analysis-score}b). Consistent with the runs examined in Section~\ref{sec:analysis-review}, the reports perturb the internal weights of the chosen proxy and leave untested whether the proxy itself governs the experimental endpoint, which would make a stated expectation a measure of internal consistency rather than of accuracy.

\subsection{Tasks That Separate the Configurations}
\label{sec:analysis-separation}

\begin{figure}[!htbp]
  \centering
  \includegraphics[width=\linewidth]{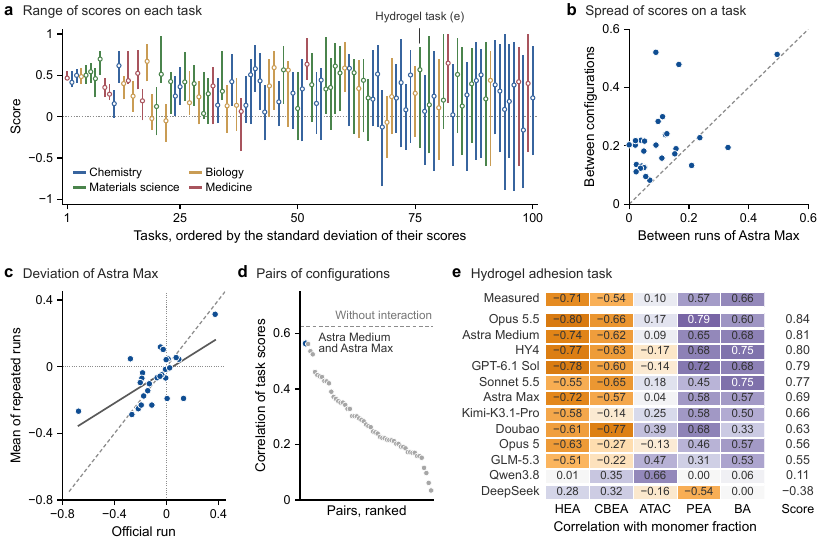}
  \caption{\textbf{Tasks that separate the configurations.} \textbf{a}, Lowest and highest score among the configurations on each public task, with the mean, ordered by the standard deviation of the scores. \textbf{b}, Spread between configurations on a task against the spread between the repeated runs of Astra Max on the same task. \textbf{c}, Astra Max's official score minus the mean of the other configurations, against the same difference computed from its repeated runs. \textbf{d}, Correlation between the per-task scores of two configurations, for the 55 pairs among the eleven configurations with the widest coverage; the dashed line marks the value expected without a configuration-by-task interaction. \textbf{e}, On the hydrogel task, the correlation between each submitted order and five monomer fractions, with the measured order at the top. Dashed lines in \textbf{b} and \textbf{c} mark equality, and the solid line in \textbf{c} is a least-squares fit.}
  \label{fig:analysis-separation}
\end{figure}

Configurations diverge sharply on specific tasks rather than maintaining a uniform gap across the benchmark. On 46 of the 100 public tasks, at least one configuration achieves a strong positive correlation ($>0.5$) while another falls below random guessing ($<0$), and the cross-configuration standard deviation has a median of 0.242 and varies by more than a factor of twenty across tasks (Figure~\ref{fig:analysis-separation}a; Appendix~\ref{app:exp-separation}). As expected, tasks with wider score dispersion exhibit lower ranking consensus among configurations (Spearman $-0.56$ with the inter-configuration agreement of Appendix~\ref{app:exp-agreement}).

\vspace{-5pt}
\paragraph{Systematic separation beyond run-to-run noise.}
Comparing cross-configuration differences against the repeated runs of Astra Max (Section~\ref{sec:repeated-runs}) confirms that these task-level gaps reflect reproducible capability differences rather than stochastic run-to-run noise. Across the 28 repeated-run tasks with at least three scored repeats, the spread between configurations exceeds the spread between repeats of one configuration on 25 tasks, by a median ratio of 6.4 (Figure~\ref{fig:analysis-separation}b), and a variance decomposition over the nine configurations scored on the same 86 tasks attributes 49\% of single-run score variance to configuration-by-task interaction versus only 20\% to run-to-run noise (Appendix~\ref{app:exp-runs}). Consequently, whether a configuration outperforms or underperforms its peers on a given task is highly repeatable: Astra Max's relative advantage in its official run and in its five repeated runs shares the same sign on 20 of the 28 tasks and correlates at $r = 0.64$ ($P=0.001$ against a noise-only null; Figure~\ref{fig:analysis-separation}c).

\vspace{-5pt}
\paragraph{Model-specific priors rather than domain specialization.}
These repeatable task advantages track the underlying base model rather than traditional scientific disciplines. Among the eleven configurations sharing at least 90 scored tasks, Astra Medium and Astra Max, which share the same base model at different reasoning efforts, exhibit the strongest per-task score correlation ($r = 0.56$, against a median of 0.26 and a maximum of 0.56 across the other 54 pairs; $P<0.0005$; Figure~\ref{fig:analysis-separation}d). By contrast, discipline and subfield grouping explain almost none of the configuration-by-task interaction (5\% at the discipline level, $P=0.08$, and indistinct from chance at the subfield level, $P=0.29$), and nine of the thirteen configurations achieve the sole highest score on at least one of the 25 most discriminating tasks. Thus, success on a task depends on whether a model's specific scientific priors align with the governing mechanism of that experiment.

\vspace{-5pt}
\paragraph{What the ranking is built on.}
On a single task the quantity a run ranks by matters more than how much it computes. A task that ranks eight organoboron emitters by fluorescence quantum yield was attempted by all thirteen configurations with quantum-chemical calculations lasting 4.7 to 8.0 hours; the five runs that ranked by the channels that drain the emitting state scored 0.55 to 0.88, whereas the four that ranked mainly by the strength of the emitting transition scored $-0.50$ to 0.17. On a task that ranks 271 hydrogel formulations by the force needed to pull them off glass, the measured order follows the composition, and the ten runs that scored above 0.5 all reproduce the signs of its four strongest compositional correlations, while the run that ranked by monomer adsorption energies on silica scored $-0.38$ (Figure~\ref{fig:analysis-separation}e). Seven of the tasks examined in Section~\ref{sec:analysis-review} are among the 25 with the widest spread.

\subsection{What Varies Between Repeated Runs}
\label{sec:analysis-repeats}

The five extra runs of Astra Max on each of 29 tasks (Section~\ref{sec:repeated-runs}) measure how much of a single score belongs to the configuration and how much to the run, and their trajectories show where the difference enters.

\begin{figure}[!htbp]
  \centering
  \includegraphics[width=\linewidth]{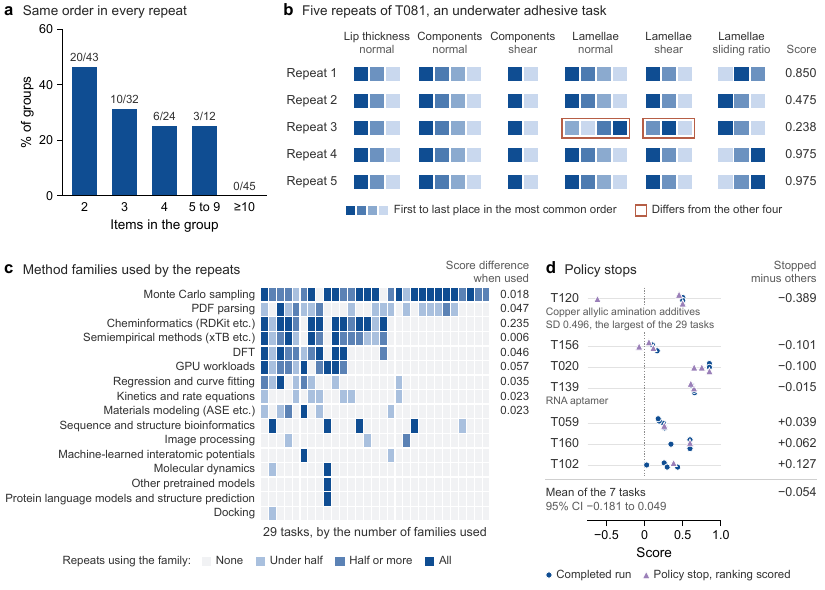}
  \caption{\textbf{What varies between repeated runs.} \textbf{a}, Share of the comparison groups that received the same order in every scored repeat of their task, by the number of items in the group. \textbf{b}, Order submitted for each group of an underwater adhesive task in its five repeats, with the items shaded by their place in the most common order of the group, and the score of each repeat. \textbf{c}, How many of the repeats of each task used each method family, and the difference in score between the repeats of a task that used a family and those that did not, for families used by some but not all repeats on at least four tasks. \textbf{d}, Scores of the repeats on the tasks where some repeats stopped under the information policy, and the mean score of the stopped repeats minus that of the others, with the mean over these tasks and its 95\% bootstrap interval.}
  \label{fig:analysis-repeats}
\end{figure}

A single run is an unbiased, if noisy, measure of a configuration on a task. The within-task and between-task spreads of Section~\ref{sec:repeated-runs} give single-run scores an intraclass correlation of 0.655 (95\% CI 0.389 to 0.846), and averaging runs raises the reliability along the curve of Figure~\ref{fig:public-runs}b. The official run that the rest of this paper uses does not flatter Astra Max on these tasks. Its official score sits a little below the mean of its repeats, by an interval that covers zero, and exceeds every repeat on one of the 29 tasks (Figure~\ref{fig:public-repeated-runs}a). The official scores of Astra Medium average 0.004 above the mean of the Astra Max repeats on the same tasks (95\% CI $-0.045$ to 0.054).

A configuration resembles itself more than it resembles the others. Measuring the agreement between two submitted rankings by the task score $S_t$ of Section~\ref{sec:evaluation}, with the second ranking in place of the measured ranks, two scored repeats of the same task agree with each other more closely than a repeat agrees with the official run of another configuration (Figure~\ref{fig:public-repeated-runs}c), and 27 of the 261 pairs of repeats submitted identical rankings. The gap runs the same way on 27 of the 29 tasks and averages 0.173 (95\% CI 0.120 to 0.231). This is the same configuration-by-task interaction that Section~\ref{sec:analysis-separation} found in single-run scores, seen from the side of the submitted order.

How often the exact order recurs depends almost entirely on how many items the group holds. Of the 156 comparison groups in these 29 tasks, 39 received an identical order in every scored repeat of their task. All 39 hold at most nine items, and the rate falls with group size, from 20 of 43 groups of two items to 6 of 24 groups of four. None of the 45 groups with ten or more items received the same order in every repeat. A repeat therefore reproduces its predecessor's answer where there are few orders to choose between (Figure~\ref{fig:analysis-repeats}a).

Which methods a run happens to use varies between repeats as well, and it barely moves the score. Of the 129 combinations of a task and a method family that at least one repeat used, only 49 were used by every repeat of that task, and on the median task more than half the families appear in some repeats and not others. Within a task, the difference in score between repeats that used a family and those that did not is at most 0.235, for cheminformatics over 4 tasks and 20 runs, and below 0.06 for the other eight families. The methods are thus part of what varies rather than what drives the score, which is the finding of Section~\ref{sec:analysis-score} seen inside one configuration (Figure~\ref{fig:analysis-repeats}c).

What the spread does track is the choice of the quantity that governs the measured endpoint. On a task that ranks 19 formulations of an underwater adhesive prototype, five repeats all finished normally and scored from 0.238 to 0.975. Three of the six groups received the same order in all five, and all three are ranked by adhesion strength. The spread comes from the one group the task defines by a sliding ratio, three lamella layouts ranked by how long they retain shear adhesion under load. That group carries a quarter of the score and drew three different orders: two repeats matched the measured order, two submitted its exact reverse, and the fifth swapped one pair. The whole distance between the best repeats and the second-worst comes from this group alone. The worst repeat took the reversed order and also departed on the two groups of lamella variants under load, swapping the leading pair in one and nearly reversing the other (Figure~\ref{fig:analysis-repeats}b).

Incidents during a run explain much less than this. Fourteen of the 137 scored repeats stopped under the information policy and submitted a provisional ranking. Across the seven tasks that have repeats of both kinds, a stop costs 0.054 on average with an interval that covers zero (95\% CI $-0.181$ to 0.049), and the stopped repeats scored higher on three of the seven. It matters on one task, the copper allylic amination additives, where the three stopped repeats averaged 0.389 below the two that finished, nearly all of it from one stopped repeat that scored $-0.619$, and where the within-task standard deviation, 0.496, is the largest of the 29 (Figure~\ref{fig:analysis-repeats}d).

For the ranking of configurations, this run-to-run variation preserves both the two-tier separation and the leading configuration while leaving fine-grained ordering within each tier unresolved. In 10,000 simulated reruns that add this noise to every official score, the leading six configurations stay above the other seven in 98.6\% of simulations and Opus 5.5 stays first in 82\%, whereas the exact thirteen-way ordering recurs in only 0.1\% (Figure~\ref{fig:public-runs}f). When tasks are resampled as well, the leading six stay above the other seven in 69.2\%.

\subsection{Scores by Discipline}
\label{sec:analysis-discipline}

\begin{figure}[!htbp]
  \centering
  \includegraphics[width=\linewidth]{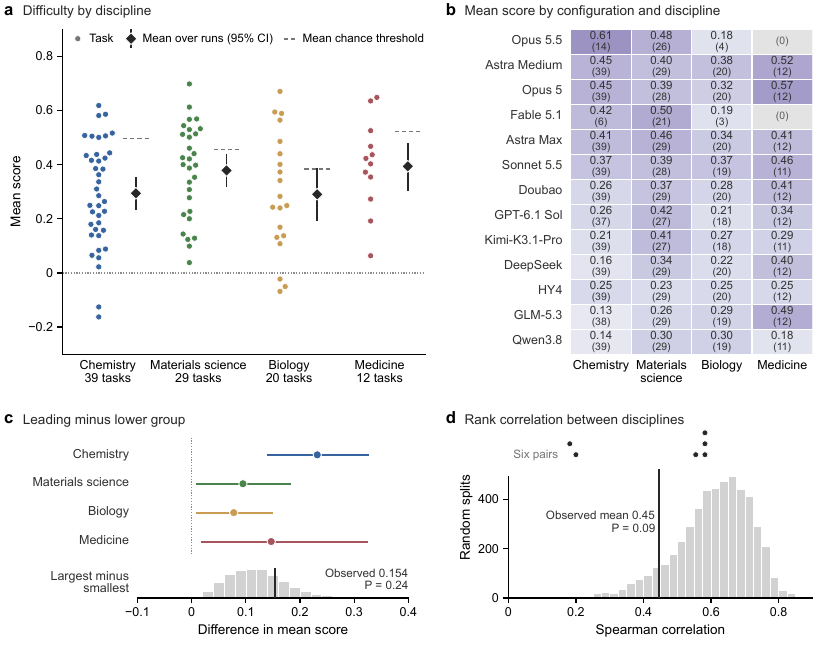}
  \caption{\textbf{Scores by discipline.} \textbf{a}, Mean score of each public task over the configurations scored on it, by discipline, with the raw mean of the discipline and its mean chance threshold. \textbf{b}, Raw mean score of each configuration in each discipline, with the number of scored tasks. \textbf{c}, Mean score of Astra Medium, Opus 5, Astra Max and Sonnet 5.5 minus that of DeepSeek, Qwen3.8, GLM-5.3, Doubao, HY4 and Kimi-K3.1-Pro on the same task, by discipline; below, the largest minus the smallest of the four differences, against its values when the disciplines of the tasks are shuffled (grey). \textbf{d}, Rank correlation between the discipline scores of the eleven configurations for the six pairs of disciplines (points) and their mean (line), against random splits of the task set into blocks of the same sizes (grey). Error bars in \textbf{a} and \textbf{c} are 95\% intervals.}
  \label{fig:analysis-discipline}
\end{figure}

The four disciplines differ in how well the configurations score, in how high a score has to be to beat chance, and in which configurations are scored at all. Over the scored runs, materials science and medicine average 0.378 and 0.394, against 0.294 in chemistry and 0.290 in biology (Figure~\ref{fig:analysis-discipline}a and Table~\ref{tab:discipline}; Appendix~\ref{app:exp-discipline}). The chance threshold of Section~\ref{sec:evaluation} falls as a task holds more experiments, and biology tasks hold the most, a median of 47 against 13.5 to 18 in the other three, so the threshold averages 0.384 in biology and 0.522 in medicine; the share of runs that beat it therefore orders the disciplines differently from the mean, at 49\% in materials science against 37\% in chemistry and 38\% in biology. Refusals fall almost entirely outside materials science, removing 3\% of its 377 official runs against 12\% to 15\% in the other three (Section~\ref{sec:run-outcomes}), which leaves Opus 5.5 and Fable 5.1 with no scored medicine task and most of their coverage in materials science (Figure~\ref{fig:analysis-discipline}b). Each discipline has a different leader.

\vspace{-5pt}
\paragraph{How far the discipline differences carry.}
Eleven configurations have a scored run in every discipline, so a comparison restricted to them rests on the same configurations in each. Astra Medium, Opus 5, Astra Max and Sonnet 5.5, the four of the leading six that remain, score above the six configurations that the paired tests separate from the leading six (Section~\ref{sec:public-results}), DeepSeek, Qwen3.8, GLM-5.3, Doubao, HY4 and Kimi-K3.1-Pro, on the same task in all four disciplines, by a margin that is widest in chemistry and narrowest in biology, 0.232 against 0.078 (Figure~\ref{fig:analysis-discipline}c). The gap does not depend detectably on the discipline: the widest minus the narrowest, 0.154, sits inside what permuting the discipline labels of the 100 tasks gives for that statistic (median 0.116, $P=0.239$ over 5,000 permutations). The ordering of the eleven configurations is no more stable across disciplines than across random blocks of the task set: their mean scores correlate at 0.45 on average between pairs of disciplines, below the median of 0.62 for random splits of the task set into blocks of the same sizes but above 9\% of those splits (Figure~\ref{fig:analysis-discipline}d), consistent with the absence of strong discipline specialization found in Section~\ref{sec:analysis-separation}. What changes clearly from one discipline to the next is how hard the tasks are, how high the chance threshold sits and which configurations have a score at all.

\subsection{A Closer Look at Selected Runs}
\label{sec:analysis-review}

\begin{figure}[!htbp]
  \centering
  \includegraphics[width=\linewidth]{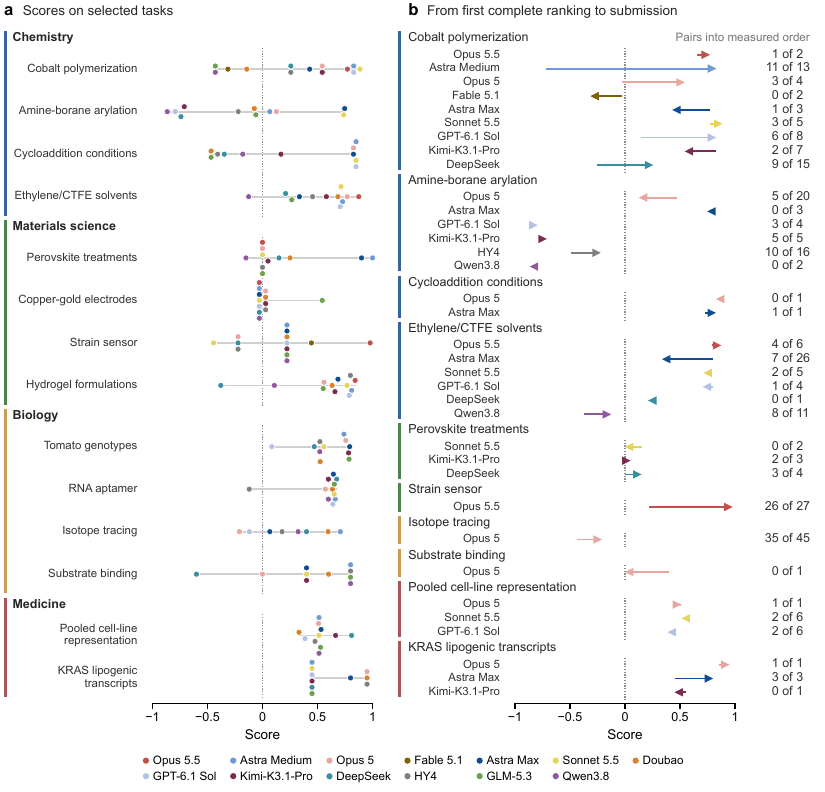}
  \caption{\textbf{Scores and revisions on selected tasks.} \textbf{a}, Score of every scored official run on the tasks examined in detail, grouped by discipline. \textbf{b}, Score of the first complete ranking and of the submission, for those of these runs whose submission differs from it; numbers count the flipped pairs that moved into the measured order.}
  \label{fig:analysis-review}
\end{figure}

The labels of Section~\ref{sec:analysis-run} cover every run, but they do not say what a submission was built on. The trajectories and reports do, and we single out here tasks whose runs show a distinctive way of working. This section follows the runs that scored well above the chance threshold of their task, one discipline at a time (Figure~\ref{fig:analysis-review}; Appendix~\ref{app:exp-review}).

In chemistry, Opus 5.5 scored 0.874 on a task that ranks twelve solvent conditions by the isolated mass of an ethylene/CTFE copolymer, the highest of the twelve configurations scored on it and well above the chance threshold of 0.503. Its run built a Hansen-type phase and solubility model, computed transfer-to-solvent and cross-propagation barriers at M06-2X, and sampled by Monte Carlo the weights it could not derive. It placed the three polar solvents last because the comonomer is largely immiscible with them, they take up 3 to 40 times less ethylene, and the lipophilic photocatalyst falls outside their solubility sphere. Each of those three statements is about the quantity the task measures.

In materials science, a task that ranks the resistance response of a mica-supported Mo-W-Nb strain sensor at four temperatures separated the configurations further than most (Section~\ref{sec:analysis-separation}). Its measured orders are nearly unrelated to temperature; ten of the thirteen configurations submitted orders monotone in temperature and scored 0.222 or $-0.222$. The Opus 5.5 run, which scored 0.978, spent 7.4 hours on a Monte Carlo model of the contacts between grains in the sensor film, and 26 of the 27 pairs it flipped after its first draft moved into the measured order. It committed to one placement, 900\,\textdegree C last, which it argued from neck growth above the anneal temperature of 800\,\textdegree C, the NbO$_2$ transition near 808\,\textdegree C and the large metallic resistivity at that temperature. Its report called the other three temperatures nearly tied and its own confidence low.

In biology, the Astra Max run on a task that ranks 110 tomato genotypes by field branching scored 0.792 against a threshold of 0.179. It converted each genotype into a functional activity, with a residual activity for the homozygote and an uncertain dosage term for the heterozygote, and fitted nothing to the measured endpoint. It also tested one of its own features before using it: having scanned both strands with Arabidopsis binding matrices, it compared the maximum scores against 2,000 permutations that preserve length and base composition, found the smallest $P$ at 0.082 and dropped the motif scores from its model.

In medicine, the clearest lead over the other configurations came from a run that finished in minutes. DeepSeek answered a task that ranks eleven cell lines in a mixed culture by their representation after drug treatment in under a quarter of an hour and 81 tool calls, at 0.809 against a threshold of 0.368, where the next configuration reached 0.664. Its run retrieved genotype, copy number and expression for the panel, requested no treatment-response endpoint, and predicted a block structure: in each arm the three lines whose driver matches the covalent target of the drug fall to the bottom, and the remaining eight sit near parity with the vehicle.

What these runs share is that the quantity they ranked by is the quantity the task measures. The margin over chance was no larger for the two that computed for six to seven and a half hours than for the two that finished within about an hour. Runs that scored poorly on the same tasks often applied the same methods to something else. On a task that ranks ligands by the conversion they give in a cobalt-catalyzed polymerization, Astra Medium, GLM-5.3 and Qwen3.8 all ran semiempirical calculations and a kinetic model, and scored 0.829, $-0.429$ and $-0.429$; the trajectories were not describing the same species, since one built the ligand with the wrong relative stereochemistry and another never added explicit hydrogens although its own tool output reported them missing. Section~\ref{sec:analysis-score} found no method family associated with score, and what the method is applied to is one reason why.

\subsection{Use of the GPU}
\label{sec:analysis-gpu}

\begin{figure}[!htbp]
  \centering
  \includegraphics[width=\linewidth]{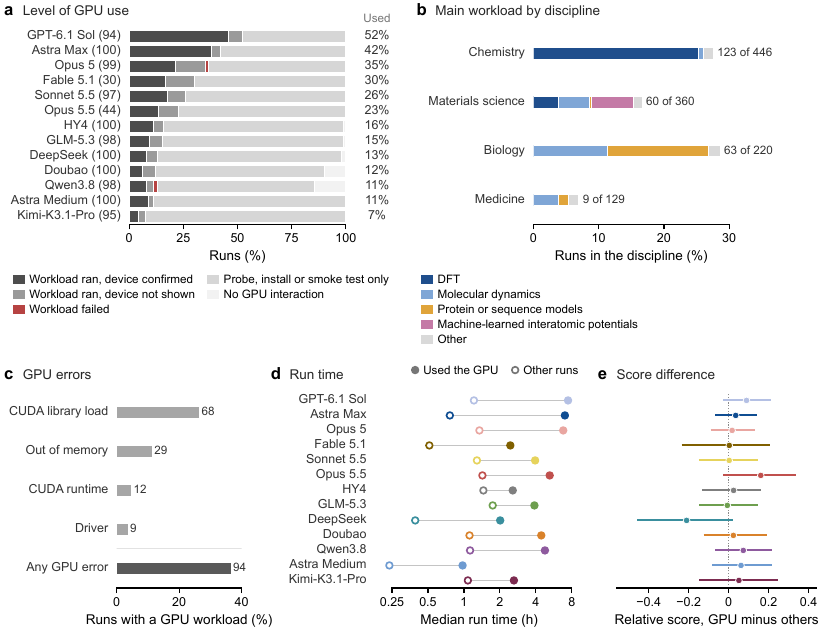}
  \caption{\textbf{Use of the GPU.} \textbf{a}, Share of runs at each level of GPU use, by configuration. \textbf{b}, Main workload of the runs that used the GPU, by discipline. \textbf{c}, Share of the runs that started a workload hitting each kind of GPU error. \textbf{d}, Median run time of runs that used the GPU and of the others, by configuration. \textbf{e}, Difference between runs that used the GPU and the other runs of the same configuration, after subtracting the mean of the other configurations on the same task, with 95\% intervals.}
  \label{fig:analysis-gpu}
\end{figure}

Every container carries one NVIDIA H20 shared by all local processes, and the task instruction asks the agent to check the available hardware before deciding what to run (Section~\ref{sec:task}). Almost every run did so, 98\% probing the GPU and 89\% within their first ten tool calls, but computing on it was rarer: a GPU workload ran without failing in 255 runs (22\%), and in 185 of those the program's own output names the device (Figure~\ref{fig:analysis-gpu}a; Appendix~\ref{app:exp-gpu}). The share that used the GPU ranges from 7\% of Kimi-K3.1-Pro runs to 52\% of GPT-6.1 Sol runs. DFT accounts for half of the 255 runs, ahead of molecular dynamics, protein and sequence models and machine-learned interatomic potentials, and the mix follows the discipline of the task (Figure~\ref{fig:analysis-gpu}b).

\vspace{-5pt}
\paragraph{What it costs and what it returns.}
Runs that used the GPU lasted a median of 4.9 hours against 1.0 hours for the rest, and the gap holds within every configuration (Figure~\ref{fig:analysis-gpu}d). The task image leaves the GPU libraries to the agent, to install like any other software it needs (Section~\ref{sec:task}), and 36\% of the 258 runs that started a workload hit a GPU error, most often a CUDA library that failed to load (Figure~\ref{fig:analysis-gpu}c). The return is small at best. Over the 63 tasks that have runs on both sides, the 255 runs that ran a workload score 0.035 above the others of the same configuration (95\% CI $-0.019$ to 0.090; Holm-corrected $P=1.0$ within the method families of Section~\ref{sec:analysis-score}); restricted to the 185 runs whose output names the device it is +0.051, with an interval just clear of zero (95\% CI 0.011 to 0.092; permutation $P=0.04$). Taken one configuration at a time the differences fall on both sides of zero and none is detectable (Figure~\ref{fig:analysis-gpu}e). This pattern suggests that GPU acceleration speeds up the evaluation of a chosen proxy without moving that proxy closer to the measured endpoint.

\subsection{Configuration Profiles}
\label{sec:analysis-profiles}

\begin{figure}[!htbp]
  \centering
  \includegraphics[width=\linewidth]{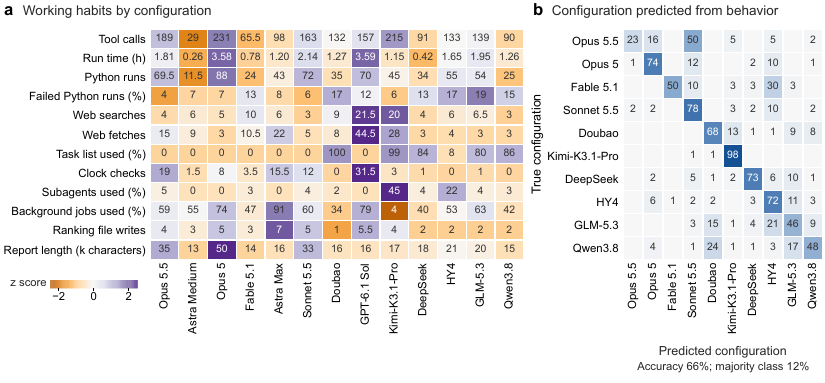}
  \caption{\textbf{Configuration profiles.} \textbf{a}, Working habits of each configuration, as medians over its runs or, for task lists, subagents and background jobs, as the percentage of runs that used them; colors show $z$ scores across the thirteen configurations, and task lists and subagents are not available in Codex. \textbf{b}, Confusion matrix of a random forest that predicts a run's configuration from its behavior, for the 861 runs of the ten configurations under Claude Code (five-fold cross-validation).}
  \label{fig:analysis-profiles}
\end{figure}

A configuration pairs one model with one agent harness at one reasoning effort (Section~\ref{sec:exp-setup}). The thirteen configurations work in recognizably different ways. Given only behavioral features (Appendix~\ref{app:exp-trajectory}), a random forest identifies which of the ten Claude Code configurations produced a run with 66\% accuracy in five-fold cross-validation, against 12\% for always guessing the most common configuration (Figure~\ref{fig:analysis-profiles}b); the configurations under Codex are left out, since the harness alone would identify them. Kimi-K3.1-Pro, Sonnet 5.5 and Opus 5 are the easiest to recognize, while Fable 5.1 is most often mistaken for HY4 and Qwen3.8 for Doubao. The habits behind the fingerprints are specific and stable (Figure~\ref{fig:analysis-profiles}a): Astra Max starts background jobs in 91\% of its runs and writes its ranking file a median of seven times, whereas Qwen3.8 and Sonnet 5.5 search least, with a median of three queries per run.

\subsection{The Same Model at Two Efforts}
\label{sec:analysis-effort}

\begin{figure}[!htbp]
  \centering
  \includegraphics[width=\linewidth]{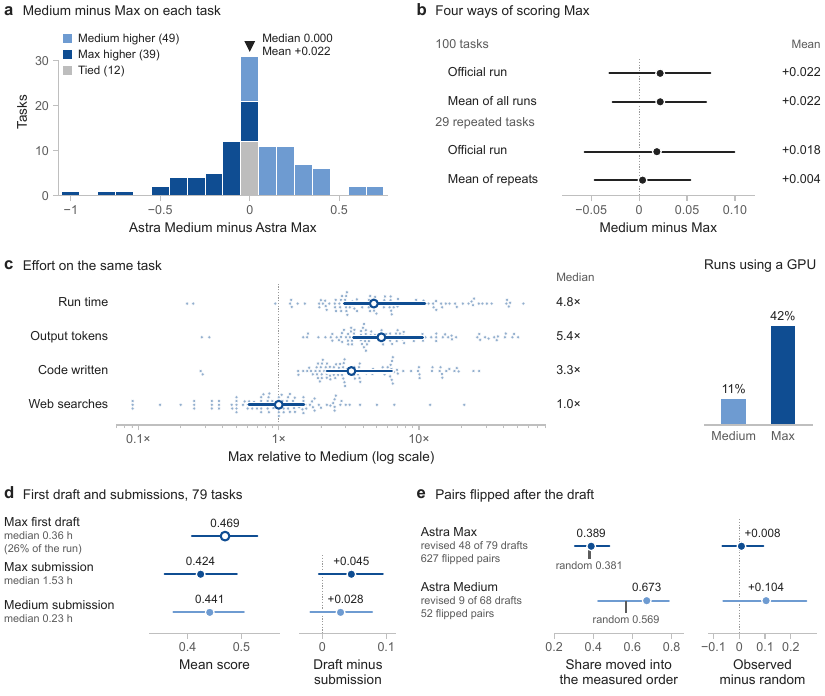}
  \caption{\textbf{The same model at two reasoning efforts.} \textbf{a}, Difference in score between Astra Medium and Astra Max on each of the 100 public tasks. \textbf{b}, Mean difference under four ways of scoring Max, with 95\% bootstrap intervals. \textbf{c}, Ratio of Max to Medium on the same task for run time, output tokens, code written and web searches, and the share of runs that used a GPU. \textbf{d}, Score of Max's first complete ranking, of its own submission and of Medium's submission on the tasks where that draft is visible, and the draft minus each submission, with 95\% intervals. \textbf{e}, Share of the item pairs flipped after the draft that moved into the measured order, against the share expected from random flips at the same distance in that draft, and the difference, with 95\% intervals.}
  \label{fig:analysis-effort}
\end{figure}

Astra Medium and Astra Max differ only in reasoning effort, and Medium's mean is 0.022 above Max on the same 100 tasks (Section~\ref{sec:public-results}), a difference indistinguishable from zero: the per-task differences center on zero, with Medium ahead on somewhat more tasks than Max (Figure~\ref{fig:analysis-effort}a), and the gap survives trimming the few tasks on which the two diverge most, so no part of the task set carries it. Four ways of scoring Max, including the mean of its five repeated runs on the 29 tasks of Section~\ref{sec:repeated-runs}, all place the difference between zero and 0.022 with an interval that covers zero (Figure~\ref{fig:analysis-effort}b).

Max runs a median of 4.8 times as long as Medium on the same task and writes several times more code and output while searching the web no more often (Figure~\ref{fig:analysis-effort}c), yet how much it spends on a task says nothing about which configuration scores higher there. The extra time goes into revising a ranking that was already as accurate. Max writes its first complete ranking about a quarter of the way into its run, by which point the whole Medium run on the same task has usually ended, and that draft already scores as well as both its own submission and Medium's (Figure~\ref{fig:analysis-effort}d). It then revises most of its runs, where Medium rarely does, but the item pairs it flips move into the measured order no more often than random flips at the same distance in its own draft (Figure~\ref{fig:analysis-effort}e). Scoring Max by its first ranking wherever one exists is enough to reverse the comparison, to 0.014 in its favor.

\section{Related Work}
\label{sec:related}

\vspace{-5pt}
\paragraph{Predicting experimental outcomes under hidden ground truth.}
Testing predictions against experimental data that are not yet public has a long history in computational science. Blind challenges such as CASP~\citep{kryshtafovych2026casp16}, CAPRI~\citep{lensink2025capri} and SAMPL~\citep{amezcua2024sampl9} withhold measured structures or binding free energies while teams submit predictions. They established the principle that the answer must stay hidden from the predictor until scoring. Their predictors are research groups and automated servers, each applying its own established method, and each challenge covers a narrow set of properties. Recent benchmarks apply this blind-prediction or expert-level testing paradigm to language models: Humanity's Last Exam~\citep{phan2025hle} tests frontier reasoning on closed-ended academic questions, BrainBench~\citep{luo2024brainbench} asks models to distinguish real from altered neuroscience abstracts, SciPredict~\citep{sehwag2026scipredict} poses 405 natural-science outcome questions with retrieval disabled, ForecastBench~\citep{karger2025forecastbench} evaluates predictions of future events, and related studies test whether models can anticipate the results of social-science experiments~\citep{ashokkumar2026llmsocialscience}. Because these evaluations query models in a single pass without tools, they measure parametric recall and reasoning rather than the ability to gather and test new scientific evidence, the gap between exam-style and discovery-oriented evaluation of AI scientists that \citet{du2026measuring} identify. On the ranking side, ProteinGym~\citep{notin2023proteingym}, TDC~\citep{huang2021tdc} and Matbench~\citep{dunn2020matbench} evaluate supervised predictors on experimental property rankings using Spearman-type metrics, while AssayBench~\citep{debrouwer2026assaybench} and its follow-up~\citep{edwards2026assayloop} frame CRISPR screens as gene-ranking tasks. \MYBENCH retains rank-based scoring against hidden ground truth, but the predictor is an autonomous agent with full web and compute access, operating zero-shot without task-specific training data: each experiment appears once, and the agent must decide for itself which evidence and calculations bear on the outcome.

\vspace{-5pt}
\paragraph{Science agents and their evaluation.}
One line of work demonstrates what science agents can do. Coscientist~\citep{boiko2023coscientist} and ChemCrow~\citep{bran2024chemcrow} call chemistry tools to plan and run syntheses; El Agente~\citep{zou2025elagente} and MDCrow~\citep{campbell2025mdcrow} carry out quantum-chemistry and molecular-dynamics workflows; hypotheses or designs from Virtual Lab~\citep{swanson2025virtuallab}, Co-Scientist~\citep{gottweis2026coscientist}, Robin~\citep{ghareeb2025robin} and Biomni~\citep{huang2025biomni} have been validated in the wet lab; and the AI Scientist~\citep{lu2024aiscientist,yamada2025aiscientistv2}, DeepScientist~\citep{weng2025deepscientist} and AlphaEvolve~\citep{novikov2025alphaevolve} iterate autonomously in domains with automatic verification. These studies evaluate each system on problems chosen by its authors, without a shared task set or scoring rule, so their results cannot be compared directly. A second line turns scientific ability into benchmarks: ScienceAgentBench~\citep{chen2025scienceagentbench} grades programs written for data-driven discovery tasks, BLADE~\citep{gu2024blade} evaluates analysis decisions on open datasets, BixBench~\citep{mitchener2025bixbench} and SciAgentGym~\citep{shen2026sciagentgym} evaluate multi-step bioinformatics and scientific tool use, and Terminal-Bench-Science~\citep{tbscience2026} collects 70 research-workflow tasks across five fields, each verified by deterministic tests. What these benchmarks share is that the ground truth follows deterministically from the provided data or specification; they test whether an agent executes a prescribed workflow correctly. In \MYBENCH, the outcome cannot be derived from the inputs alone and is settled only by wet-lab measurement---making the choice of an appropriate physical or chemical proxy part of the challenge itself, and a primary source of failure when agents compute quantities that do not govern the measured endpoint (\S\ref{sec:analysis-review}).

\vspace{-5pt}
\paragraph{Long-horizon, containerized agent benchmarks.}
A single \MYBENCH run can last up to eight hours, with hundreds of tool calls and long model outputs, so it is also a long-horizon agent evaluation. This line began in software engineering, where SWE-bench~\citep{jimenez2024swebench} and SWE-Lancer~\citep{miserendino2025swelancer} ask agents to resolve issues in real repositories. MLE-bench~\citep{chan2025mlebench}, PaperBench~\citep{starace2025paperbench} and RE-Bench~\citep{wijk2024rebench} extend the horizon to hours of machine-learning engineering and research replication, and METR's time-horizon metric~\citep{kwa2025measuring} makes ``how long a task an agent can complete'' a single scale of capability. OSWorld~\citep{xie2024osworld}, $\tau$-bench~\citep{yao2024taubench}, TheAgentCompany~\citep{xu2024theagentcompany}, Toolathlon~\citep{li2026toolathlon} and Agents' Last Exam~\citep{sun2026ale} extend this kind of evaluation to graphical interfaces, conversational tool use and complex multi-application workflows. Terminal-Bench~\citep{merrill2026terminalbench,terminalbench4} packages tasks as reproducible container environments run through the Harbor framework~\citep{harbor2026}; its latest release allows internet access and uses a uniform eight-hour limit. Long-Horizon-Terminal-Bench~\citep{li2026longhorizonterminal} pushes single tasks to about 85 minutes and close to ten million tokens on average. \MYBENCH tasks are also packaged and run in Harbor, but depart from software and workflow benchmarks in three ways. First, grading compares submitted rankings against external wet-lab measurements in an isolated container rather than by running in-container unit tests, leaving no verifier script for an agent to inspect or game. Second, while longer time horizons and heavier compute consistently improve performance on software benchmarks~\citep{kwa2025measuring}, spending more time, tokens or GPU hours in \MYBENCH yields no detectable gain in prediction accuracy (\S\ref{sec:public-results}, \S\ref{sec:analysis-effort}). Third, because open web access makes published outcomes discoverable in principle, our protocol combines an isolated grader with anonymized tasks, an information policy with audited tools, automated screening of tool outputs for source identifiers, and unpublished private tasks (\S\ref{sec:private}, \S\ref{sec:audit}), and runs that encounter source identifiers show no significant score advantage in any configuration. Table~\ref{tab:comparison} summarizes how \MYBENCH compares with representative evaluations across these axes.

\begin{table}[t]
\centering
\caption{\textbf{\MYBENCH compared with related evaluations.} \emph{Wet-lab outcome}: the target is a measured experimental result. \emph{Agentic compute}: the model runs code and scientific software of its own choosing. \emph{Hours-long runs}: a single run may last hours (Terminal-Bench 4.0: 8\,h limit). \emph{Open web}: retrieval is allowed during the run (\pmark: allowed with an instruction not to look up answers, but not audited). \cmark{} yes, \pmark{} partly, \xmark{} no, -- not stated in the release. $^\dagger$AssayBench evaluates agent baselines alongside direct LLM prediction.}
\label{tab:comparison}
\small
\setlength{\tabcolsep}{4pt}
\resizebox{\linewidth}{!}{%
\begin{tabular}{lllcccc}
\toprule
Benchmark & Domain & Output & \makecell{Wet-lab\\outcome} & \makecell{Agentic\\compute} & \makecell{Hours-long\\runs} & \makecell{Open\\web} \\
\midrule
Humanity's Last Exam~\citep{phan2025hle} & Multi-discipline & Closed-ended QA & \xmark & \xmark & \xmark & \xmark \\
BrainBench~\citep{luo2024brainbench} & Neuroscience & 2-way choice & \cmark & \xmark & \xmark & \xmark \\
SciPredict~\citep{sehwag2026scipredict} & Phys./Chem./Bio. & Answer / value & \cmark & \xmark & \xmark & \xmark \\
ProteinGym~\citep{notin2023proteingym} & Protein fitness & Spearman ranking & \cmark & \xmark & \xmark & \xmark \\
AssayBench$^\dagger$~\citep{debrouwer2026assaybench} & CRISPR screens & Gene ranking & \cmark & \pmark & \xmark & \xmark \\
ScienceAgentBench~\citep{chen2025scienceagentbench} & Data-driven science & Program & \xmark & \cmark & \xmark & \xmark \\
Terminal-Bench-Science~\citep{tbscience2026} & Five sci.\ domains & Tested artifact & \xmark & \cmark & -- & -- \\
Agents' Last Exam~\citep{sun2026ale} & Tool-use tasks & Task state / answer & \xmark & \cmark & \cmark & \pmark \\
Terminal-Bench 4.0~\citep{terminalbench4} & Mixed & Tested state & \xmark & \cmark & \cmark & \pmark \\
Long-Horizon-TB~\citep{li2026longhorizonterminal} & Mixed & Graded subtasks & \xmark & \cmark & \cmark & -- \\
\midrule
\rowcolor{blue!6} \textbf{\MYBENCH (ours)} & \shortstack[l]{Chem./Mat./\\Bio./Med.} & Ranking & \cmark & \cmark & \cmark & \shortstack{\cmark\\[-2pt]{\scriptsize policy + screening}} \\
\bottomrule
\end{tabular}}
\end{table}

\section{Conclusion}
\label{sec:conclusion}
\enlargethispage{\baselineskip}

We introduced the \textbf{Science Prediction Universe (\MYBENCH)}, a benchmark that evaluates whether autonomous language-model agents can predict the outcomes of real wet-lab experiments across chemistry, materials science, biology and medicine when the measured endpoints are withheld. Across 100 public tasks from recent Nature Portfolio studies and 16 private tasks from unpublished wet-lab campaigns, thirteen agent configurations spanning twelve frontier models separate into two distinct tiers yet reach a best task-adjusted Spearman correlation of 0.478 on the public benchmark, while twelve of the thirteen rank the unpublished private experiments significantly above chance. Even with up to eight hours of container execution, web search and GPU access, longer runs and heavier simulations did not yield detectably more accurate predictions. Trajectory analysis suggests that the bottleneck lies in scientific judgment rather than tool execution: agents share monotone priors, agree more with one another than with measurement (0.52 vs.\ 0.33), and often compute tractable proxies that do not govern the experimental endpoint. Because a correct prediction of an unperformed experiment is the strongest evidence of scientific understanding, \MYBENCH measures how far current agents are from that standard, and the private tasks will keep it measurable as the public ones age. Bridging the gap from autonomous computation to experimental reality will require agents that can identify which mechanism controls a measurement and test their own physical assumptions before committing to a prediction.

\bibliography{custom}
\bibliographystyle{unsrtnat}

\clearpage
\appendix
\section{Authors}
\label{app:authors}

Each contributor is listed with their affiliations (superscripts), keyed to the numbered list at the end of this section.

\noindent\textbf{Project Team}\\
Zengji Tu\textsuperscript{1,*}, Xiang Liu\textsuperscript{1,*}, Jiarui Sun\textsuperscript{1}, Ruoyu Wu\textsuperscript{1}, Zhihui Qi\textsuperscript{1}, Chengda Wu\textsuperscript{1}, Yutao Zhou\textsuperscript{1}, Bo Yang\textsuperscript{1}, Ziyan Zhang\textsuperscript{1}, Jialong Wu\textsuperscript{1}, Liang Chen\textsuperscript{1,\dag,\ddag}, Kuan Li\textsuperscript{1,\dag}\par
\vspace{4pt}

\noindent\textbf{Advisory Committee}\\
Steven Dillmann\textsuperscript{2}, Yuanqi Du\textsuperscript{3}, Soojung Yang\textsuperscript{4}, Shizhe Diao\textsuperscript{5}, Jianjun Wu\textsuperscript{6}, Junlang Liu\textsuperscript{7}, Xuming Hu\textsuperscript{8}, Lin Chang\textsuperscript{6,9,\dag}, Baobao Chang\textsuperscript{10}, Xiaowen Chu\textsuperscript{8}, Philip S. Yu\textsuperscript{11}, Tong-Yi Zhang\textsuperscript{8,12}\par
\vspace{4pt}

\noindent\textbf{Wet-Lab Collaborators}\\
Hao Cheng\textsuperscript{13}, Qin Han\textsuperscript{13}, Yanye Feng\textsuperscript{14}, Chenru Duan\textsuperscript{15}, Chuncheng Liu\textsuperscript{15}, Yan Liu\textsuperscript{15}, Yuchen Gao\textsuperscript{15}, Zhao Li\textsuperscript{15}, Ronghua Zhang\textsuperscript{15}, Tory Xu\textsuperscript{15}, Renjian Song\textsuperscript{16}, Yaokai Fu\textsuperscript{16}\par
\vspace{4pt}

\noindent{\footnotesize \textsuperscript{*}\,Core contributors.

 \textsuperscript{\dag}\,Corresponding authors: Liang Chen \href{mailto:liangchen@unipat.ai}{\texttt{liangchen@unipat.ai}}, Kuan Li \href{mailto:kuanli@unipat.ai}{\texttt{kuanli@unipat.ai}}, Lin Chang.

 \textsuperscript{\ddag}\,Project leader: Liang Chen.\par}

\vspace{6pt}

\noindent\textbf{Affiliations}\par
\vspace{2pt}
{\small
\noindent 1.~UniPat AI\\
2.~Stanford University\\
3.~Cornell University\\
4.~FutureHouse\\
5.~The Hong Kong University of Science and Technology\\
6.~State Key Laboratory of Photonics and Communications, School of Electronics, Peking University\\
7.~Department of Chemistry and Chemical Biology, Harvard University\\
8.~The Hong Kong University of Science and Technology (Guangzhou)\\
9.~Frontiers Science Center for Nano-optoelectronics, Peking University\\
10.~Peking University\\
11.~University of Illinois Chicago\\
12.~Shanghai University\\
13.~Yurogen Biosystems LLC\\
14.~ABclonal Biotechnology Co., Ltd.\\
15.~Deep Principle\\
16.~Bota Biosciences\par}

\section{Limitations, Ethics and Release}
\label{app:limitations}

\vspace{-5pt}
\paragraph{Benchmark scope and experimental noise.}
The public tasks evaluate agents retrospectively against published measurements, where fine-grained differences between neighboring conditions may occasionally fall within experimental uncertainty; to account for task size and measurement ties, we report the fraction of tasks exceeding the chance threshold alongside mean scores (\S\ref{sec:evaluation}). The private tasks provide a prospective test free of published answers, though their current sample size (16 tasks across four experimental series) yields wider confidence intervals (\S\ref{sec:private-results}). Expanding the domain coverage and scale of the private wet-lab suite is an ongoing priority.

\vspace{-5pt}
\paragraph{End-to-end configurations and coverage.}
Each evaluated configuration represents an end-to-end pairing of a model, reasoning effort and agent harness (Codex or Claude Code), and agents autonomously choose their run time, tool calls and GPU usage; consequently, cross-configuration comparisons reflect the combined system rather than isolated model weights, and our analyses of test-time effort, apart from the comparison of the two Astra efforts (\S\ref{sec:analysis-effort}), are observational (\S\ref{sec:exp-setup}). In addition, provider biosecurity filters caused Opus 5.5 and Fable 5.1 to refuse most tasks outside materials science (\S\ref{sec:run-outcomes}). Because statistical task adjustments cannot observe how a configuration would perform on refused domains, mean scores should always be interpreted jointly with task coverage. Future controlled ablations will isolate harness effects and evaluate fixed compute checkpoints.

\vspace{-5pt}
\paragraph{Release and contamination.}
Unlike static question-answering benchmarks, which rely on answers being difficult to memorize~\citep{phan2025hle}, \MYBENCH derives every target outcome of a public task from published literature, so these outcomes may eventually enter future pretraining corpora~\citep{sun2025contamination}. Leaderboard submissions therefore record each model's training cutoff alongside its evaluation set. The private tasks are withheld from public release for controlled evaluation; we plan to continuously expand the private suite and retire public tasks if contamination is detected.

\vspace{-5pt}
\paragraph{Ethics and broader impact.}
All public tasks derive from published studies, and neither the released dataset nor this paper redistributes raw experimental values or reference rankings. The private tasks were designed with partner laboratories from unpublished wet-lab campaigns whose measurements remain confidential. The benchmark involves no human subjects, and all agents execute within isolated containers with full trajectory logging. While accurate outcome prediction can accelerate experimental design and reduce wet-lab waste, overconfident mispredictions could misdirect experimental effort; we therefore report coverage, comparisons against chance and qualitative failure analyses alongside aggregate scores.
\section{Results by Discipline}
\label{app:task_details}

\begin{table}[!htbp]
\centering
\caption{\textbf{Raw mean score by discipline on the public tasks.} Each cell averages a configuration's scored official runs in one discipline, with the number of scored tasks in parentheses; the header gives the number of public tasks in each discipline, and -- marks a discipline without a scored task. Rows follow Table~\ref{tab:public-results}.}
\label{tab:discipline}
\small
\begin{tabular}{lccccc}
\toprule
Configuration & Chemistry (39) & Materials science (29) & Biology (20) & Medicine (12) & All (100) \\
\midrule
Opus 5.5 & 0.606 (14) & 0.475 (26) & 0.185 (4) & -- & 0.491 (44) \\
Astra Medium & 0.451 (39) & 0.398 (29) & 0.384 (20) & 0.520 (12) & 0.431 (100) \\
Opus 5 & 0.453 (39) & 0.394 (28) & 0.315 (20) & 0.565 (12) & 0.422 (99) \\
Fable 5.1 & 0.416 (6) & 0.502 (21) & 0.191 (3) & -- & 0.453 (30) \\
Astra Max & 0.409 (39) & 0.461 (29) & 0.335 (20) & 0.406 (12) & 0.409 (100) \\
Sonnet 5.5 & 0.374 (39) & 0.393 (28) & 0.370 (19) & 0.459 (11) & 0.388 (97) \\
Doubao & 0.263 (39) & 0.372 (29) & 0.276 (20) & 0.408 (12) & 0.315 (100) \\
GPT-6.1 Sol & 0.263 (37) & 0.422 (27) & 0.211 (18) & 0.340 (12) & 0.308 (94) \\
Kimi-K3.1-Pro & 0.206 (39) & 0.411 (27) & 0.273 (18) & 0.292 (11) & 0.287 (95) \\
DeepSeek & 0.160 (39) & 0.342 (29) & 0.224 (20) & 0.398 (12) & 0.254 (100) \\
HY4 & 0.249 (39) & 0.234 (29) & 0.249 (20) & 0.253 (12) & 0.245 (100) \\
GLM-5.3 & 0.129 (38) & 0.261 (29) & 0.286 (19) & 0.488 (12) & 0.242 (98) \\
Qwen3.8 & 0.139 (39) & 0.305 (29) & 0.301 (19) & 0.179 (11) & 0.224 (98) \\
\bottomrule
\end{tabular}
\end{table}

Table~\ref{tab:discipline} gives the raw mean score of each configuration in each discipline of the public tasks, and Section~\ref{sec:analysis-discipline} reads the differences between the disciplines. Opus 5.5 and Fable 5.1 have no scored medicine task and only four and three scored biology tasks because of their refusals (Section~\ref{sec:run-outcomes}). The two Astra configurations, Opus 5 and Sonnet 5.5 lead the lower group by the widest margin in chemistry, where they average 0.374 to 0.453 and DeepSeek, GLM-5.3 and Qwen3.8 average 0.129 to 0.160. GPT-6.1 Sol scores 0.263 in chemistry and 0.422 in materials science, its only discipline above the lower group. The disciplines differ in their source studies and task formats, so these comparisons are descriptive.

\section{Task Example}
\label{app:task_example}

Every task ships the same general instruction followed by a task-specific part. Below we reproduce the instruction of the organoboron emitter task of \S\ref{sec:analysis-separation} with light abridgment (marked [\dots]). The public package also contains \texttt{experiment.md}, \texttt{protocol.json}, \texttt{targets.csv}, structure images and the machine-readable contract; the hidden answers and the scorer live outside the agent's container.

\begin{tcolorbox}[taskbox, title={Task \texttt{organoboron-absolute-fluorescence-yield-ranking} (v1.0.0)}]
\textbf{General instruction (shared by all tasks).} You are a general-purpose AI scientist tasked with predicting experimental outcomes across scientific disciplines. [\dots] You may use all computational tools and search resources available in your environment, including web search, scientific literature, databases, code execution, statistical modeling, machine learning, and numerical simulation. [\dots] Proactively check the available CPUs, GPUs, and software [\dots] For scientifically grounded methods that may be useful but whose benefits are uncertain, prioritize small-scale trial calculations before deciding whether to expand the computation. [\dots] do not search for, access, or use the original study, its supplementary materials, or data that can directly reveal the experimental results, and do not use the results of the target experiments or experiments of the same type. [\dots] After forming an initial ranking, continue to look for and actually try methods that could improve key judgments within the time budget, prioritizing resources for the most uncertain comparisons. [\dots] You have a strict time limit of 8 hours (28,800 seconds).

\textbf{Prediction question.} Eight structurally specified organoboron molecular materials are examined in dilute dichloromethane solution. Predict the ordering of their absolute fluorescence quantum yields under the supplied common measurement definition. Molecular identity and sample characterization are given; all target optical outcomes remain hidden.

Rank each comparison group from lowest to highest using that group's specified endpoint and unit. Submit rankings only; absolute values are not required.

\textbf{Experimental definition.} Read \texttt{/workspace/public/experiment.md}, \texttt{protocol.json} and \texttt{targets.csv}. [\dots] Prepare dilute DCM solutions of each specified isolated molecular material at the common molar concentration and measure the absolute fluorescence quantum yield using the stated quantum-efficiency system. [\dots] Use one comparison group across the complete eight-molecule panel.

\begin{itemize}
\item \texttt{optical\_panel} --- Absolute fluorescence quantum yield (dimensionless): The source-reported absolute fluorescence quantum yield in DCM at 1e-5 mol/L, measured using C9920-02G. Rank all eight molecular materials from lower to higher. The table's fractional values are used directly, without treating them as percent numbers.
\end{itemize}

\textbf{Comparison group.} \texttt{optical\_panel}: Eight molecular solutions at the same concentration in DCM; 8 items; weight 1.0; ascending absolute fluorescence quantum yield (dimensionless).

\textbf{Allowed and prohibited information.} The original study, all its supplementary materials (including methods), source data, related reporting, summaries, mirrors and caches are prohibited. [\dots] Measured fluorescence quantum yields, optical absorption/emission outcomes, excited-state lifetimes, chiroptical outcomes and rankings of these or closely analogous organoboron fluorophores under comparable conditions, and models fitted to such outcomes, are prohibited. [\dots] If a permitted query unexpectedly exposes prohibited external results, immediately stop prediction and further search; record only the triggering step, source and stopping reason in \texttt{report.md} without restating results. [\dots]

\textbf{Resources.} 8 CPU cores, 16,384 MiB system RAM and one NVIDIA H20 GPU [\dots] The 8-hour clock starts at first delivery of the task inputs; retries do not reset it.

\textbf{Submission.} Write UTF-8 JSON to \texttt{/logs/artifacts/rankings.json} [\dots] Include the exact task identity and exactly one entry per group; each order must contain every assigned item once, from lowest to highest. [\dots] Write an English account of scientific reasoning, calculations, assumptions and uncertainty to \texttt{/logs/artifacts/report.md}; source records to \texttt{sources.json}; supporting scripts and calculations to \texttt{analysis/}. [\dots]

\textbf{Ranking scores.} The primary score is standard Spearman rank correlation, from $-1$ to $1$, computed separately in each group and combined with fixed group weights. [\dots] Random complete rankings have expected correlation 0.
\end{tcolorbox}

\section{Experimental Details and Additional Results}
\label{app:exp-details}
\label{app:exp-setup}
\label{app:exp-coverage}
\label{app:exp-paired}

\subsection{Grader Outcomes and Recovery of Formatting-Rejected Files}
\label{app:exp-recovery}

\vspace{-5pt}
\paragraph{Recovery procedure.}
In our main evaluation (Section~\ref{sec:run-outcomes}), 94 official public runs and 3 official private runs produced a \texttt{rankings.json} file that the strict grader rejected due to JSON schema departures (Figure~\ref{fig:public-rejected-files}a). No language model or prompt was used to recover or rewrite these files. Instead, because every departure merely renamed or renested JSON keys while leaving the ordered lists of experiment identifiers intact, we recovered the orderings using a deterministic Python schema-normalization parser that maps the five observed structural variants back to the canonical schema \texttt{\{"task\_id": ..., "rankings": [\{"group\_id": ..., "order": [...]\}]\}}:
\begin{enumerate}[leftmargin=*,itemsep=1pt,topsep=2pt]
  \item \emph{Object keyed by group ID} (38 public files): \texttt{rankings} is a dictionary mapping each \texttt{group\_id} directly to its ordered list of experiment IDs (or to a sub-object containing the ordering list) instead of a list of group objects.
  \item \emph{Alternative ordering key} (31 public files): each group object inside \texttt{rankings} stores the ordered list of experiment IDs under an alternative key name rather than \texttt{order}, most often \texttt{item\_ids} and otherwise \texttt{ranking}, \texttt{ranked\_item\_ids}, \texttt{items}, \texttt{ordering} or \texttt{ordered\_item\_ids}.
  \item \emph{Alternative top-level key} (20 public files): the list or dictionary of comparison groups is placed under a top-level key other than \texttt{rankings}, \texttt{groups} in every such file but one and \texttt{predictions} in that one.
  \item \emph{Alternative group identifier key} (4 public files): each group object uses \texttt{group} instead of \texttt{group\_id} to name the comparison group.
  \item \emph{Renamed identity fields} (1 public file): the task identity keys use variant field names, \texttt{task} and \texttt{version} in place of \texttt{task\_id} and \texttt{task\_version}, while preserving the exact task, group and experiment IDs.
\end{enumerate}
The parser extracts the ordered list of experiment IDs for each comparison group without altering any item's position and passes the normalized JSON to the standard grader of Section~\ref{sec:evaluation}. Two independent implementations of this deterministic recovery yield identical rankings and task scores agreeing to within $10^{-16}$. Table~\ref{tab:outcomes-grader} compares configuration means on the 100 public tasks before and after recovering these formatting-rejected files.

\vspace{-5pt}
\paragraph{Runs without a usable ranking.}
Ten of the 15 official public runs without a usable ranking (Section~\ref{sec:run-outcomes}) stopped under the information policy: eight wrote no file, and two runs of GPT-6.1 Sol wrote an empty list of rankings. Two runs of Kimi-K3.1-Pro and two of Sonnet 5.5 ended without a submission and without citing the policy. The fifteenth wrote a file the grader accepted and was then killed at the 16\,GiB memory limit, so it is recorded as a memory limit rather than a submission (Table~\ref{tab:outcomes-grader}).

\begin{table}[!htbp]
  \centering
  \caption{\textbf{Outcomes and the grader's scoring on the public tasks.} Each configuration has 100 official runs. Mean at $-1$ is the grader's default convention, which scores every run without a valid file at $-1$; the next column replaces the formatting-rejected files by their recovered scores, and Scored mean is the raw mean over the runs that remain. Five runs left a valid file before refusing, and the grader scored those files rather than giving them $-1$. Two runs of GPT-6.1 Sol stopped under the information policy with an empty list of rankings, which the grader also scores at $-1$; they are counted under No file, as is the run of GPT-6.1 Sol that the memory limit killed after the grader had accepted its file. Kimi-K3.1-Pro's five runs without a file are three information-policy stops and two runs that ended without a submission, and Sonnet 5.5's two also ended without a submission.}
  \label{tab:outcomes-grader}
  \small
  \begin{tabular}{lrrrrrr}
    \toprule
    Configuration & Rejected & Refusal & No file & Mean at $-1$ & After recovery & Scored mean \\
    \midrule
    Astra Medium & 0 & 0 & 0 & 0.431 & 0.431 & 0.431 \\
    Astra Max & 0 & 0 & 0 & 0.409 & 0.409 & 0.409 \\
    GPT-6.1 Sol & 0 & 2 & 4 & 0.259 & 0.259 & 0.308 \\
    Opus 5 & 18 & 1 & 0 & 0.168 & 0.421 & 0.422 \\
    Sonnet 5.5 & 0 & 1 & 2 & 0.347 & 0.347 & 0.388 \\
    DeepSeek & 3 & 0 & 0 & 0.221 & 0.254 & 0.254 \\
    Qwen3.8 & 8 & 0 & 2 & 0.110 & 0.199 & 0.224 \\
    GLM-5.3 & 16 & 1 & 1 & 0.025 & 0.217 & 0.242 \\
    HY4 & 12 & 0 & 0 & 0.079 & 0.245 & 0.245 \\
    Doubao & 33 & 0 & 0 & $-0.126$ & 0.315 & 0.315 \\
    Kimi-K3.1-Pro & 4 & 0 & 5 & 0.174 & 0.223 & 0.287 \\
    Opus 5.5 & 0 & 56 & 0 & $-0.329$ & $-0.329$ & 0.491 \\
    Fable 5.1 & 0 & 69 & 1 & $-0.537$ & $-0.537$ & 0.453 \\
    \bottomrule
  \end{tabular}
\end{table}

\subsection{Paired Comparisons and Run-to-Run Variation}
\label{app:exp-runs}

\begin{figure}[!htbp]
  \centering
  \includegraphics[width=\linewidth]{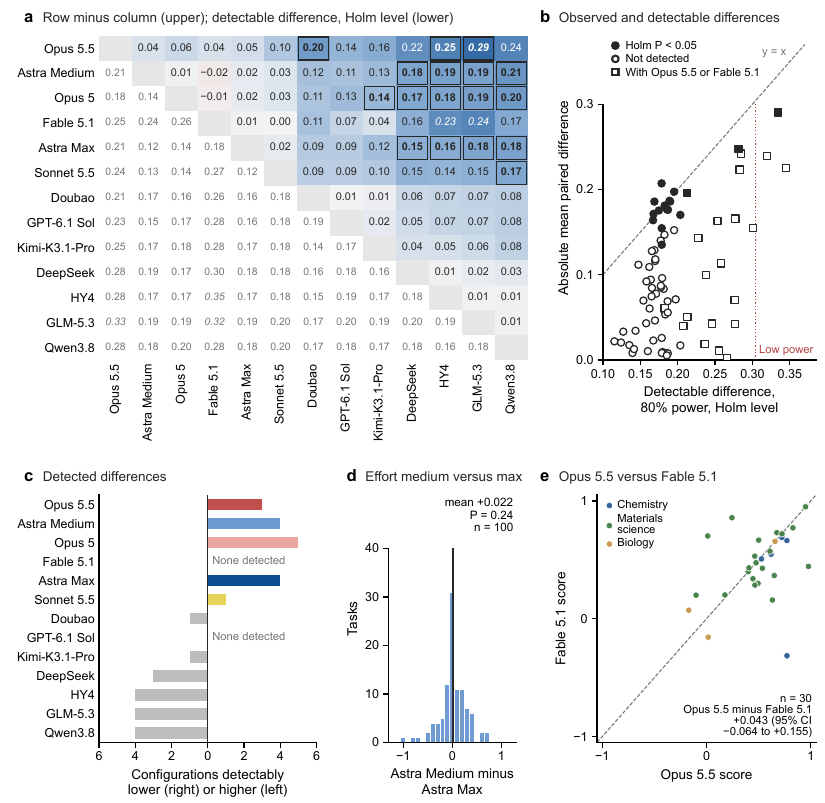}
  \caption{\textbf{Paired comparisons on the public tasks.} \textbf{a}, Upper triangle, mean paired difference (row minus column) on the tasks scored for both configurations, with bold boxes for pairs that pass Holm correction (two-sided Wilcoxon signed-rank test); lower triangle, difference detectable with 80\% power at $\alpha=0.05/78$, the strictest level the Holm correction applies; pairs whose detectable difference exceeds 0.304 are in italics in both triangles. \textbf{b}, Observed against detectable difference for each pair. \textbf{c}, Number of configurations detectably below (right) or above (left) each configuration. \textbf{d}, Per-task difference between Astra Medium and Astra Max, with the uncorrected Wilcoxon $P$. \textbf{e}, Scores of Opus 5.5 and Fable 5.1 on their 30 shared tasks.}
  \label{fig:public-paired}
\end{figure}

\begin{figure}[!htbp]
  \centering
  \includegraphics[width=\linewidth]{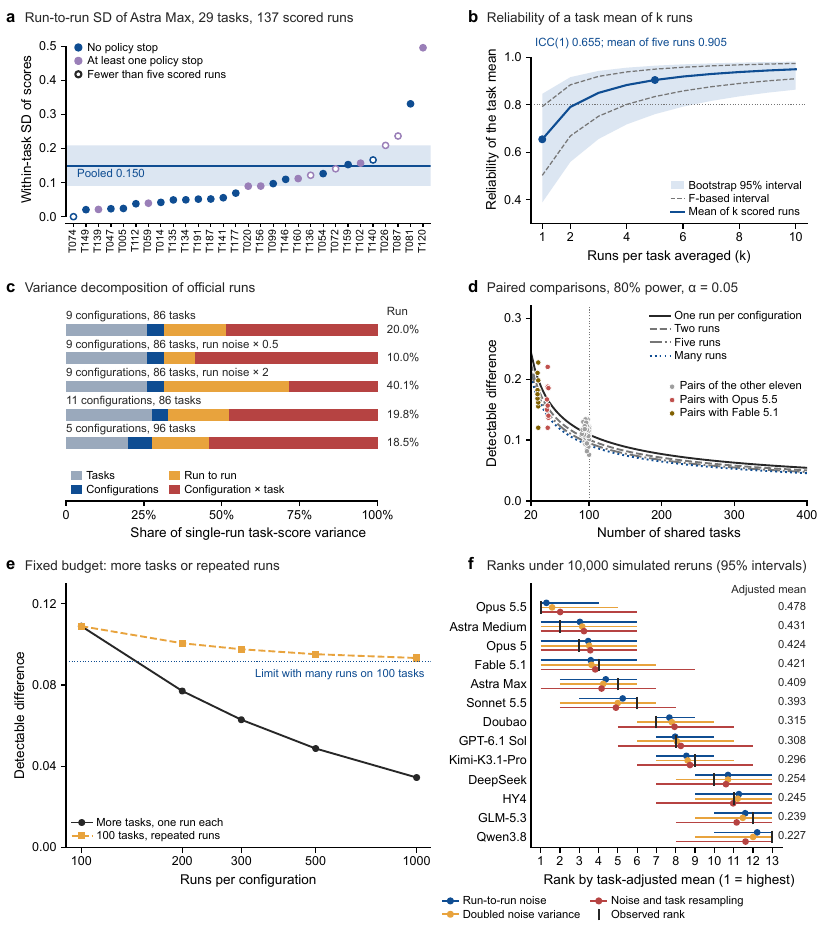}
  \caption{\textbf{Run-to-run variation and rank stability on the public tasks.} \textbf{a}, Standard deviation of Astra Max scores across the repeated runs on each of 29 tasks. \textbf{b}, Reliability of a task mean over k runs. \textbf{c}, Shares of the variance of single-run task scores due to tasks, configurations, run-to-run noise and the configuration-by-task interaction. \textbf{d}, Difference detectable with 80\% power at an uncorrected $\alpha=0.05$, against the number of shared tasks. \textbf{e}, Detectable difference at the same power and level, for a fixed number of runs spent on new tasks or on repeated runs of 100 tasks. \textbf{f}, Ranks of the thirteen task-adjusted means in 10,000 simulated reruns, with 95\% intervals and the observed ranks (ticks).}
  \label{fig:public-runs}
\end{figure}

Figures~\ref{fig:public-paired} and~\ref{fig:public-runs} provide additional diagnostics on pairwise configuration comparisons (Section~\ref{sec:public-results}), run-to-run score variation and rank stability (Section~\ref{sec:repeated-runs}), and the variance decomposition across tasks and configurations (Section~\ref{sec:analysis-separation}).

\vspace{-5pt}
\paragraph{Detectable differences.}
For each pair, the smallest difference detectable with 80\% power is $z\,s_d/\sqrt{n}$, where $s_d$ is the standard deviation of the paired differences, $n$ the number of shared tasks and $z$ the sum of the normal quantiles for the power and the test level: 4.26 at $\alpha=0.05/78$, the strictest level the Holm correction applies, and 2.80 at an uncorrected $\alpha=0.05$. We report the corrected value (Figure~\ref{fig:public-paired}a, b). No difference is detected within either group of Section~\ref{sec:public-results}, where the gaps fall below the minimum detectable effect of a single-run design, which the Holm correction puts between 0.115 and 0.266 at 80\% power.

\subsection{Effort}
\label{app:exp-effort}
\label{app:exp-two-efforts}

Median run times among scored official runs are 0.26\,h for Astra Medium, 0.42\,h for DeepSeek, 0.78\,h for Fable 5.1, 1.15\,h for Kimi-K3.1-Pro, 1.20\,h for Astra Max, 1.26\,h for Qwen3.8, 1.27\,h for Doubao, 1.65\,h for HY4, 1.81\,h for Opus 5.5, 1.95\,h for GLM-5.3, 2.14\,h for Sonnet 5.5, 3.58\,h for Opus 5 and 3.59\,h for GPT-6.1 Sol, and median tool calls range from 29 for Astra Medium to 231 for Opus 5. We tested the correlation of score with log run time, log tool calls, log web fetches and GPU use in each of the thirteen configurations, with Holm correction over the thirteen for each measure; none of the 52 correlations is detected, and the three with an uncorrected $P$ below 0.05 are Astra Medium's run time and tool calls ($\rho=0.21$ and 0.22) and GPT-6.1 Sol's run time ($\rho=0.24$). With task and configuration effects removed, the correlation between score and log run time is 0.043 (95\% CI $-0.026$ to 0.113), and those with log tool calls, log web fetches and GPU use are 0.031 to 0.040, each with a 95\% CI that includes zero.

\subsection{Information Policy}
\label{app:exp-policy}

\vspace{-5pt}
\paragraph{Detector.}
The detector of Section~\ref{sec:audit} scans tool outputs for the DOI of the source study, its PubMed or PMC identifier, its full title or most of its title, and the DOI or title of a preprint version. By configuration, it flags 23 runs of Doubao, 19 of Kimi-K3.1-Pro, 17 of HY4, 12 of GPT-6.1 Sol, 9 of GLM-5.3, 8 each of Qwen3.8 and Opus 5, 4 of Sonnet 5.5, 3 each of Astra Medium and DeepSeek, 2 of Fable 5.1 and 1 each of Astra Max and Opus 5.5. The flagged runs exceed the chance threshold in 41 of 101 scored runs (across 49 tasks), against 196 of the 462 unflagged runs on those tasks, with no configuration showing a significant difference after Holm correction. The pooled difference of +0.053 in Section~\ref{sec:audit} is driven by Opus 5 (+0.203 over 8 flagged runs; Holm $P=0.27$) and Doubao (+0.128 over 23; Holm $P=0.26$), while five of the thirteen configurations vary in the opposite direction.

\vspace{-5pt}
\paragraph{Presented order.}
The score of the presented order is computed with the grader's rule from the public input order of each task; it beats the chance threshold on 6 of the 100 public tasks, matching random expectation. Twenty-two scored runs on 15 tasks submitted that order (11 by GPT-6.1 Sol, 3 by Astra Max, 2 by DeepSeek and one each by Astra Medium, Doubao, GLM-5.3, HY4, Kimi-K3.1-Pro and Qwen3.8). Removing all 22 raises GPT-6.1 Sol's raw mean from 0.308 to 0.348 and changes no other configuration's raw mean by more than 0.007.

\subsection{Private Tasks}
\label{app:exp-private}

\vspace{-5pt}
\paragraph{Exclusions.}
Seven runs that hit the 16\,GiB memory limit during ESMFold structure prediction were rerun and are scored (DeepSeek 2, Doubao 1, HY4 3 and Qwen3.8 1). The run that reached the 8-hour limit is from Qwen3.8 on the electrochemical window task. The three recovered files are from GLM-5.3 on the single-substitution expression task and from Opus 5 on the ELISA response and SPR affinity tasks. Opus 5 on the single-substitution binding task, GPT-6.1 Sol and Sonnet 5.5 on the combination expression task, Opus 5.5 on the \textit{N}-sulfonylation task and Fable 5.1 on the \textit{N}-benzylation task wrote a valid file and then refused, so all five runs are excluded as refusals.

\vspace{-5pt}
\paragraph{Scores by task.}
The electrochemical window task has a mean of $-0.347$. The three reaction-condition tasks average 0.488 (\textit{N}-benzylation), 0.615 (\textit{N}-sulfonylation) and 0.169 (reductive amination). Of the eight VHH tasks, five average 0.33 to 0.49, while the single-substitution binding and expression tasks and the combination expression task average between $-0.113$ and $-0.003$. The three Magrolimab tasks average 0.340 (CHO production), 0.639 (ELISA response) and 0.242 (SPR affinity).

\subsection{Cost Estimation}
\label{app:exp-cost}

\begin{table}[!htbp]
  \centering
  \caption{\textbf{List prices used for the cost estimate.} USD per million tokens, checked on 20 October 2026 (GPT-6.1 Sol and Claude Opus 5 on 23 October, Kimi-K3.1-Pro on 26 October, Claude Sonnet 5.5 on 29 October). GPT-6.1 Sol: short-context rates; 26 of its 22,608 requests exceed 272K input tokens and are billed at the long-context rates (4.00, 0.20, 5.00 and 15.00). Cache writes are billed at the five-minute rate; the one-hour rate is given in parentheses where it differs. n/a: the provider reports no cache writes, and tokens not read from the cache are billed at the input price. DeepSeek: off-peak rates. Qwen3.8: international rates for explicit caching. Doubao: converted from yuan at 6.72 CNY per USD.}
  \label{tab:prices}
  \small
  \begin{tabular}{lrrrr}
    \toprule
    Model & Input & Cache read & Cache write & Output \\
    \midrule
    GPT-6 Astra & 10.00 & 1.00 & 12.50 & 50.00 \\
    GPT-6.1 Sol & 2.00 & 0.10 & 2.50 & 10.00 \\
    Claude Opus 5.5 & 4.00 & 0.20 & 5.00 (8.00) & 20.00 \\
    Claude Opus 5 & 5.00 & 0.50 & 6.25 (10.00) & 25.00 \\
    Claude Sonnet 5.5 & 2.00 & 0.10 & 2.50 (4.00) & 10.00 \\
    DeepSeek-V4.1-Flash-0910 & 0.15 & 0.003 & n/a & 0.60 \\
    Qwen3.8-Max-0902 & 2.00 & 0.17 & 2.50 & 6.00 \\
    GLM-5.3-Flash & 0.15 & 0.03 & n/a & 0.50 \\
    Doubao-Seed-Evolving-0909 & 0.893 & 0.179 & n/a & 4.464 \\
    HY4-Preview & 0.834 & 0.042 & n/a & 2.501 \\
    Kimi-K3.1-Pro & 2.00 & 0.20 & n/a & 10.00 \\
    Claude Fable 5.1 & 10.00 & 0.25 & 12.50 (20.00) & 50.00 \\
    \bottomrule
  \end{tabular}
\end{table}

Table~\ref{tab:prices} lists the provider token rates used to estimate API costs. Each run's cost sums uncached input, cache-write, cache-read and output tokens at these list prices, excluding container and GPU infrastructure overhead.

\subsection{Trajectory Analysis}
\label{app:exp-trajectory}

\vspace{-5pt}
\paragraph{Activities and method families.}
Each tool call receives one primary activity by a fixed priority: subagent, web search, web fetch or shell download, software installation, scientific computation, submission files, other Python, code writing, notes and data files, reading the task inputs, environment probes, waiting and monitoring, viewing files, planning and time checks, and other. Scientific computation is executed Python that imports a module of a method family, calls a fitting, rate-equation or sequence-alignment routine, draws Monte Carlo samples or runs a GPU workload, or a program of a method family that actually runs. A call is credited with the time from its step to the next step, divided evenly among the calls of the step. Figure~\ref{fig:analysis-workflow}a merges the activities into eight classes; in a random sample of 60 calls checked by hand, 56 labels were correct.

\vspace{-5pt}
\paragraph{Rankings and consensus.}
To find where an ordering first appears (Section~\ref{sec:analysis-rankings}), we search the trajectory in time order for the item identifiers of the group in the submitted order: in the model's messages and reasoning, in the code and arguments of each call, and in each tool output. A run's first draft (first complete ranking) is the earliest ranking in its trajectory that has the submission format, contains every group and every item, and differs from the input order. For the monotone prior, each eligible group is compared with every numerical column of the public inputs that takes at least three values within the group, keeping the largest absolute Spearman correlation. In the leave-one-out consensus comparison (Appendix~\ref{app:exp-agreement}), the consensus for a configuration uses only the other configurations scored on the same task, with ties broken by item identifier.

\vspace{-5pt}
\paragraph{Score associations and classifier.}
For each behavioral or method feature, the score of a run is regressed on the feature with task and configuration fixed effects by least squares over the 1,155 scored runs, with intervals and two-sided $P$ values from 10,000 bootstrap resamples of tasks and Holm correction applied separately to the 16 method features and 8 behavioral features. Before correction, two method families have intervals above zero: machine-learned interatomic potentials (+0.141 over 29 runs) and DFT (+0.070 over 258). The configuration classifier (Section~\ref{sec:analysis-profiles}) is a 500-tree random forest evaluated by stratified five-fold cross-validation on the 861 scored runs of the ten Claude Code configurations.

\subsection{Agreement Between Configurations}
\label{app:exp-agreement}

\begin{figure}[!htbp]
  \centering
  \includegraphics[width=\linewidth]{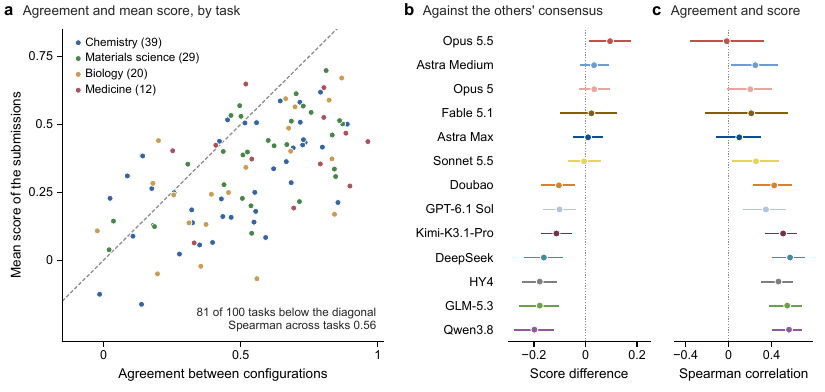}
  \caption{\textbf{Agreement between configurations.} \textbf{a}, Agreement between configurations on each public task, the mean Spearman correlation between the orders of two different configurations, against the mean score of these submissions; the dashed line marks equality. \textbf{b}, Mean score of each configuration minus that of the consensus of the other configurations on the same tasks. \textbf{c}, Spearman correlation, across the runs of each configuration, between a run's agreement with that consensus and its score. Lines are 95\% bootstrap intervals over tasks.}
  \label{fig:analysis-agreement}
\end{figure}

\vspace{-5pt}
\paragraph{Shared priors.}
The configurations agree with one another more than with the measurements. With agreement measured as in Section~\ref{sec:analysis-repeats}, two configurations agree at 0.52 on an average task (95\% CI 0.48 to 0.57), whereas their mean score against the measurements is 0.33, and on 81 of the 100 tasks they agree with one another more than with the measurements (Figure~\ref{fig:analysis-agreement}a). Agreement does carry some information about accuracy, since the tasks on which the configurations converge are also the tasks they score well on (Spearman 0.56 across tasks). It is no guarantee: on one oncology task, eleven configurations agree at 0.90 and average 0.27.

\vspace{-5pt}
\paragraph{Consensus.}
Averaging each item's position over the configurations gives a consensus ranking that scores 0.081 above the average submission (95\% CI 0.054 to 0.110), higher on 73 of the 100 tasks. Averaging cancels errors that differ between configurations and keeps those they share. Measured against the consensus of the other configurations on the same tasks, Opus 5.5 scores above that consensus and the other five of the leading six do not differ detectably, while for each of the seven below them the average ranking of the others predicts better than their own submission (Figure~\ref{fig:analysis-agreement}b). These thirteen intervals are nominal, not corrected over the set. Within most configurations, runs that agree more with that consensus also score higher (Figure~\ref{fig:analysis-agreement}c).

\vspace{-5pt}
\paragraph{Stable errors.}
Repeated runs (Section~\ref{sec:repeated-runs}) and several configurations can show which tasks are unstable, but they cannot reveal an error that all of them share. On a task that ranks six copper-gold electrodes by hydrogen-isotope incorporation during acetylene reduction, whose measured order is nearly unrelated to composition (absolute Spearman 0.03), seven configurations submitted the same order, monotone in composition, and four its exact reverse, scoring $-0.03$ and 0.03. GLM-5.3 computed reaction energies on each alloy surface with a machine-learned interatomic potential, submitted a non-monotonic order and scored 0.54.

\subsection{Tasks That Separate the Configurations}
\label{app:exp-separation}

\vspace{-5pt}
\paragraph{Spread and simulations.}
A task's spread is the standard deviation of the configuration scores on it. The comparison with run-to-run noise uses the residuals of the additive fit of Section~\ref{sec:exp-setup}, comparing their mean square on a task with the mean square of the repeated Astra Max scores about their own mean on the same task ($F$ test with Holm correction over the 28 tasks with at least three scored repeats). Null distributions without a configuration-by-task interaction are simulated from the additive fit plus independent normal noise with the pooled within-task standard deviation of 0.150 (Section~\ref{sec:repeated-runs}; 2,000 draws).

\subsection{Scores by Discipline}
\label{app:exp-discipline}

\vspace{-5pt}
\paragraph{What the quantities cover.}
A task's discipline is the one it was given when it was built (Section~\ref{sec:stats}); the 100 public tasks hold 39 chemistry, 29 materials science, 20 biology and 12 medicine tasks. A discipline mean averages the scored runs in that discipline, so a configuration with more scored tasks there weighs more, and its interval resamples the tasks of the discipline. The share above chance counts scored runs rather than tasks, and the median task size quoted in Section~\ref{sec:analysis-discipline} counts the experiments of a task, not its comparison groups.

\vspace{-5pt}
\paragraph{The difference between the groups and the two nulls.}
Astra Medium, Astra Max, DeepSeek, Qwen3.8, GLM-5.3, Doubao, HY4, Kimi-K3.1-Pro, Opus 5, GPT-6.1 Sol and Sonnet 5.5 have a scored run in all four disciplines. On each task the mean of Astra Medium, Opus 5, Astra Max and Sonnet 5.5 is taken minus the mean of DeepSeek, Qwen3.8, GLM-5.3, Doubao, HY4 and Kimi-K3.1-Pro, over the configurations that have a score on that task, and the task differences are averaged within the discipline; Opus 5.5, Fable 5.1 and GPT-6.1 Sol are left out as in Section~\ref{sec:analysis-discipline}. The discipline labels of the 100 tasks are permuted 5,000 times to test the widest-minus-narrowest gap, and ordering stability is compared against random splits of the task set into four blocks of 39, 29, 20 and 12 tasks. Chemistry holds somewhat more of the tasks that spread the configurations out most, with 13 of the 25 widest-spread tasks of Section~\ref{sec:analysis-separation} against 9.8 expected ($P=0.097$).

\subsection{Selected Runs in Detail}
\label{app:exp-review}

\vspace{-5pt}
\paragraph{Chemistry.}
On the cobalt polymerization task of Section~\ref{sec:analysis-review}, we traced the score gap to the species each run computed on rather than to the methods it used, and noted that the Astra Medium run rebuilt its structure after an unexpected hydrogen transfer and tested the extra coordination it had assumed. On the amine-borane task, one run built copper complexes in two oxidation states and an alternative aryl-transfer path, one computed boronate formation and competing reactions, and one scored acid-base equilibria with hand-set factors using aqueous constants in an organic solvent. On the cycloaddition task, the Astra Medium run, one of three tied for the top score, modeled the competition between ring closure and intermediate capture by a second substrate so dilution could act on selectivity, whereas lower-scoring runs used a conversion-times-selectivity proxy.

\vspace{-5pt}
\paragraph{Materials science.}
On the perovskite task, the recurring difficulty was mapping a computed local property to non-radiative recombination: runs used uncalibrated scoring weights or substituted parent-molecule properties for the surface ligand. On the copper-gold task (Appendix~\ref{app:exp-agreement}), the GLM-5.3 run, the only one to clear 0.029, computed adsorption energies on alloy surfaces and allowed a non-monotonic order, whereas other runs treated non-monotonic results as secondary. On the task that ranks 271 hydrogel formulations by pull-off strength (Section~\ref{sec:analysis-separation}), the Opus 5.5 run, at 0.841 against a chance threshold of 0.097, took the expected rank under a mixture of three unfitted model families, and its refinements were all about the test interface rather than the bulk gel: the acidification of the unbuffered conditioning saline by the gel's own monomer, the charge of the glass in contact following that pH, and a partial re-equilibration of the gel surface during the immersion before contact.

\vspace{-5pt}
\paragraph{Biology.}
On the tomato task, the Astra Max run withdrew a sequence feature when a randomized control did not support it; on the aptamer task, the HY4 run, which scored $-0.120$, guessed the parent sequence and assumed every mutation was at best neutral; and on the isotope-tracing task, adding expression proxies did not close the gap between the model and the actual culture conditions. The DeepSeek run, which scored highest on the aptamer task at 0.674 against a threshold of 0.131, started from the direction of the measurement: the fluorescence reported is quenched on binding, so the strongest binder gives the lowest reading, and the ordering by affinity is then the same at every RNA concentration the task varies.

\vspace{-5pt}
\paragraph{Medicine.}
On the mixed-culture task of Section~\ref{sec:analysis-review}, the block structure the leading run submitted rests on the covalent target of each drug, and its own report marks the ordering within each block as far less certain than the blocks themselves. On a task that ranks lipogenic transcripts across a KRAS state and two PIKfyve inhibitors, the Doubao run, one of three that scored 0.950 against a threshold of 0.650, built its order from a homeostatic route, in which blocked lysosomal lipid export deprives the membrane of lipid and activates the sterol-regulatory machinery that co-induces the two enzymes measured. It also wrote down the routes that oppose it, including lipid catabolism through TFEB and the albumin-bound fatty acids in serum that bypass the lysosome, and it checked with a first-order turnover band that an 8-hour transcriptional response can be as large as the chronic difference between the two KRAS states, so the middle of the order could not be assumed away.

\vspace{-5pt}
\paragraph{Checks that miss the endpoint.}
Among the runs that scored poorly, the recurring pattern is that the run verifies something other than the quantity the task asks for. A DeepSeek run reported that redocking recovered the reference binding pose and stated high confidence in its weakest-binding call, then scored $-0.600$ on a task where four configurations scored 0.800. An assumption built into a model also survives any amount of sampling inside it: all six candidate orderings that a DeepSeek run compared on the cycloaddition task placed the four low-temperature conditions ahead of the four at 30 degrees Celsius, so the condition with the highest measured yield could not be selected however many samples it drew. A calculation can move a ranking the wrong way as well: on the amine-borane task, the Astra Max run found an alternative aryl-transfer path favorable, raised one additive on that basis, and fell from 0.818 to 0.747.

\subsection{GPU Use}
\label{app:exp-gpu}

A call counts as GPU work when the executed code places a computation on the device through a CUDA tensor or array, a GPU build of a quantum-chemistry or dynamics package, or a command-line program given a CUDA device; the device is confirmed when the program's own output names it (excluding \texttt{nvidia-smi} probes). The GPU calls of a run that used the device take a median of 0.10 hours (2.5\% of its run time, a lower bound because background jobs are undercounted).

\section{Scores on Individual Runs}
\label{app:score-examples}

\begin{figure}[!t]
  \centering
  \includegraphics[width=\linewidth]{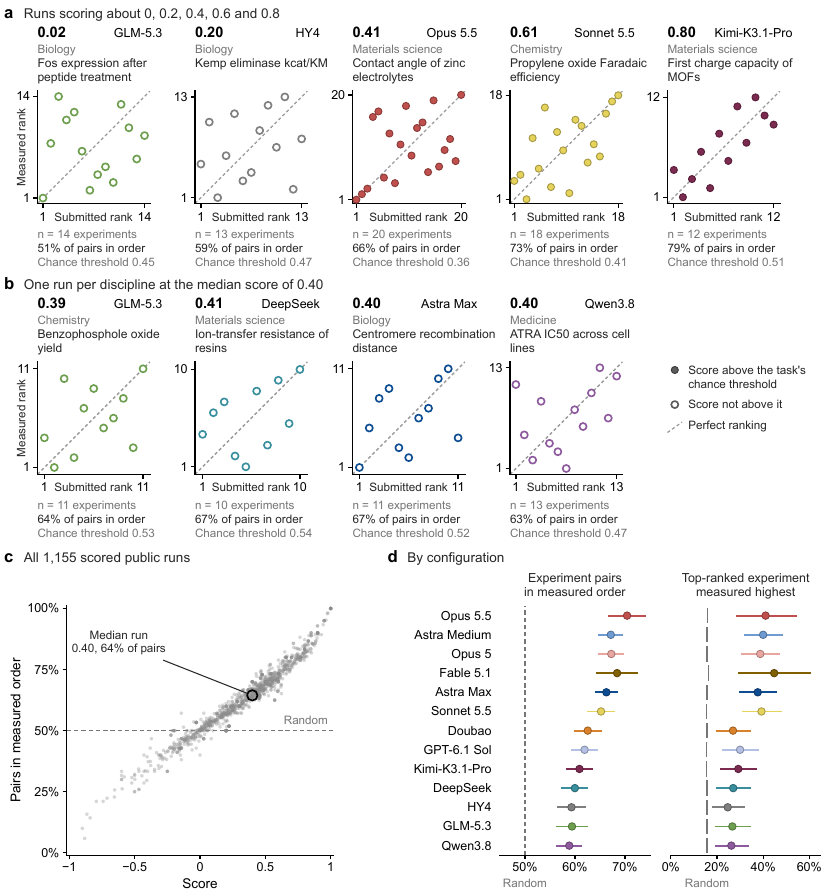}
  \caption{\textbf{Scores on individual runs.} \textbf{a}, Measured rank against submitted rank for five official runs scoring about 0, 0.2, 0.4, 0.6 and 0.8, where rank 1 is the lowest. \textbf{b}, The same for one run per discipline at the median score of 0.40. \textbf{c}, Share of experiment pairs in the measured order against score for all 1,155 scored public runs, with the median run circled. \textbf{d}, By configuration, mean share of pairs in the measured order and share of runs whose top-ranked experiment is the measured top, with 95\% bootstrap intervals over tasks.}
  \label{fig:score-examples}
\end{figure}

A task score is the Spearman correlation between the submitted and measured orders. Figure~\ref{fig:score-examples} shows what such scores look like on real runs.

\vspace{-5pt}
\paragraph{Example runs.}
The examples are runs on public tasks with one comparison group, no tied measurements and 10 to 20 experiments (125 runs on 12 tasks), so each score is a single Spearman correlation. The first row takes the run closest to each of 0, 0.2, 0.4, 0.6 and 0.8, and the second takes one run per discipline closest to 0.40, the median score of all scored public runs; no task is used twice. Near 0.4 the submitted order follows the measured one only loosely. The four median runs, with 10 to 13 experiments, all stay below their tasks' chance thresholds of 0.47 to 0.54, while the first-row run at 0.41, with 20 experiments, clears a threshold of 0.36.

\vspace{-5pt}
\paragraph{Pairs in the measured order.}
A more direct reading is the share of experiment pairs with different measured values that a run places in the measured order, weighting groups as in the task score. It rises almost linearly with the score ($r=0.98$; Figure~\ref{fig:score-examples}c) and has a median of 64\% over the scored runs, against 50\% for random rankings. Configuration means range from 59\% (Qwen3.8) to 70\% (Opus 5.5), so even the leading configuration gets about three pairs in ten wrong; every 95\% interval lies above 50\% (Figure~\ref{fig:score-examples}d). The top-ranked experiment of a group is the measured top 25\% (HY4) to 45\% (Fable 5.1) of the time, against 16\% at random.

\end{document}